\documentclass[twoside,11pt]{article}
\usepackage{jmlr2e}

\jmlrheading{18}{2017}{1-45}{7/16; Revised 10/17}{11/17}{16-374}{Christophe Dupuy and Francis Bach}

\ShortHeadings{Online Inference for Latent Variable Models}{Dupuy and Bach}
\firstpageno{1}

\usepackage{times}
\usepackage{graphicx}
\usepackage{float}

\usepackage{subfigure} 

\usepackage{color}

\usepackage[mathcal]{eucal}

\usepackage{natbib}

\usepackage{algorithm}
\usepackage{algorithmic}

\usepackage{hyperref}

\usepackage{amsmath}
\usepackage{amsfonts}
\usepackage{stmaryrd}
\usepackage{enumitem}
\usepackage[normalem]{ulem}
\usepackage[figuresright]{rotating}

\usepackage{footnote}
\usepackage{empheq}

\title{Online but Accurate Inference for Latent Variable Models \\with Local Gibbs Sampling}

\newcommand{\mysec}[1]{Section~\ref{sec:#1}}

\begin{document}
\author{Christophe Dupuy \email chirstophe.ih.dupuy@gmail.com\\
\addr Technicolor\\
INRIA\\
D´epartement d'Informatique de l'ENS, Ecole normale supérieure, CNRS, PSL Research University\\
Paris, France.
       \AND
       \name Francis Bach \email francis.bach@ens.fr\\
\addr INRIA\\
D´epartement d'Informatique de l'ENS, Ecole normale supérieure, CNRS, PSL Research University\\
Paris, France.}
\editor{Franc\c{o}is Caron}

\maketitle

\begin{abstract}%
We study parameter inference in large-scale latent variable models.~We first propose a unified treatment of online inference for latent variable models from a non-canonical exponential family, and draw explicit links between several previously proposed frequentist or Bayesian methods.~We then propose a novel inference method for the frequentist estimation of parameters, that adapts MCMC methods to online inference of latent variable models with the proper use of local Gibbs sampling.~Then, for latent Dirichlet allocation,we provide an extensive set of experiments and comparisons with existing work, where our new approach outperforms all previously proposed methods.
In particular, using Gibbs sampling for latent variable inference is superior to variational inference in terms of test log-likelihoods.~Moreover, Bayesian inference through variational methods perform poorly, sometimes leading to worse fits with latent variables of higher dimensionality.
\end{abstract} 


\section{Introduction}

Probabilistic graphical models provide general modelling tools for complex data, where it is natural to include assumptions on the data generating process by adding latent variables in the model. Such latent variable models are  adapted to a wide variety of unsupervised learning tasks~\citep{koller2009probabilistic,murphy2012machine}.
In this paper, we focus on parameter inference in such latent variable models where the main operation needed for the standard expectation-maximization (EM) algorithm is {intractable}, namely  dealing with   conditional distributions over latent variables given the observed variables; latent Dirichlet allocation (LDA)~\citep{LDA} is our motivating example, but many hierarchical models exhibit this behavior, e.g., ICA with heavy-tailed priors. For such models, there exist two main classes of methods to deal efficiently with intractable exact inference in large-scale situations: sampling methods or variational methods.

\emph{Sampling methods} can  handle arbitrary distributions and lead to simple inference algorithms while converging to exact inference. However it may be slow to converge and non scalable to big datasets in practice.  In particular, although efficient implementations have been developed, for example for LDA \citep{SAME2014,GPUGibbsLDA2009}, MCMC methods may not deal efficiently yet with continuous streams of data for our general class of models.  

On the other hand, \emph{variational inference} builds an approximate model for the posterior distribution over latent variables---called variational---and infer parameters of the true model through this approximation. The fitting of this variational distribution is formulated as an optimization problem where efficient (deterministic) iterative techniques such as gradient or coordinate ascent methods apply. This approach leads to scalable inference schemes~\citep{SVI2013}, but due to approximations, there always remains a gap between the variational posterior and the true posterior distribution, inherent to algorithm design, and that will not vanish when the number of samples and the number of iterations increase.

Beyond the choice of approximate inference techniques for latent variables, parameter inference may be treated either from the \emph{frequentist} point of view, e.g., using maximum likelihood inference, or a \emph{Bayesian} point of view, where the posterior distribution of the parameter given the observed data is approximated. With massive numbers of observations, this posterior distribution is typically peaked around the maximum likelihood estimate, and the two inference frameworks should not differ much~\citep{van2000asymptotic}.

In this paper, we focus on methods that make a single pass over the data to estimate parameters. We make the following contributions:
\begin{enumerate}
\item We review and compare existing methods for online inference for latent variable models from a non-canonical exponential family in \mysec{method}, and draw explicit links between several previously proposed frequentist or Bayesian methods. Given the large number of existing methods, our unifying framework allows to understand differences and similarities between all of them.
\item We propose in \mysec{intract} a novel inference method for the frequentist estimation of parameters, that adapts MCMC methods to online inference of latent variable models with the proper use of ``local'' Gibbs sampling. In our online scheme, we apply Gibbs sampling to the current observation, which is ``local'', as opposed to ``global'' batch schemes where Gibbs sampling is applied to the entire dataset.
\item After formulating LDA as a non-canonical exponential family in \mysec{LDA}, we provide an extensive set of experiments   in \mysec{results}, where our new approach outperforms all previously proposed methods. In particular, using Gibbs sampling for latent variable inference is superior to variational inference in terms of test log-likelihoods. Moreover, Bayesian inference through variational methods perform poorly, sometimes leading to worse fits with latent variables of higher dimensionality.
\end{enumerate}

\section{Online EM}
\label{sec:method}
We consider an \emph{exponential family} model on random variables $(X,h)$ with parameter $\eta\in\mathcal{E}\subseteq{\mathbb{R}^d}$ and with density \citep{ExpFam1998}: 
\begin{equation}
p(X,h|\eta) = a(X,h)\exp\big[\langle\phi(\eta), S(X,h)\rangle -\psi(\eta)\big].
\label{eqn:exp_fam}
\end{equation}
We assume that $h$ is hidden and $X$ is observed. The vector $\phi(\eta) \in \mathbb{R}^d $ represents the natural parameter, $S(X,h) \in \mathbb{R}^d$ is the vector {of} sufficient statistics, $\psi(\eta)$ is the log-normalizer, and $a(X,h)$ is the underlying base measure. We consider a \emph{non-canonical} family as in many models (such as LDA), the natural parameter $\phi(\eta)$ does not coincide with the model parameter $\eta$, that is,  $\phi(\eta)\not\equiv\eta$; we however assume that $\phi$ is injective.

We consider $N$ \emph{i.i.d.}~observations  $(X_i)_{i=1,\ldots,N}$ from a distribution $t(X)$, which may be of the form {$P(X|\eta^*)=\int_h p(X,h|\eta^*)\mathrm{d}h$} for our model above and a certain  $\eta^* \in \mathcal{E}$ (well-specified model) or not (misspecified model). Our goal is to obtain a predictive density $r(X)$ built from the data and using the model defined in (\ref{eqn:exp_fam}), with the maximal expected log-likelihood
$ \mathbb{E}_{t(X)} \log r(X)$.

\subsection{Maximum Likelihood Estimation}

In the frequentist perpective, the predictive distribution $r(X)$ is of the form $p(X|\hat{\eta})$, for a well-defined estimator $\hat{\eta} \in \mathcal{E}$. The most common method is the EM algorithm \citep{EM1977}, which is an algorithm that aims at maximizing the likelihood of the observed data, that is,
\begin{equation}
\label{eqn:ML}
\max\limits_{\eta\in\mathcal{E}} \;\; \sum_{i=1}^N \log p(X_i|\eta).
\end{equation}
More precisely, the EM algorithm is an iterative process to find the maximum likelihood (ML) estimate given observations $ (X_i)_{i=1,\ldots,N}$ associated to hidden variables $(h_i)_{i=1,\ldots,N} $. It
may be seen as the iterative construction of lower bounds of the log-likelihood function~\citep{bishop2006pattern}. In the exponential family setting~(\ref{eqn:exp_fam}), we have, by Jensen's inequality, given the model defined by $\eta' \in \mathcal{E}$ from the previous iteration, and for any parameter $\eta \in \mathcal{E}$:
\begin{align*}
\log p(X_i|\eta)  & = \log \int p(X_i,h_i | \eta) d h_i \\[-.1cm]
& \geq   \int \!  p(h_i|X_i,\eta') \log  \frac{ p(X_i,h_i | \eta)}{ p(h_i|X_i,\eta')} d h_i  \\[-.1cm]
& =  { \int  \!\! p(h_i|X_i,\eta')  \left(\langle \phi(\eta),S(X_i,h_i)\rangle \!-\! \psi(\eta) \right)d h_i \! -\! C_i(\eta') } \\[-.1cm]
& =    \langle \phi(\eta), \mathbb{E}_{p( {h_i}|X_i,\eta')}\left[S(X_i, {h_i})\right] \rangle - \psi(\eta)
- C_i(\eta'),
\end{align*}
for a certain constant $C_i(\eta')$,
with equality if $\eta'  = \eta$. Thus, EM-type algorithms build locally tight lower bounds of the log-likelihood in (\ref{eqn:ML}), which are equal to 
\[ \textstyle \langle \phi(\eta), \sum_{i=1}^N s_i  \rangle - N \psi(\eta) + \mbox{cst},\] 
for appropriate values of $s_i \in \mathbb{R}^d$ obtained by computing conditional expectations with the distribution of $h_i$ given ${X_i}$ for the current model defined by $\eta'$ (E-step), i.e., ${s_i =  \mathbb{E}_{p({h_i}|X_i,\eta')}\left[S(X_i,{h_i})\right]}$. Then this function of~$\eta$ is maximized to obtain the next iterate (M-step). In standard EM applications, these two steps are assumed tractable. In \mysec{intract}, we will only assume that the M-step is tractable while the E-step is intractable.

Standard EM will consider ${s_i =   \mathbb{E}_{{p(h_i|X_i,\eta')}}\left[S(X_i,h)\right]}$ for the previous value of the parameter~$\eta$ for all~$i$, and hence, at every iteration, all observations $X_i$, $i=1,\dots,N$ are considered for latent variable inference, leading to a slow ``batch'' algorithm for large $N$.

Incremental EM~\citep{incrementalEM1998} will only update a single element $s_i$ coming from  a single observation $X_i$ and update the corresponding part of the sum $\sum_{j=1}^N s_j$ without changing other elements. In the extreme case where a single pass over the data is made,
then the M-step at iteration $i$ maximizes  
\[
\langle \phi(\eta), \sum_{j=1}^i \mathbb{E}_{p(h_j|X_j,\eta_{j-1})}\left[S(X_j,h_j)\right] \rangle - i \psi(\eta),
\]
 with respect to $\eta$. In the next section, we provide a (known) other interpretation of this algorithm.

\subsection{Stochastic Approximation}

Given our frequentist objective $\mathbb{E}_{t(X)} \log p(X|\eta)$ to maximize defined as an expectation, we may consider two forms of stochastic approximation~\citep{kushner:yin:2003}, where observations $X_i$ sampled from $t(X)$ are processed only once.
 The first one is stochastic gradient ascent, of the form
\[ \eta_{i} =   \eta_{i-1} + \rho_i \frac{\partial \log p(X_i| \eta)}{\partial \eta},\]
or appropriately renormalized version thereof, i.e.,  ${\eta_{i} =   \eta_{i-1} + \rho_i H^{-1} \frac{\partial \log p(X_i| \eta)}{\partial \eta}}$,
with several possibilities for the $d \times d$ matrix $H$, such as the negative Hessian of the partial or the full log-likelihood, or the negative covariance matrix of gradients, which can be seen as versions of natural gradient---see \citet{Titterington,delyon1999,OnlineEM}. This either leads to slow convergence (without~$H$) or expensive iterations (with~$H$), with the added difficulty of choosing a proper scale and decay for the step-size $\rho_i$.

A key insight of~\citet{delyon1999,OnlineEM} is to use a different formulation of stochastic approximation, \emph{not explicitly based on stochastic gradient ascent}. Indeed, they consider the stationary equation $ {\mathbb{E}_{t(X)} \big[\frac{\partial \log p(X| \eta)}{\partial \eta}\big]=0}$
and expand it using the exponential family model~(\ref{eqn:exp_fam}) as follows:
\begin{align*}
\frac{\partial \log p(X| \eta)}{\partial \eta} &  = \frac{\partial \log \int  p(X,h| \eta) d h }{\partial \eta} \\
& = \phi'(\eta) \mathbb{E}_{p(h|X,\eta) } \left[S(X,h)\right] -  {\psi'}(\eta).
\end{align*}
Given standard properties of the exponential family, namely 
\[
\psi'(\eta)=\phi'(\eta) \mathbb{E}_{p(h,X|\eta)}\left[S(X,h)\right]\footnotemark[1],
\] and assuming invertibility of $\phi'(\eta)$, this leads to the following stationary equation:
\begin{align*}
\mathbb{E}_{  t(X) } \left[ \mathbb{E}_{p(h|X,\eta) } \left[S(X,h)\right] \right] = \mathbb{E}_{p(h,X|\eta)}\left[S(X,h)\right].
\end{align*}
\footnotetext[1]{Proof: Given~(\ref{eqn:exp_fam}), $\int_{X,h}p(X,h|\eta)\mathrm{d}(X,h)=1\Rightarrow \psi(\eta)=\log\left[\int_{X,h} a(X,h)e^{\langle\phi(\eta),S(X,h)\rangle}\mathrm{d}(X,h)\right]$. We then derive this identity with respect to $\eta$, which gives:
\begin{align*}
\psi'(\eta) & = \frac{\int_{X,h} \phi'(\eta)S(X,h)a(X,h)e^{\langle\phi(\eta),S(X,h)\rangle}\mathrm{d}(X,h)}{\int_{X,h} a(X,h)e^{\langle\phi(\eta),S(X,h)\rangle}\mathrm{d}(X,h)}\\
& = \frac{\phi'(\eta)\int_{X,h} S(X,h)a(X,h)e^{\langle\phi(\eta),S(X,h)\rangle}\mathrm{d}(X,h)}{e^{\psi(\eta)}}\\
& = \phi'(\eta)\int_{X,h} S(X,h)p(X,h|\eta)\mathrm{d}(X,h)\\
& = \phi'(\eta)\mathbb{E}_{p(h,X|\eta)}\left[S(X,h)\right].
\end{align*} }

This stationary equation states that at optimality the sufficient statitics have the same expectation for the full model $p(h,X|\eta)$ and the joint ``model/data'' distribution ${t(X)p(h|X,\eta)}$. 

Another important insight of~\citet{delyon1999,OnlineEM} is to consider the change of variable ${s(\eta) = \mathbb{E}_{p(h,X|\eta)}\left[S(X,h)\right]}$ on sufficient statistics, which is equivalent to \[
 \eta = \eta^\ast(s) \in \arg\max\  \langle \phi(\eta),s \rangle - \psi(\eta),
 \]
(which is the usual M-step update). See~\citet{OnlineEM} for detailed assumptions allowing this inversion. We may then rewrite the equation above as
\[
 \mathbb{E}_{  t(X) } \big( \mathbb{E}_{p(h|X,\eta^\ast(s) ) } \left[S(X,h)\right] \big) = s.
\]
This is a non-linear equation in $s\in\mathbb{R}^d$, with an expectation with respect to $t(X)$ which is only accessed through i.i.d.~samples $X_i$, and thus a good candidate for the Robbins-Monro algorithm to solve stationary equations (and not to minimize functions)~\citep{kushner:yin:2003}, which takes the simple form:
\[
s_i = s_{i-1} - \rho_i \big(  s_{i-1} - \mathbb{E}_{p(h_i|X_i,\eta^\ast(s_{i-1}) )} \left[S(X_i,h_i)\right] \big),
\]
with a step-size $\rho_i$. It  may be rewritten as
\begin{equation}
\hspace*{-.25cm}
\left\{
\begin{array}{lcl}
s_{i} & \!\!\!= \!\!\!& (1-\rho_i)s_{i-1} +\rho_i\mathbb{E}_{p(h_i|X_{i},\eta_{i-1})}[S(X_{i},h_i)]\\
\eta_{i} &\!\!\!= \!\!\! &\eta^*(s_{i}),
\end{array}
\right.
\hspace*{-.25cm}
\label{eqn:OEM}
\end{equation}
which has a particularly simple interpretation:
instead of computing the   expectation for all observations as in   full EM, this stochastic version keeps tracks of old sufficient statistics through the variable~$s_{i-1}$ which is updated towards the current value ${\mathbb{E}_{p(h_i|X_{i},\eta_{i-1})}[S(X_{i},h_i)]}$. The parameter~$\eta$ is then updated to the value $\eta^*(s_{i})$. \citet{OnlineEM} show that this update is asymptotically equivalent to the natural gradient update with three main improvements: (a) no matrix inversion is needed, (b) the algorithm may be accelerated through Polyak-Ruppert averaging~\citep{polyak1992acceleration}, i.e., using the average $\bar{\eta}_N$ of all $\eta_i$ instead of the last iterate $\eta_N$, and (c) the step-size is particularly simple to set, as we are taking \emph{convex} combinations of sufficient statistics, and hence only the decay rate of $\rho_i$ has to be chosen, i.e., of the form $\rho_i = i^{-\kappa}$, for $\kappa \in (0,1]$, without any multiplicative constant.

\subsubsection{Incremental View} For the specific stepsize $\rho_i= {1}/{i}$, the online EM algorithm (\ref{eqn:OEM}) corresponds exactly to the incremental EM presented above \citep{incrementalEM1998}, as then 
\[
s_{i}= \frac{1}{i}\sum_{j=1}^{i}\mathbb{E}_{p(h_j|X_{j},\eta_{j-1})}[S(X_{j},h_j)].
\]
See \citet{incrementalMM2014} for a detailed convergence analysis of incremental algorithms, in particular showing that step-sizes larger than $1/i$ are preferable (we observe this in practice in \mysec{results}).

\subsubsection{Monte Carlo Methods}
There exist alternative methods to the EM algorithm based on Monte Carlo sampling to compute the maximum likelihood. For instance, the Monte Carlo EM method (MCEM) \citep{wei1990monte} is a general Bayesian approach (i.e., $\eta$ is a random variable) to approximate the maximizer of the posterior distribution $p(\eta|X,h)$ with Monte Carlo sampling. More precisely, in the MCEM method, similarly to EM, a surrogate function of the log-likelihood is used, given by:
\[
Q(\eta,\eta_t) = \int_h \log[p(\eta|X,h)]p(h|X,\eta_t)\mathrm{d}h.
\]
The function $Q$ is approximated by sampling the latent variables $h$ from the current conditional $p(h|X,\eta_t)$:
\[
\hat{Q}(\eta,\eta_t) = \sum_{i=1}^P \log p(\eta|X,h^i),
\]
where $(h^i)_{i=1,\ldots,P}$ are the samples drawn from the conditional $p(h|X,\eta_t)$. The approximation $\hat{Q}$ is then maximize with respect to $\eta$. Note that this method is a batch method, namely, samples are drawn over all the dataset.

Other sequential Monte Carlo methods (SMC) use importance sampling to estimate the conditional distributions $p(h|X)$ using an auxiliary density function. This auxiliary distribution is then reweighted at each iteration. In our case where the conditional distribution $p(h|X)$ is intractable to compute, these methods fail to accurately approximate this distribution \citep{cappe2005inference,kantas2015particle} and our Gibbs sampling scheme is more relevant. The SMC methods are also more adapted to models with dependency between observations.

The two Monte Carlo methods mentioned above also consist in sufficient statistics updates for the class of models considered here.

\section{Online EM with Intractable Models}
\label{sec:intract}
The  online EM updates in (\ref{eqn:OEM})  lead to a scalable algorithm for optimization when the local E-step is tractable.
However, in many  latent variable models---e.g., LDA, hierarchical Dirichlet processes \citep{HDP}, or ICA \citep{ICA2004}---it is intractable to compute   the conditional expectation ${\mathbb{E}_{p(h|X,\eta)}[S(X,h)]}$.

Following \citet{OnlineEM_LDA}, we propose to leverage the scalability of online EM updates (\ref{eqn:OEM}) and locally approximate the conditional distribution $p(h|X,\eta)$ in the case this distribution is intractable to compute. We will however consider different approximate methods, namely  Gibbs sampling or variational inference. Our method is thus restricted to models where the hidden variable $h$ may naturally be splitted in two or more groups of simple random variables.
Our algorithm is described in Algorithm~\ref{alg:GoEM} and may be instantiated with two approximate inference schemes which we now describe.

\begin{algorithm}[H]
   \caption{Gibbs / Variational online EM}
   \label{alg:GoEM}
\begin{algorithmic}
   \STATE {\bfseries \underline{Input:}} $\eta_0$, $s_0$, $\kappa\in(0,1]$.
   \FOR{$i= 1,\ldots, N$ }
   \STATE $\bullet$ Collect observation $X_i$,
   \STATE $\bullet$ Estimate $p(h_i|X_{i},\eta_{i-1})$ with sampling (\texttt{G-OEM}) or variational inference (\texttt{V-OEM}),
   \STATE $\bullet$ Apply (\ref{eqn:OEM}) to sufficient statistics $s_{i}$ and parameter $\eta_{i}$ with $\rho_i=  {1}/{i^\kappa}$,
   \ENDFOR
   \STATE {\bfseries \underline{Output:}} $\bar{\eta}_N = \frac{1}{N} \sum_{i=1}^N \eta_i$ or $\eta_N$.
\end{algorithmic}
\end{algorithm}

\subsection{Variational Inference:  {\texttt{V-OEM}}}
\label{sec:VOEM}

While variational inference had been considered before for online estimation of latent variable models, in particular for LDA for incremental EM~\citep{SinglePassLDA}, using it for online EM (which is empirically faster) had not been proposed and allows to use bigger step-sizes (e.g., $\kappa=1/2$). These methods are based on maximizing the negative variational ``free-energy''
\begin{equation}
\label{eqn:free}
\mathbb{E}_{q(h|\eta)}\left[   \log \frac{ p(X,h|\eta)  }{  q(h|\eta) }\right],
\end{equation}
with respect to $q(h|\eta)$ having a certain factorized form adapted to the model at hand, so that efficient coordinate ascent may be used. See, e.g., \citet{SVI2013}. We now denote  online EM  with variational approximation of the conditional distribution $p(h|X,\eta)$ as \texttt{V-OEM}.

\subsection{Sampling Methods:  {\texttt{G-OEM}}}

MCMC methods to approximate the conditional distribution of latent variables with online EM have   been considered by
 \citet{OnlineEM_LDA}, who apply locally the  Metropolis-Hasting (M-H) algorithm \citep{MetropolisAlgo1953,MC1970}, and show results on simple synthetic datasets. 
While   Gibbs sampling is widely used for many models such as LDA due to its simplicity and lack of external parameters, M-H requires a proper proposal distribution with frequent acceptance and fast mixing, which may be hard to find in high dimensions.
We provide a different simpler local scheme  based on Gibbs sampling (thus adapted to a wide variety of models), and propose a thorough favorable comparison on synthetic and real datasets with existing methods.

The Gibbs sampler  is used to estimate posterior distributions by alternatively sampling parts of the variables given the other ones \citep [see][for details]{ExplainingGibbs}, and is standard and easy to use in many common latent variable models.
In the following, the online EM method with Gibbs estimation of the conditional distribution $p(h|X,\eta)$ is denoted \texttt{G-OEM}. 

As mentioned above, the online EM updates correspond to a stochastic approximation algorithm and thus are robust to random noise in the local E-step.  As a result, our sampling method is particularly adapted as it is a random estimate of the E-step---see a theoretical analysis by~\citet{OnlineEM_LDA}, and thus
we only need to compute a few Gibbs samples for the estimation of $p(h|X_{i},\eta_{i-1})$.
A key contribution of our paper is to reuse sampling techniques that have proved competitive in the batch set-up and to compare them to existing variational approaches. 

\subsection{``Boosted'' Inference}
\label{sec:boost}

 As the variational and MCMC estimations of $p(h|X_i,\eta_{i-1})$ are done with iterative methods, we can boost the inference of Algorithm~\ref{alg:GoEM} by applying the update in the parameter $\eta$ in~(\ref{eqn:OEM}) after each iteration of the estimation of $p(h|X_i,\eta_{i-1})$. In the context of LDA, this was proposed by~\citet{SinglePassLDA} for incremental EM and we extend it to all versions of online EM. 
{With this boost, we expect that the global parameters $\eta$ converge faster, as they are   updated more often. In the following, we denote by \texttt{G-OEM++} (resp. \texttt{V-OEM++}) the method \texttt{G-OEM} (resp. \texttt{V-OEM}) augmented with this boost.
}

 \subsection{Variational Bayesian Estimation}
  \label{sec:VB}
 In the Bayesian perspective where $\eta$ is seen as a random variable, we either consider a  distribution based on model averaging, e.g., 
 $r(X) = \int p(X|\eta) q(\eta) d\eta$
 where ${q(\eta) \propto \prod_{i=1}^N p(X_i | \eta) p(\eta)}$ is the posterior distribution, or
 \[
 r(X) = p(X|\bar{\eta}),
 \]
 where $\bar{\eta}$ is the summary (e.g., the mean) of the posterior distribution $q(\eta)$, or of an approximation, which is usually done in practice~\citep[see, e.g.,][]{hoffmanstructured} and is asymptotically equivalent when $N$ tends to infinity.
 
The main problem is that, \emph{even when the conditional distribution of   latent variables is tractable},  it is intractable to manipulate the \emph{joint} posterior distribution over the latent variables $h_1,\dots,h_N$, and the parameter $\eta$. Variational inference techniques consider an approximation where hidden variables are independent of the parameter $\eta$, i.e., such
 that 
\[
 p(\eta,h_1,\dots,h_N | X_1,\dots,X_N) \approx q(\eta) \prod_{i=1}^N q(h_i),
\]
which corresponds to the maximization of the following lower bound---called Evidence Lower BOund (ELBO)---on the log-likelihood $\log p(X_1,\dots,X_n)$~\citep{bishop2006pattern}:
\[
\int \!\!q(\eta) \prod_{i=1}^N\! q(h_i) \log \frac{p(\eta) \prod_{i=1}^n p(X_i,h_i|\eta)}{q(\eta) \prod_{i=1}^N q(h_i)} d \eta dh_1 \cdots dh_N.
\]
 The key insight from~\citet{OnlineLDA,StreamingVB} is to consider the variational distribution $q(\eta)$ as the global parameter, and the cost function above as a sum of \emph{local} functions that depend on the data $X_i$ and the variational distribution $q(h_i)$.  Once the local variational distribution $q(h_i)$ is maximized out, the sum structure may be leveraged in similar ways than for frequentist estimation, either by direct (natural) stochastic gradient~\citep{OnlineLDA} or incremental techniques that accumulate sufficient statistics~\citep{StreamingVB}. A nice feature of these techniques is that they extend directly to models with intractable latent variable inference, by making additional assumptions on $q(h_i)$ (see for example the LDA situation in \mysec{LDA}).
 
  In terms of actual updates, they are similar to online EM in Section~\ref{sec:VOEM}, with a few changes, but which turn out to lead to significant differences in practice.
  The similarity comes from the expansion of the ELBO as
$$ \mathbb{E}_{q(\eta)}
  \Big[{\sum_{i=1}^N}
  \mathbb{E}_{q(h_i)} \log \frac{p(X_i,h_i|\eta)}{ q(h_i)}
  \Big] + \mathbb{E}_{q(\eta)}\left[\log \frac{p(\eta)}{q(\eta)}\right].
$$
The left hand side has the same structure than the variational EM update in (\ref{eqn:free}), thus leading to similar updates, while the right hand side corresponds to the ``Bayesian layer'', and the maximization with respect to $q(\eta)$ is similar to the M-step of EM (where $\eta$ is seen as a parameter).

 Like online EM techniques presented in \mysec{intract}, approximate inference for latent variable is used, but, when using Bayesian stochastic variational inference techniques, there are two additional sources of inefficiencies: (a) extra assumptions regarding the independence of $\eta$ and $h_1,\dots,h_N$, and (b)~the lack of explicit formulation as the minimization of an expectation, which prevents the simple use of the most efficient stochastic approximation techniques (together with their guarantees). While (b) can simply slow down the algorithm, (a) may lead to results which are far away from exact inference, even for large numbers of samples (see examples in \mysec{results}).

Beyond variational inference, Gibbs sampling has been recently considered  by \citet{StreamingGibbs2016}: their method consists in sampling hidden variables for the current document given current parameters, but (a) only some of the new parameters are updated by incrementally aggregating the samples of the current document with current parameters, and (b) the method is slower than \texttt{G-OEM} (see \mysec{results}).

\section{Application to LDA}
\label{sec:LDA}

LDA \citep{LDA} is a probabilistic model that infers hidden topics given a text corpus where each document of the corpus can be represented as topic probabilities. In particular, the assumption behind LDA is that each document is generated from a mixture of topics and the model infers the hidden topics and the topic proportions of each document. 
In practice,  inference is done using Bayesian variational EM \citep{LDA}, Gibbs sampling \citep{GibbsLDA, wallach2006topic} or stochastic variational inference \citep{OnlineLDA, StreamingVB, SinglePassLDA}.

\subsubsection{Hierarchical Probabilistic Model.}
Let ${\mathcal{C}=\{X_1,\ldots,X_D\}}$ be a corpus of $D$ documents, $V$ the number of words in our vocabulary and $K$ the number of latent topics in the corpus. Each topic~$\beta^k$ corresponds to a discrete distribution on the $V$ words (that is an element of the simplex in $V$ dimensions). A hidden discrete distribution $\theta_i$ over the $K$ topics  (that is an element of the simplex in $K$ dimensions) is attached to each document $X_i$.
LDA is a generative model applied to a corpus of text documents which assumes that each word of the $i^{th}$ document $X_i$ is generated as follows:
\begin{itemize}
\item Choose $\theta_i\sim\mathrm{Dirichlet}(\alpha)$,
\item For each word $x_n\in X_i=\left(x_1,\ldots,x_{N_{X_i}}\right)$:
\begin{itemize}
\item[--] Choose a topic $z_n\sim\mathrm{Multinomial}(\theta_i)$,
\item[--]  Choose a word $x_n\sim\mathrm{Multinomial}(\beta^{z_n})$.
\end{itemize}
\end{itemize}
In our settings, an observation is a document ${X_i= (x_1,\ldots,x_{N_{X_i}} )}$ where for all ${1\leq n\leq N_{X_i}}$, ${x_n\in \{0,1\}^V}$ and ${\sum_{v=1}^Vx_{nv}=1}$. Each observation $X_i$ is associated with the hidden variables $h_i$, with ${h_i\equiv(Z_i=(z_1,\ldots,z_{N_{X_i}}),\theta_i)}$.  The vector $\theta_i$ represents the topic proportions of document $X_i$ and $Z_i$ is the vector of topic assignments of each word of $X_i$. The variable $h_i$ is local, i.e., attached to one observation $X_i$. The parameters of the model are global, represented by ${\eta\equiv(\beta,\alpha)}$, where $\beta$ represents the topic matrix and $\alpha$ represents the Dirichlet prior on topic proportions.

We derive the LDA model in Section~\ref{sec:LDAexp} to find $\phi$, $S$, $\psi$ and $a$ such that the joint probability $p(Z,\theta|X,\alpha,\beta)$ is in a non-canonical exponential family (\ref{eqn:exp_fam}). 

We may  then readily apply all algorithms from \mysec{intract} by estimating the conditional expectation $\mathbb{E}_{Z,\theta|X,\alpha,\beta}[S(X,Z,\theta)]$ with either variational inference (\texttt{V-OEM}) or Gibbs sampling (\texttt{G-OEM}). {See Sections~\ref{sec:Voem} and~\ref{sec:Goem} for online EM derivations}.  Note that the key difficulty of LDA is the presence of two interacting hidden variables $Z$ and $\theta$. 

\subsection{LDA and Exponential Families}
\label{sec:LDAexp}
An observation $X$ is a document of length $N_X$, where ${X=(x_1,\ldots,x_{N_X})}$, each word is represented by ${x_n\in \{0,1\}^V}$ with ${\sum_{v=1}^Vx_{nv}=1}$. Our corpus $\mathcal{C}$ is a set of $D$ observations ${\mathcal{C}=(X^1,\ldots,X^D)}$. For each document $X^i$  a hidden variable $\theta^i$ is associated, corresponding to the topic distribution of document $X^i$. For each word $x_n$ of document $X^i$ a hidden variable $z_n\in\{0,1\}^K$ is attached, corresponding to the topic assignment of word $x_n$. We want to find $\phi$, $S$, $\psi$ and $a$ such that, the joint probability is in the exponential family (\ref{eqn:exp_fam}):
\begin{align*}
p(X,Z,\theta|\beta,\alpha) = a(X,Z,\theta)\exp\left[\langle\phi(\beta,\alpha),S(X,Z,\theta)\rangle - \psi(\beta,\alpha) \right],
\end{align*}
given an observation $X$ and hidden variables $Z$ and $\theta$. For the LDA model, we have:

\begin{align*}
p(X,Z,\theta|\beta,\alpha) = & \prod_{n=1}^{N_X}p(x_n|z_n,\beta)p(z_n|\theta)p(\theta|\alpha)= \prod_{n=1}^{N_X} \prod_{k=1}^K\prod_{v=1}^V\left[(\beta^{k}_{v})^{x_{nv}} \theta_{k}\right]^{z_{nk}}p(\theta|\alpha),\\
\end{align*}
which we can expand as:
\begin{align*}
 p(X,Z,\theta|\beta,\alpha) &= \exp\left[ \sum_{n=1}^{N_X} \sum_{k=1}^K z_{nk} \log\theta_{k} \right] \times \exp \left[\sum_{n=1}^{N_X} \sum_{k=1}^K \sum_{v=1}^V x_{nv} z_{nk}\log\beta^{k}_{v}\right]\\
& \times \exp\left[ \sum_{k=1}^K (\alpha_k-1)\log\theta_k +  B(\alpha)\right],
\end{align*}
with $B(\alpha)=\log\left[\Gamma\left(\sum_{i=1}^K \alpha_i\right)\right]- \sum_{i=1}^K \log[\Gamma(\alpha_i)]$, where $\Gamma$ is the gamma function.
We deduce the non-canonical exponential family setting $\phi$, $S$, $\psi$ $a$:
\begin{eqnarray}
S(X,Z,\theta) & = & \left(\renewcommand*{\arraystretch}{1.8}
\begin{array}{c}
S^1_{kv}\equiv \left[\sum\limits_{n=1}^{N_X} z_{nk} x_{nv}\right]_{kv}\\
S^2_{k}\equiv \left[\log \theta_k\right]_k
\end{array}\right),\label{eqn:expS2}\\
\phi(\beta,\alpha) & = &\left(\renewcommand*{\arraystretch}{1.5}
\begin{array}{l}
\phi^1_{kv} \equiv \left[\log\beta^k_{v}\right]_{kv}\\
\phi^2_{k} \equiv \left[\alpha_{k}\right]_{k}\\
\end{array}
\right),\label{eqn:expPhi2}
\end{eqnarray}
with $S^1,\phi^1\in\mathbb{R}^{K\times V}$ and $S^2,\phi^2\in\mathbb{R}^{K}$,
\begin{equation}
\psi(\beta,\alpha)=\sum_{i=1}^K \log[\Gamma(\alpha_i)] - \log\left[\Gamma\left(\sum_{i=1}^K \alpha_i\right)\right],
\label{eqn:expPsi2}
\end{equation}
and
\[
a(X,Z,\theta) = \exp\left[\sum_{k=1}^K \left(\sum_{n=1}^{N_X}z_{nk}-1\right) \log\theta_{k}\right].
\]
The one-to one mapping between the sufficient statistics $s=\binom{s^1}{s^2}$ and $(\beta,\alpha)$ is defined by:
\[
(\beta,\alpha)^*[s] = \left\{\renewcommand*{\arraystretch}{1.6}
\begin{array}{cl}
\arg\max_{\beta\geq 0,\alpha\geq 0} & \langle\phi(\beta,\alpha), s\rangle - \psi(\beta,\alpha)\\
\text{s.t.} & \beta^\top\mathbf{1} =\mathbf{1},
\end{array}
\right.
\]
where $\mathbf{1}$ denotes the vector whose all entries equal 1.
The objective ${\langle\phi(\beta,\alpha), s\rangle - \psi(\beta,\alpha)}$ is concave in $\beta$ from the concavity of $\log$ and concave in any $\alpha_k$ for $\alpha\geq 0$ as the function $B(\alpha)$ is concave as the negative log-partition of the Dirichlet distribution.
We use the Lagrangian method for $\beta$:
\[
L(\beta,\lambda) = \sum_{k=1}^K\sum_{v=1}^V s^1_{kv}\log\beta^k_{v} + \lambda^\top(\beta^\top\mathbf{1}-\mathbf{1}),
\]
with $\lambda\in\mathbb{R}^K$. The derivative of $L$ is set to zero when:
\[
\forall (k,v), \frac{s^1_{kv}}{\beta^k_{v}} + \lambda_k = 0 \Rightarrow\lambda_k = -\sum_{v=1}^Vs^1_{kv},
\]
as $\sum_{v=1}^V\beta^k_{v}=1$. We then have ${\left(\beta^*(s)\right)_{kv} = \frac{s^1_{kv}}{\sum_j s^1_{kj}}}$. This mapping satisfies the constraint ${\beta\geq 0}$ because for any observation $X$ and hidden variable $Z$, we have ${S^1(X,Z)_{kv}\geq 0}$. This comes from~(\ref{eqn:expS2}) and  the fact that ${\forall (n,k,v), (x_{nv},z_{nk})\in\{0,1\}^2}$.
We find the condition on $\alpha$ by setting the derivatives to 0, which gives ${\forall k\in\llbracket 1,K\rrbracket}$: 
\[ 
s^2_k- \Psi([\alpha^*(s)]_k) + \Psi\left(\sum_{i=1}^K[\alpha^*(s)]_i\right) =0,
\] 
where ${\Psi:x\mapsto\frac{\partial}{\partial x}[\log\Gamma] (x)}$ is the digamma function.
Finally, $(\alpha^*(s),\beta^*(s))$ satisfies $\forall (k,v)$:
\begin{equation}
\label{eqn:MstepLDA}
\left\{\renewcommand*{\arraystretch}{1.8}
\begin{array}{ccl}
\left(\beta^*(s)\right)_{kv} & \equiv & \left[\frac{s^1_{kv}}{\sum_j s^1_{kj}}\right]_{kv}\\
\Psi([\alpha^*(s)]_k) - \Psi\left(\sum\limits_{i=1}^K[\alpha^*(s)]_i\right) & = & s^2_k .
\end{array}
\right.
\end{equation}
The parameter $\alpha^*$ is usually estimated with gradient ascent \citep{LDA,OnlineLDA}. We can also estimate $\alpha$ with the fixed point iteration \citep{EstimateDirichlet} which consists in repeating the following update until convergence:
\[
\textstyle \alpha^{new}_k = \Psi^{-1}\left( \Psi\left(\sum_{i=1}^K\alpha^{old}_i\right)  +s^2_k\right).
\]
We use the fixed point iteration to estimate $\alpha^*$ as it is more stable in practice. We study different updates for $\alpha$ in Appendix~\ref{app:alpha}.

We can now apply Algorithm \ref{alg:GoEM} to LDA. The only missing step is the estimation of the conditional expectation $\mathbb{E}_{Z,\theta|X,\alpha_t,\beta_t}[S(X,Z,\theta)]$, with ${X=(x_1,\ldots,x_{N_X})}$ and ${Z=(z_1,\ldots,z_{N_X})}$. We explain how to approximate this expectation with variational inference and Gibbs sampling.

\subsection{Variational Online EM Applied to LDA (\texttt{V-OEM})}
\label{sec:Voem}
In this section we explain how to approximate $\mathbb{E}_{Z,\theta|X,\alpha_t,\beta_t}[S(X,Z,\theta)]$ with variational inference, in the frequentist setting. See \citet{SVI2013} for detailed derivations of variational inference for LDA in the Bayesian setting (from which the updates in the frequentist setting may be easily obtained). The idea behind variational inference is to maximize the Evidence Lower BOund (ELBO), a lower bound on the probability of the observations: 
\[
{p(X)\geq \mathrm{ELBO}(X,p,q)},
\]
where $q$ represents the variational model. In the case of LDA, the variational model is often set with a Dirichlet($\gamma$) prior on $\theta$ and  a multinomial prior on $Z$ \citep{SVI2013}:
\begin{equation}
q(Z,\theta) = q(\theta|\gamma)\prod_{n=1}^{N_X}q(z_n|\zeta_n).
\label{eqn:meanfield2}
\end{equation}
We then maximize the ELBO with respect to $\gamma$ and $\zeta$, which is equivalent to minimizing the Kullback-Leibler (KL) divergence between the variational posterior and the true posterior: 
\begin{equation}
\max_{\gamma,\zeta}\;\mathrm{ELBO}(X,p,q)\Leftrightarrow \min_{\gamma,\zeta} \;\mathrm{KL}[p(Z,\theta|X)||q(\theta,Z)].
\label{eqn:ELBO2}
\end{equation}
We solve this problem with block coordinate descent, which leads to iteratively updating $\gamma$ and $\zeta$ as follows:
\begin{align}
\label{eqn:varphi2}\zeta_{nk} &\propto \prod_{v=1}^V \left(\beta^k_{v}\right)^{x_{nv}} \exp\left[ \Psi(\gamma_k) \right],\\
\label{eqn:vargam2}\gamma_k &= \textstyle \alpha_k + \sum_{n=1}^{N_X}\zeta_{nk}.
\end{align}
We then approximate $\mathbb{E}_{Z,\theta|X,\alpha_t,\beta_t}[S(X,Z,\theta)]$ with the variational posterior.
Given (\ref{eqn:expS2}) and (\ref{eqn:meanfield2}), we have:
\begin{align}
\mathbb{E}_{p(Z,\theta|X)}[S(X,Z,\theta)] &\approx\mathbb{E}_{q(Z,\theta)}[S(X,Z,\theta)]
\label{eqn:varexp2}=\left[\begin{array}{c}
\left(\textstyle\sum_{n=1}^{N_{X_{t+1}}} \zeta_{nk} x_{nv}\right)_{kv}\\\\
\left(\Psi(\gamma_k) - \Psi\left(\textstyle\sum_{j=1}^K \gamma_j\right) \right)_k
\end{array}
\right].
\end{align}
The variational approximation of $\mathbb{E}_{p(Z,\theta|X)}[S(X,Z,\theta)]$ is then done in two steps:
\begin{enumerate}
\item Iteratively update $\zeta$ with (\ref{eqn:varphi2}) and $\gamma$ with (\ref{eqn:vargam2}),
\item $\mathbb{E}_{p(Z,\theta|X)}[S(X,Z,\theta)]\leftarrow \mathbb{E}_{q(Z,\theta|\gamma,\zeta)}[S(X,Z,\theta)]$ with equation (\ref{eqn:varexp2}).
\end{enumerate}
As $\gamma$ and $\zeta$ are set to minimize the distance between the variational posterior and the true posterior~(\ref{eqn:ELBO2}) we expect that this approximation is close to the true expectation. However, as the variational model is a simplified version of the true model, there always remains a gap between the true posterior and the variational posterior.

\subsection{Gibbs Online EM Applied to LDA (\texttt{G-OEM})}
\label{sec:Goem}
In this section we explain how to approximate $\mathbb{E}_{Z,\theta|X,\alpha_t,\beta_t}[S(X,Z,\theta)]$ with Gibbs sampling.
\subsubsection{Expectation of $S^1$.}
Given (\ref{eqn:expS2}), we have ${\forall k\in\llbracket 1,K\rrbracket}$, ${\forall v\in\llbracket 1,V\rrbracket}$:
\begin{align*}
\mathbb{E}_{Z,\theta|X,\alpha,\beta}\left[(S^1(X,Z))_{kv}\right]  & =  \mathbb{E}_{Z,\theta|X,\alpha,\beta}\left[\sum_{n=1}^{N_X}z_{nk} x_{nv}\right]\\
&= \sum\limits_{n=1}^{N_X} \int_{Z,\theta} z_{nk}x_{nv}p(z_{n},\theta|X,\beta,\alpha)\mathrm{d}\theta\mathrm{d}z = \sum\limits_{n=1}^{N_X} x_{nv} p(z_{nk}=1|X,\beta,\alpha).
\end{align*}
We see that we only need the probability of $z$, and can thus use collapsed Gibbs sampling~\citep{GibbsLDA}.
We have, following Bayes rule:
\begin{align*}
p(z_{nk}=1&|z_{-n},X,\beta,\alpha) \propto p(x_{n}|z_{nk}=1,\beta)p(z_{nk}=1|z_{-n},\alpha),
\end{align*}
where $z_{-n}$ is the topic assignments except index $n$. In the LDA model, each word $x_n$ is drawn from a multinomial with parameter $\beta^{z_n}$, which gives:
\[
{p(x_{n}|z_{nk}=1,\beta)=\sum_{v=1}^V x_{nv}\beta^k_{v}}.
\] 
In the following, we use the notation ${\beta^k_{x_n }\equiv \sum_{v=1}^V x_{nv}\beta^k_{v}}$ for the sake of simplicty.  We then use the fact that the topic proportions $\theta$ has a ${\mathrm{Dirichlet}(\alpha)}$ prior, which implies that ${Z|\alpha}$ follows a Dirichlet-multinomial distribution (or multivariate P\'olya distribution). As a result, the conditional distribution is:
\[
p(z_{nk}=1|z_{-n},\alpha) = \frac{N_{-n,k} +\alpha_k}{ (N_{X}-1) +\sum_j\alpha_j},
\]
with $N_{-n,k}$ the number of words assigned to topic $k$ in the current document, except index $n$.
Finally, we have the following relation \citep{GibbsLDA}:
\begin{equation}
p(z_{nk}=1|z_{-n},X,\beta,\alpha) \propto \beta^k_{x_n}\times \frac{N_{-n,k} +\alpha_k}{ (N_{X}-1) +\sum_j\alpha_j}.
\label{eqn:Gibbs}
\end{equation}
We estimate $p(z_{nk}=1|X,\beta,\alpha)$ with Gibbs sampling by iteratively sampling topic assignments $z_n$ for each word, as detailed in Algorithm~\ref{alg:Gibbs}. We average over the last quarter of samples to reduce noise in the final output. We then incorporate the output in Algorithm~\ref{alg:GoEM}.

\subsubsection{Expectation of $S^2$.}
Given (\ref{eqn:expS2}), we also have ${\forall k\in\llbracket 1,K\rrbracket}$, ${\forall v\in\llbracket 1,V\rrbracket}$:
\begin{align*}
\mathbb{E}_{Z,\theta|X,\alpha,\beta}[(S^2(X,Z))_k]  &=  \mathbb{E}_{Z,\theta|X,\alpha,\beta}[\log\theta_k].
\end{align*}
On the one hand, we have:
\begin{align*}
p(Z,\theta|X,\beta,\alpha) &=p(Z|\theta,X,\beta,\alpha)p(\theta|X,\beta,\alpha)=C(\alpha)\prod_{k=1}^K \theta_k^{ \left(\sum_{n=1}^{N_X} z_{nk}\right) +\alpha_k -1},
\end{align*}
with $C(\alpha)=\frac{\Gamma\left(\sum_{i=1}^K\alpha_i\right)}{\prod_{i=1}^K\Gamma(\alpha_i)}$.
On the other hand:
\begin{align*}
p(Z,\theta|X,\beta,\alpha) &\propto p(\theta|Z,\alpha)p(Z|X,\beta).
\end{align*}
We deduce from the two identities:
\begin{align*}
p(\theta|Z,\alpha)\propto \prod_{k=1}^K \theta_k^{ \left(\sum_{n=1}^{N_X} z_{nk}\right) +\alpha_k -1} \Rightarrow \theta|Z,\alpha \sim \mathrm{Dirichlet}\left( \textstyle\alpha + \sum_{n=1}^{N_X} z_{n} \right).
\end{align*}
Finally, the expectation is:
\begin{align*}
\mathbb{E}_{Z,\theta|X,\alpha,\beta}[(S^2(X,Z))_k]  = & \mathbb{E}_{Z,\theta|X,\alpha,\beta}[\log\theta_k]\\
=&\mathbb{E}_{Z|X,\beta,\alpha}\left[\mathbb{E}_{\theta|Z,\alpha}\left[\log\theta_k\right]\right]\\
=& \mathbb{E}_{Z|X,\beta,\alpha}\left[\Psi\left([\alpha(s)]_k +\textstyle \sum_{n=1}^{N_X}z_{nk}\right)\right] - \Psi\left(\textstyle \sum_{i=1}^K[\alpha(s)]_i + N_X\right),
\end{align*}
as the distribution of $\theta|Z,\alpha$ is $\mathrm{Dirichlet}\left(\alpha+\sum\limits_{n=1}^{N_X} z_{n} \right)$. We use the values of $z$ sampled with Algorithm~\ref{alg:Gibbs} to estimate this expectation. More precisely, keeping notations of Algorithm~\ref{alg:Gibbs}:
\begin{align*}
\mathbb{E}_{Z|X,\beta,\alpha} & \left[\Psi\left([\alpha(s)]_k + \textstyle\sum_{n=1}^{N_X}z_{nk}\right)\right] \approx \frac{1}{P}\sum_{t=1}^P \Psi\left([\alpha(s)]_k + \textstyle\sum_{n=1}^{N_X}z^t_{nk}\right).
\end{align*}

\begin{algorithm}[t]
   \caption{Gibbs sampling scheme to approximate $p(z_{nk}=1|X,\beta,\alpha)$}
   \label{alg:Gibbs}
\begin{algorithmic}
   \STATE {\bfseries \underline{Input:}} $\beta$, $\alpha$, $X$.
   \STATE {\bfseries \underline{Initialization:}} ${Z^0_n\sim\mathrm{Mult}([\bar{\beta}^k_{x_n}]_{k=1,\ldots,K})}$, with ${\bar{\beta}^k_{x_n}=\frac{\beta^k_{x_n}}{\sum_j \beta^j_{x_n}}}$ ${\forall n\in\llbracket 1,N_X\rrbracket}$.
   \FOR{$t=1,2,\ldots, P$ }
   \STATE Compute random permutation $\sigma^t$ on $\llbracket 1,N_X\rrbracket$,
   \FOR{$n=0,1,\ldots, N_X$ }
   \STATE $\bullet$ Set ${Z^t_{-n} = \left\{(z_{\sigma^t(i)}^t)_{1\leq i<n},(z_{\sigma^t(i)}^{t-1})_{n< i\leq N_X}\right\}}$,
   \STATE $\bullet$ Compute $\forall k$, ${p(z_{\sigma^t(n)k}=1|Z^t_{-n},X,\beta,\alpha)}$ with Equation~(\ref{eqn:Gibbs}),
   \STATE $\bullet$ Sample ${z_{\sigma^t(n)}^t\sim \mathrm{Mult}\left[ p(z_{\sigma^t(n)}|Z^t_{-n},X,\beta,\alpha) \right]}$,
   \ENDFOR
   \ENDFOR
   \FOR{$n=0,1,\ldots, N_X$ }
   \STATE Set ${Z^t_{-n} = \left\{(z_{i}^t)_{1\leq i<n},(z_{i}^{t})_{n< i\leq N_X}\right\}}$ for $t\geq\frac{3}{4}P$,
   \STATE \begin{small}
   $p(z_{nk}=1|X,\beta,\alpha)\leftarrow \frac{4}{P}\sum\limits_{t=\frac{3}{4}P}^Pp(z_{nk}=1|Z^t _{-n},X,\beta,\alpha)$
   \end{small}
   \ENDFOR
   \STATE {\bfseries \underline{Output:}} $\forall k$, $\forall n$: $(z_n^t)_{t=1,\ldots,P}$, $p(z_{nk}=1|X,\beta,\alpha)$.
\end{algorithmic}
\end{algorithm}
 
\subsection{Bayesian Approach}
\label{sec:bayes}
In a Bayesian setting, we consider $\beta$ as a random variable, with $\beta\sim\mathrm{Dirichlet}(b\mathbf{1})$, with $b\in\mathbb{R}$ and $\mathbf{1}\in\mathbb{R}^V$ denotes the vector whose all entries equal $1$. The variational distribution of the global parameter $\beta$ is then set to $q(\beta^k|\lambda^k)=\mathrm{Dirichlet}(\lambda^k)$, with $\lambda^k\in\mathbb{R}^{V}$ $\forall k=1,\ldots,K$. The main difference with the frequentist methods above (\texttt{G-OEM} and \texttt{V-OEM}) is to optimize the ELBO with respect to the variational parameters $(\lambda^k)_k$. In practice, it is equivalent to replace $\beta^k_v$ by $\exp\left[\mathbb{E}_q[\log\beta^k_v]\right]$ in all the updates above (i.e., in Equation~(\ref{eqn:varphi2}) for \texttt{V-OEM} and in Equation~(\ref{eqn:Gibbs}) for \texttt{G-OEM}). The variational parmater $\lambda^k\in\mathbb{R}^{V}$ is updated with stochastic gradient on the ELBO, which gives, at iteration $t$:
\begin{equation}
\lambda^k(t+1) = \rho_t\lambda^k(t) + (1-\rho_t)\hat{\lambda}^k,
\label{eqn:bayes}
\end{equation}

with $\hat{\lambda}^k\in\mathbb{R}^{ V}$, $\hat{\lambda}^k_{v}= b + D \mathbb{E}_q\left[ S^1_{kv}\right]$, where $D$ is the total number of documents in the dataset and $b$ is the prior on $\beta^k$ \citep{SVI2013}.

\section{Application to Hierarchical Dirichlet Process (HDP) \citep{HDP}}
The HDP model is a generative process to model documents from an infinite set of topics $\beta_k$, $k=1,2,3,\ldots$. Each topic is a discrete distribution of size $V$, the size of the vocabulary. Each topic is associated to a weight $\pi_k\in[0,1]$, representing the importance of the topic in the corpus. For each document $d$, the (infinite) topic proportions $\nu_d$ are drawn from $\nu_d\sim\mathrm{Dirichlet}(b\pi)$. We then generate words with a similar scheme to LDA scheme. More formally a corpus is generated as follows:
\begin{enumerate}
\item Draw an infinite number of topics $\beta_k\sim\mathrm{Dirichlet}(\eta)$, for $k\in\{1,2,3,\ldots\}$;
\item Draw corpus breaking proportions $\bar{\pi}_k\sim\mathrm{Beta}(1,\alpha)$, for $k=1,2,3,\ldots$; with ${\pi_k=\sigma_k(\bar{\pi})}$;
\item For each document $d$:
\begin{enumerate}
\item Draw document-level topic proportions: $\nu_{d}\sim\mathrm{Dirichlet}(b\pi)$;
\item For each word $n$ in $d$:
\begin{enumerate}
\item Draw topic assignment $z_{dn}\sim\mathrm{Multinomial}(\nu_d)$;
\item Draw word $w_n\sim \mathrm{Multinomial}(\beta_{z_{dn}})$.
\end{enumerate}
\end{enumerate}
\end{enumerate}
In practice, we set the initial number of topics to $T=2$. We then increase the number of topics used in the corpus using Gibbs sampling and $p(z_{dn}>T|X,\eta)\propto b(1-\sum_{i=1}^T\pi_i)$. See Section~\ref{sec:HDPinf} for details.

\subsection{HDP and Exponential Families}
We consider an exponential family model on random variables $(X,h)$ with parameter $\eta\in\mathcal{E}\subseteq\mathbb{R}^d$ and with density:
\[
p(X,h|\eta) = a(X,h)\exp\left[\langle\phi(\eta),S(X,h)\rangle - \Psi(\eta)\right].
\]
In the case of HDP, an observation $X$ is a document of length $N_X$, where $X=(x_1,\ldots,x_{N_X})$, $x_n\in\{0,1\}^V$ and $\sum_{v=1}^V x_{nv}=1$. In the frequentist approach, the parameters of the model are global, represented by $\eta\equiv(\beta,\pi)$, where $\beta$ represents the corpus topics, $\pi$ represents the corpus breaking proportions. Our corpus $\mathcal{C}$ is a set of $D$ observations $\mathcal{C}=(X^1,\ldots,X^D)$. For each document $X^d$, the associated hidden variables are $\nu_{d}\in [0,1]^{K}$ corresponding to document-level topic proportions. For each word $x_n$ of document $X^d$, a hidden variable $z_n\in\{0,1\}^T$ is attached, corresponding  to the topic assignment of word $x_n$. 

We want to find $\phi$, $S$, $\psi$ and $a$ such that, the joint probability is in the exponential family:
\begin{align*}
p(X,Z,\nu|\beta,\pi) = a(X,Z,\nu)\exp\left[\langle\phi(\beta,\pi),S(X,Z,\nu)\rangle - \psi(\beta,\pi) \right],
\end{align*}
given an observation $X$ and hidden variables $Z$ and $\nu$. For the HDP model, we have:
\begin{align*}
p(X,Z,\nu|\beta,\pi) = & p(\nu|\pi)\prod_{n=1}^{N_X} p(x_n|z_n,\beta)p(z_n|\nu)\\
= & W(\pi)\prod_{k\in\mathbb{N}^*}\left(\nu_k\right)^{b\pi_k - 1}\prod_{n=1}^{N_X} \prod_{k}\left(\nu_k\right)^{z_{nk}}\prod_{v}\left(\beta_{k,v}\right)^{x_{nv}z_{nk}}\\
= & \exp\left[-\psi(\pi)\right]\exp\left[\sum_k \log \nu_k \left(\sum_{n=1}^{N_X} z_{nk} -1\right) \right]\\
& \times \exp\left[ \sum_{k}(b\pi_k)\log\nu_k  \right] \\
& \times \exp\left[ \sum_{k,v}\log\beta_{k,v}\sum_{n=1}^{N_X}x_{nv}z_{nk} \right],
\end{align*}
with $\psi(\pi) = \sum_{k}\log\Gamma(b\pi_k) - \log\Gamma(b)$.We deduce the exponential family setting $\phi,S,a$:
\begin{eqnarray}
S(X,Z,\nu) & = & \left(\renewcommand*{\arraystretch}{1.8}
\begin{array}{c}
S^1_{k}\equiv \left[ \log\nu_{k}\right]_{k}\\
S^2_{kv}\equiv \left[\sum\limits_{n=1}^{N_X} z_{nk} x_{nv}\right]_{kv}\\

\end{array}\right),\label{eqn:HDPexpS2}\\
\phi(\beta,\pi) & = &\left(\renewcommand*{\arraystretch}{1.5}
\begin{array}{l}
\phi^1_{k} \equiv \left[b\pi_k\right]_{k}\\
\phi^2_{kv} \equiv \left[\log\beta_{k,v}\right]_{kv}\\
\end{array}
\right),\label{eqn:HDPexpPhi2}
\end{eqnarray}
with
\[
a(X,Z,\nu) = \exp\left[\sum_k \log \nu_k \left(\sum_{n=1}^{N_X} z_{nk} -1\right) \right].
\]
The one-to one mapping between the sufficient statistics $s=(s^1,s^2)^\top$ and $(\beta,\pi)$ is defined by:
\[
(\beta,\pi)^*[s] = \left\{\renewcommand*{\arraystretch}{1.6}
\begin{array}{cl}
\arg\max_{\beta\geq 0,1\geq\pi\geq 0} & \langle\phi(\beta,\pi), s\rangle - \psi(\beta,\pi)\\
\text{s.t.} & \beta^\top\mathbf{1} =\mathbf{1},
\end{array}
\right.
\]
where $\mathbf{1}$ denotes the vector whose all entries equal 1.

With the same computation than LDA, $\beta^*(s)_{kv} \equiv \left[\frac{s^2_{kv}}{\sum_j s^2_{kj}}\right]$. We find $\pi^*(s)$ by solving:
\begin{align*}
\pi^*(s) & = \arg\max_{1\geq\pi\geq 0} \sum_{k=1}^K \left(b\pi_k s^1_k - \log\Gamma(b\pi_k)\right) + \log\Gamma(b\sum_k\pi_k),
\end{align*}
which gives:
\[
\Psi(b\pi^*(s)_k) - \Psi\left(b\sum_i\pi^*(s)_i\right) = s^1_k.
\]
where $\Psi:x\mapsto\frac{\partial}{\partial x}\left[\log\Gamma\right](x)$ is the digamma function. We estimate $(b\pi)^*$ with the fixed point iteration which consists in repeating the following update until convergence:
\[
\left(b\pi\right)^{new}_k = \Psi^{-1}\left(\Psi\left( \sum_i (b\pi_i)^{old}\right)+s^1_k\right).
\]

Finally, $(\beta,\pi)^*[s]$ satisfies $\forall(k,v)$:
\[\boxed{
\left\{\renewcommand*{\arraystretch}{1.6}
\begin{array}{ccl}
(\beta^*(s))_{kv} & = & \left[\frac{s^2_{kv}}{\sum_j s^2_{kj}}\right]\\
\Psi(b\pi^*(s)_k) - \Psi\left(b\sum_i\pi^*(s)_i\right) & = & s^1_k.
\end{array}
\right.}
\]

\subsection{Inference with Online EM}
\label{sec:HDPinf}
In this section, we explain how to approximate $\mathbb{E}_{Z,\nu|X,\eta}[S(X,Z,\nu)]$ with Gibbs sampling from a frequentist and a Bayesian perspective. In particular, as the total number of topics is infinite, we need to keep track of the previously used topics and iteratively extend the number of topics considered.
\subsubsection{Gibbs Online EM (\texttt{G-OEM})}
In our frequentist \texttt{G-OEM} approach, $\eta$ is a parameter. The Gibbs sampling scheme to approximate $\mathbb{E}_{Z,\nu|X,\eta}[S(X,Z,\nu)]$ is different from LDA and a probability of adding a new topic to the current list is computed at each iteration, as explained below.

\subsubsection{Expectation of $S^1$.}
We have:
\begin{align*}
\mathbb{E}_{Z,\nu|X,\eta}[S^1(X,Z,\nu)]_k & = \mathbb{E}_{Z,\nu|X,\eta}\left[\log\nu_k\right]\\
& = \mathbb{E}_{Z|X,\eta}\left[\Psi\left(b\pi_k+\sum_{n=1}^{N_X}z_{nk}\right)\right] - \Psi\left(b\sum_i \pi_i+N_X\right),
\end{align*}
and we use the values of $z$ sampled with Gibbs sampling to compute:
\[
\mathbb{E}_{Z|X,\eta}\left[\Psi\left(b\pi_k+\sum_{n=1}^{N_X}z_{nk}\right)\right]\approx\frac{1}{P}\sum_{t=1}^P\Psi\left(b\pi_k + \sum_{n=1}^{N_X} z_{nk}^t \right)
\]

\subsubsection{Expectation of $S^2$.}
We have:
\[
\mathbb{E}_{Z,\nu|X,\eta}[S^2(X,\nu)]_{kv} = \mathbb{E}_{Z,\nu|X,\eta}\left[\sum_{n=1}^{N_X}z_{nk}x_{nv}\right] = \sum_{n=1}^{N_X}x_{nv}p(z_{nk}=1|X,\eta)
\]

\subsubsection{Sampling $z|X,\eta$.}
If $T$ is the current number of topics, we have:
\begin{align*}
\forall k\in\{1,\ldots,T\},\;\;p(z_{nk}=1|z^{-n},X,\eta) & \propto (N_{k}^{-n}+b\pi_k) \times p(x_{n}|z_{ni}=1,c_{ik}=1,\eta)\\
&\propto (N_{k}^{-n}+b\pi_k) \times \beta_{k,x_n},
\end{align*}
and the probability of sampling a new topic is given by:
\[
p(z_n>T|z^{-n},X,\eta) \propto b\left(1-\sum_{t=1}^T\pi_k\right)/V.
\]
When a new topic is generated, we initialize the probability $\pi_{T+1}$ with $\bar{\pi}_{T+1}\sim\mathrm{Beta}(1,\alpha)$ and ${\pi_{T+1}=\bar{\pi}_{T+1}\prod_{t=1}^T\left(1-\bar{\pi}_t\right)}$.

\subsubsection{Bayesian Approach: \texttt{VarGibbs} \citep{BNP_HDP}}
In a Bayesian settings where $\beta^k\sim\mathrm{Dirichlet}(\eta)$; $q(\beta^k|\lambda)=\mathrm{Dirichlet}(\lambda^k)$ and $\pi_k\sim \mathrm{Beta}(1,a)$; $q(\pi_k|a_k,b_k)=\mathrm{Beta}(a_k,b_k)$, the sampling scheme is different as we also sample $\pi$ and an auxiliary variable $s_{dk}$ corresponding to the number of ``tables'' serving ``dish'' $k$ in ``restaurant'' $d$ (in the fomulation of HDP as a Chinese restaurant process; see \citet{BNP_HDP} for details).\\
\underline{Sampling $z$:}
\[
p(z_{nk}=1|z^{-n},\lambda,\pi) \propto (N_{dk}^{-n}+b\pi_k)\frac{N_{kx_n}^{-n}+\lambda_{kx_n}}{N_{k}^{-n} + \sum_v\lambda_{kv}}.
\]
\underline{Sampling $s$:}
\[
p(s_{dk}|N_{dk},b\pi_k) = \frac{\Gamma(b\pi_k)}{\Gamma(b\pi_k+N_{dk})}S(N_{dk},s_{dk})\left(b\pi_k\right)^{s_{dk}},
\]
with $S(n,m)$ are unsigned Stirling number of the first kind.\\
\underline{Sampling $\pi$:}
\[
p(\bar{\pi}_k)\propto \bar{\pi}_k^{a_k-1+\sum_{d\in\mathbb{S}}s_{dk}}\left(1-\bar{\pi}_k\right)^{b_k-\alpha + \sum_{d\in\mathbb{S}}\sum_{j=k+1}^\infty s_{dj}}
\]
We then set:

\begin{equation}
\begin{cases}
\hat{\lambda}_{kv} = \eta + D\sum\limits_{n=1}^{N_X}z_{nk}x_{nv}\\[0.5cm]
\hat{a}_k = 1+D s_{dk}\\[0.5cm]
\hat{b}_k  =  \alpha+D\sum\limits_{j=k+1}^\infty s_{dj}
\end{cases}
\label{eqn:hdpbayes}
\end{equation}
and $\lambda^{t+1}=(1-\rho_t)\lambda^t + \rho_t\hat{\lambda}$; $a^{t+1}=(1-\rho_t)a^t + \rho_t\hat{a}$; $b^{t+1}=(1-\rho_t)b^t + \rho_t\hat{b}$.

In practice, for each document we sample the hidden variables $z$ for each word and compute the topic counts $N_{dk}$ for topic $k$ in document $d$, then we sample the variable $s$. Finally, we perform the online EM algorithm by making the approximation ${\mathbb{E}_{p(h|X,\eta)}[S(X,h)]\approx\mathbb{E}_{q(h)}[S(X,h)]}$, which corresponds to equation~\eqref{eqn:hdpbayes}. Note that in this Bayesian approach, the parameters ${(\lambda,a,b)}$ represent the distribution parameters of the random variables $\beta$ and $\pi$.

\section{Evaluation}
\label{sec:results}
We evaluate our method by computing the likelihood on held-out documents, that is $p(X|\beta,\alpha)$ for any test document $X$. For LDA, the likelihood is intractable to compute. We approximate $p(X|\beta,\alpha)$ with the ``left-to-right'' evaluation algorithm \citep{wallachEvaluation} applied to each test document.
This algorithm is a mix of particle filtering and Gibbs sampling. On any experiments, this leads essentially to the same log-likelihood than Gibbs sampling with sufficiently enough samples---e.g., 200.  
In the following, we present results in terms of log-perplexity, defined as the opposite of the log-likelihood $  -\log p(X|\eta)$. The lower the log-perplexity, the better the corresponding model. In our experiments, we  compute the average test log-perplexity on  $N_t$  documents.
We compare eight different methods: 

\begin{itemize}
\item \texttt{G-OEM} (our main algorithm): Gibbs online EM. {Online EM algorithm with Gibbs estimation of the conditional distribution $p(h|X,\eta)$ (Algorithm~\ref{alg:Gibbs}). } Frequentist approach and step-size {$\rho_i= {1}/{\sqrt{i}}$};
\item \texttt{V-OEM++}: variational online EM (also a new algorithm). {Online EM algorithm with variational estimation of the conditional distribution $p(h|X,\eta)$, augmented with inference boosting from Section~\ref{sec:boost}. Frequentist approach and step-size $\rho_i= {1}/{\sqrt{i}}$;}
\item \texttt{OLDA}: online LDA \citep{OnlineLDA}. Bayesian approach which maximizes the ELBO from Section~\ref{sec:VB}, with natural stochastic gradient ascent { and a step-size $\rho_i= {1}/{\sqrt{i}}$};
\item \texttt{VarGibbs}: Sparse stochastic inference for LDA \citep{mimno2012sparse}. This method also maximizes the ELBO but estimates the variational expectations ${q(Z,\theta)}$ with Gibbs sampling instead of iterative maximization of variational parameters---see Section~\ref{sec:Voem};
\item \texttt{SVB}: streaming variational Bayes \citep{StreamingVB}. A variational Bayesian equivalent of \texttt{V-OEM} with  step-size $\rho_i= {1}/{ {i}}$;
\item \texttt{SPLDA}: single pass LDA \citep{SinglePassLDA}. The difference with \texttt{V-OEM++} is that {$\rho_i= {1}/{i}$} and the updates in $\alpha$ done with a Gamma prior (see Appendix~\ref{app:alpha});
\item \texttt{SGS}: streaming Gibbs sampling \citep{StreamingGibbs2016}. This method is related to \texttt{G-OEM} with {$\rho_i={1}/{i}$}. In this method, $\alpha$ is not optimized and set to a constant $C_{\alpha}$. For comparison purposes, for each dataset, we set $C_\alpha$ to be the averaged final parameter $\hat{\alpha}$ obtained with \texttt{G-OEM} on the same dataset: $C_\alpha=\frac{1}{K}\sum_k\hat{\alpha}_k$. {For each observation, only the last Gibbs sample is considered, leading to extra noise in the output};
\item \texttt{LDS}: Stochastic gradient Riemannian Langevin dynamics sampler \citep{Patterson2013}. The authors use the Langevin Monte Carlo methods on probability simplex and apply their online algorithm to LDA. For this method and only this method, we set to $P=200$ the number of internal updates.

\end{itemize}
For existing variational methods---\texttt{OLDA}, \texttt{SVB}, \texttt{SPLDA}---$\beta$ is a random variable with prior $q(\beta)$. We estimate the likelihood $p(X|\hat{\beta},\alpha)$ with the ``left-to-right'' algorithm by setting $\hat{\beta}=\mathbb{E}_q[\beta]$ for Bayesian methods. For simplicity, we only present our results obtained with \texttt{G-OEM} and \texttt{V-OEM++}. Indeed, the inference boost presented in Section~\ref{sec:LDA} is only beneficial for \texttt{V-OEM}. A detailed analysis is presented in Appendix~\ref{app:boostinf}.

\begin{table}[!b]
\begin{center}
\begin{sc}\begin{tabular}{|l|cccc|}
\hline
 & Category & $\mathbb{E}_{Z|X,\eta}[S(X,Z)]$ & Step-size $\rho_t$ & Update for $\alpha$\\
\hline
\texttt{G-OEM}    & frequentist & Gibbs sampling & free & fixed point\\
\texttt{V-OEM}    & frequentist & variational    & free & fixed point\\
\hline
\texttt{OLDA}     & Bayesian    & variational    & free & gradient ascent\\
\texttt{VarGibbs} & Bayesian    & Gibbs sampling & free & $\alpha$ fixed \\
\texttt{SVB}      & Bayesian    & variational    & fixed: $1/t$ & gradient ascent \\
\texttt{SPLDA}    & frequentist & variational    & fixed: $1/t$ & Gamma prior\\
\texttt{SGS}      & frequentist & Gibbs sampling & fixed: $1/t$ & $\alpha$ fixed\\
\hline
\end{tabular}
\end{sc}
\end{center}
\vskip -0.5cm
\caption{Comparison of existing methods for LDA.}
\label{tab:comparison}
\end{table}

\subsection{Explicit Links for LDA}
In this section, we propose to make the links between the methods listed above explicit, using the framework described in Section~\ref{sec:LDA} for the particular LDA model. We present in Table~\ref{tab:comparison} a summary of the compared method.
\subsubsection{Category} In the frequentist approach, $\beta$ is a parameter and is updated with Equation~(\ref{eqn:MstepLDA}), as the ``M-step'' in online EM.

In a Bayesian setting, $\beta$ is a random variable with prior $\beta\sim\mathrm{Dirichlet}(b\mathbf{1})$, with $b\in\mathbb{R}$ and~ ${\mathbf{1}\in\mathbb{R}^V}$ denotes the vector whose all entries equal $1$. The variational distribution of the global parameter $\beta$ is then set to $q(\beta^k|\lambda^k)=\mathrm{Dirichlet}(\lambda^k)$, with $\lambda^k\in\mathbb{R}^{V}$ $\forall k=1,\ldots,K$. The variational parameter $\lambda^k$ is updated by maximizing the ELBO with stochastic gradient ascent (Equation~(\ref{eqn:bayes})).

\subsubsection{Estimation of $\mathbb{E}_{Z|X,\eta}[S(X,Z)]$} For LDA, the expectation $\mathbb{E}_{Z|X,\eta}[S(X,Z)]$ can either be estimated with Gibbs sampling---using Equation~(\ref{eqn:Gibbs})---or with variational approximation---using Equation~(\ref{eqn:ELBO2}).

\subsubsection{Step-size} Some of the methods listed above (\texttt{SVB}, \texttt{SPLDA} and \texttt{SGS}) are incremental, which means the sufficient statistics are incrementally aggregated $s_t=s_{t-1} + \mathbb{E}_{Z_t|X_{t},\eta}[S(X_{t},Z_t)]$ . For LDA, it exactly corresponds to a step-size $\rho_t=1/t$ in the online EM setting, even though the link is not explicit in the corresponding papers.

For the other listed methods, the step-size exponent $\kappa$ is chosen arbitrarily in ${[0.5,1)}$, with ${\rho_t=1/t^\kappa}$. However, results are mostly presented with $\kappa=1/2$ and $\rho_t=1/\sqrt{t}$.

\subsection{General Settings}
\subsubsection{Initialization}
We initialize randomly $\eta\equiv (\beta,\alpha)$. For a given experiment, we initialize all the methods with the same values of $(\beta,\alpha)$ for fair comparison, except \texttt{SPLDA} that has its own initilization scheme---see \citet{SinglePassLDA} for more details.

\subsubsection{Minibatch}
{
We consider minibatches of size 100 documents for each update in order to reduce noise \citep{OnlineEMliang2009}.
In the case of online EM in Equation~(\ref{eqn:OEM}), we estimate an expectation for each observation of the minibatch. We update the new sufficient statistics $s$ towards the average of the expectations over the minibatch. We do the same averaging for all the presented methods.
}
\subsubsection{Number of Local Updates} For all the presented methods, we set the number of passes through each minibatch to $P=20$. For \texttt{G-OEM}, this means that we perform 20 Gibbs sampling for each word of the minibatch. All other methods access each document 20 times (e.g., 20 iterations of variational inference on each document). For \texttt{G-OEM}, inference with larger values for $P$ (e.g., $P=50$ or $P=100$) leads to very similar results.

\subsubsection{Datasets}
We apply the methods on {six} differents datasets, summarized in Table~\ref{tab:data} ($\overline{N_X}$ is  the average length of documents). Following~\citet{LDA}, the synthetic dataset has been generated from 10 topics and the length of each document drawn from a Poisson(60). The 10 topics are inferred with online LDA \citep{OnlineLDA} from 50,000 reviews of the IMDB dataset with a vocabulary size of 10,000. We only consider the entries of the 1,000 most frequent words of this dataset that we normalize to satisfy the constraint $\sum_v \beta^k_{v}=1$.

\begin{table}
\begin{center}
\begin{sc}\begin{tabular}{|lccc|}
\hline
Dataset & \#documents & $\overline{N_X}$ & \#words\\
\hline
Synthetic  & 1,000,000 & 60 & 1,000  \\
Wikipedia\footnotemark[1]  & 1,010,000 & 162.3 & 7702 \\
IMDB\footnotemark[2] & 614,589 & 82.2 & 10,000\\
Amazon movies\footnotemark[3] & 338,565 & 75.4 & 10,000 \\
New York Times\footnotemark[4] & 299,877 & 287.4 & 44,228\\
Pubmed\footnotemark[4] & 2,100,000 & 82.0 & 113,568\\
\hline
\end{tabular}
\end{sc}
\end{center}
\vskip -0.5cm
\caption{Datasets.}
\label{tab:data}
\end{table}

The words in the datasets IMDB, Wikipedia, New York Times, Pubmed {and Amazon movies} are filtered by removing the stop-words and we select the most frequent words of the datasets. For the synthetic dataset, IMDB, Pubmed {and Amazon movies}, the size of the test sets is $N_t=$ 5,000 documents. For Wikipedia and New York Times, the test sets contain $N_t=$ 2,000 documents.

We run the methods on 11 differents train/test splits of each dataset. For all the presented results, we plot the median from the 11 experiments as a line---solid or dashed. For the sake of readability, we only present the same plots with error bars between the third and the seventh decile in Appendix~\ref{app:Kcomp} and Appendix~\ref{app:ite}.

\footnotetext[1]{Code available from \citet{OnlineLDA} }
\footnotetext[2]{Dataset described in \citet{JMARS}}
\footnotetext[3]{Data from \citet{AmazonData}}
\footnotetext[4]{\label{data:UCI}UCI dataset \citep{UCIdata}}

\subsubsection{Computation Time}
For each presented method and dataset, the computational time is reported in Table~\ref{tab:time}. 
Although all methods have the same running-time complexities, coded in Python, sampling methods (\texttt{G-OEM}, \texttt{VarGibbs} and \texttt{SGS}) need an actual loop over all documents while variational methods (\texttt{OLDA}, \texttt{SVB}, \texttt{SPLDA} and \texttt{V-OEM++}) may use vector operations, and may thus be up to twice faster. This could be mitigated by using
efficient implementations of Gibbs sampling on minibatches \citep{GPUGibbsLDA2009,SAME2014,StreamingGibbs2016}. Note also that to attain a given log-likelihood, our method \texttt{G-OEM} is significantly faster and often attains log-likelihoods not attainable by other methods (e.g., for the dataset New York Times).

\subsubsection{Step-size}
In the following, we compare the results of our methods \texttt{G-OEM} and \texttt{V-OEM++} with ${\kappa=1/2}$, i.e., the step-size $\rho_t = 1 / \sqrt{t}$, {without averaging.} Detailed analysis of different settings of our method can be found in Appendix~\ref{app:Gibbssettings}. In particular, we compare different step-sizes and the effect of averaging over all iterates. We also compare the performance of \texttt{OLDA} with different step-sizes in Appendix~\ref{app:rho} and observe that results are very similar for all the step-sizes that we try. Note that for incremental methods (\texttt{SVB}, \texttt{SPLDA}, \texttt{SGS}), the step-size is fixed to $\rho_t=1/t$. For \texttt{LDS}, we run the method with parameters as close as possible to our method for fair comparison.

\begin{table}[t]
\begin{center}
\begin{sc}
\begin{tabular}{|lcccc|}
\hline
  & IMDB & Wikipedia & NYT & Pubmed \\
\hline
\texttt{G-OEM}   & 13h  & 55h  & 30h  & 58h \\
\texttt{V-OEM++} &  9h  & 37h  & 20h  & 54h \\
\texttt{OLDA}    &  7h  & 33h  &  8h  & 30h \\
\texttt{VarGibbs}& 12h  & 50h  & 28h  & 54h \\
\texttt{SVB}     &  7h  & 34h  &  9h  & 30h \\
\texttt{SPLDA}   &  9h  & 37h  & 20h  & 54h \\
\texttt{SGS}     & 11h  & 48h  & 27h  & 50h \\
\texttt{LDS}     &  7h  & 17h  & 12h  & 40h \\
\hline
\end{tabular}
\end{sc}
\end{center}
\vskip -0.5cm
\caption{Average computational time (in hours) for each method---$K=128$.}
\label{tab:time}
\end{table}

\subsection{Results on LDA}
Results obtained with the presented methods applied to LDA on different datasets for different values of the number $K$ of topics are presented in Figure~\ref{fig:Kcomp}. Performance through iterations (i.e., as the number of documents increases) is presented in Figure~\ref{fig:ite}.
We first observe that for all experiments, our new method \texttt{G-OEM} performs better---often significantly---than all existing methods. In particular, it is highly robust to diversity of datasets.

\subsubsection{Influence of the Number of Topics $K$}
\begin{figure}
\begin{center}
\subfigure[Dataset: Synthetic]{\includegraphics[width=0.48\columnwidth]{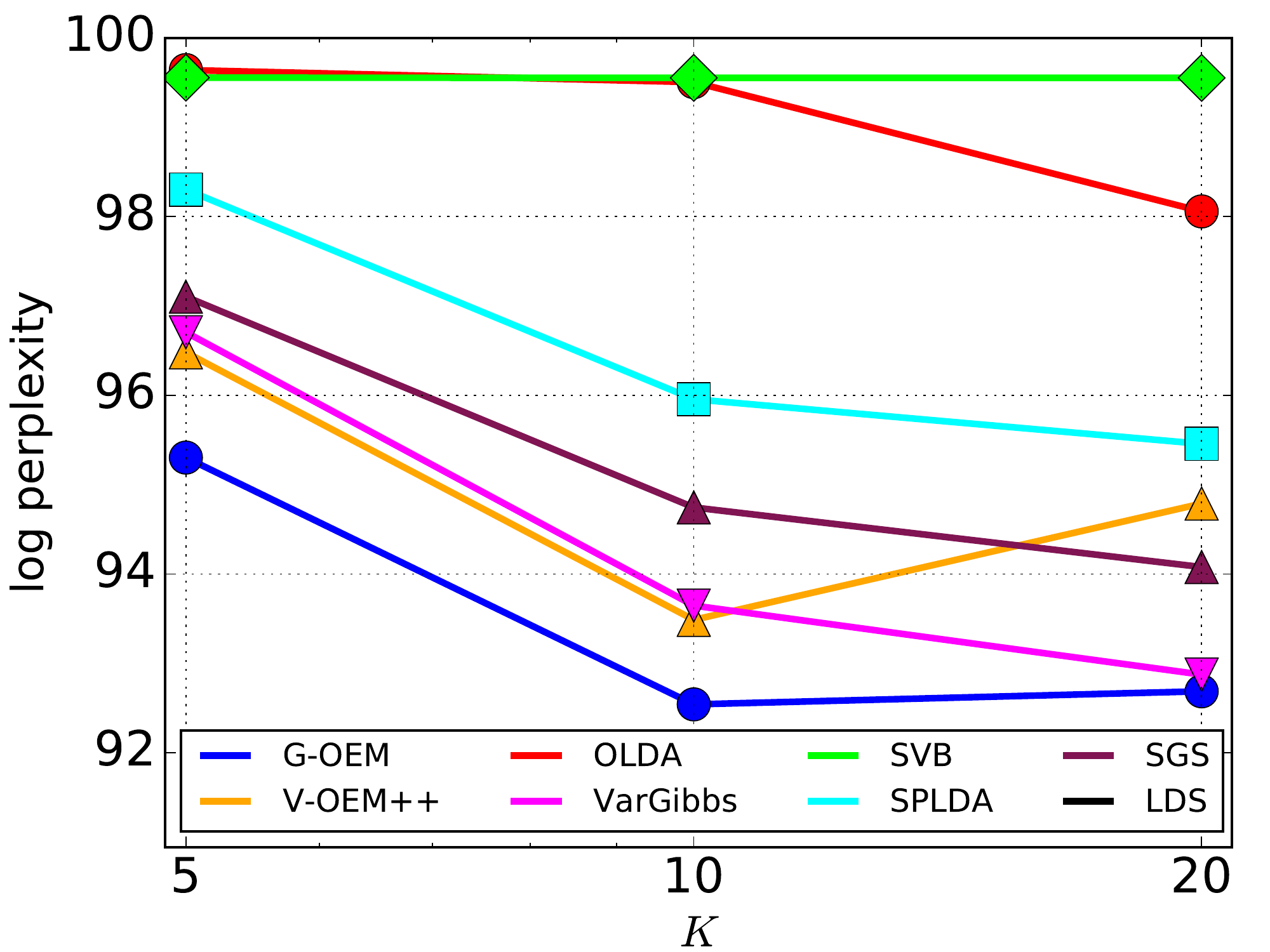}\label{fig:Synth-Kcomp}}
\subfigure[Dataset: IMDB]{\includegraphics[width=0.48\columnwidth]{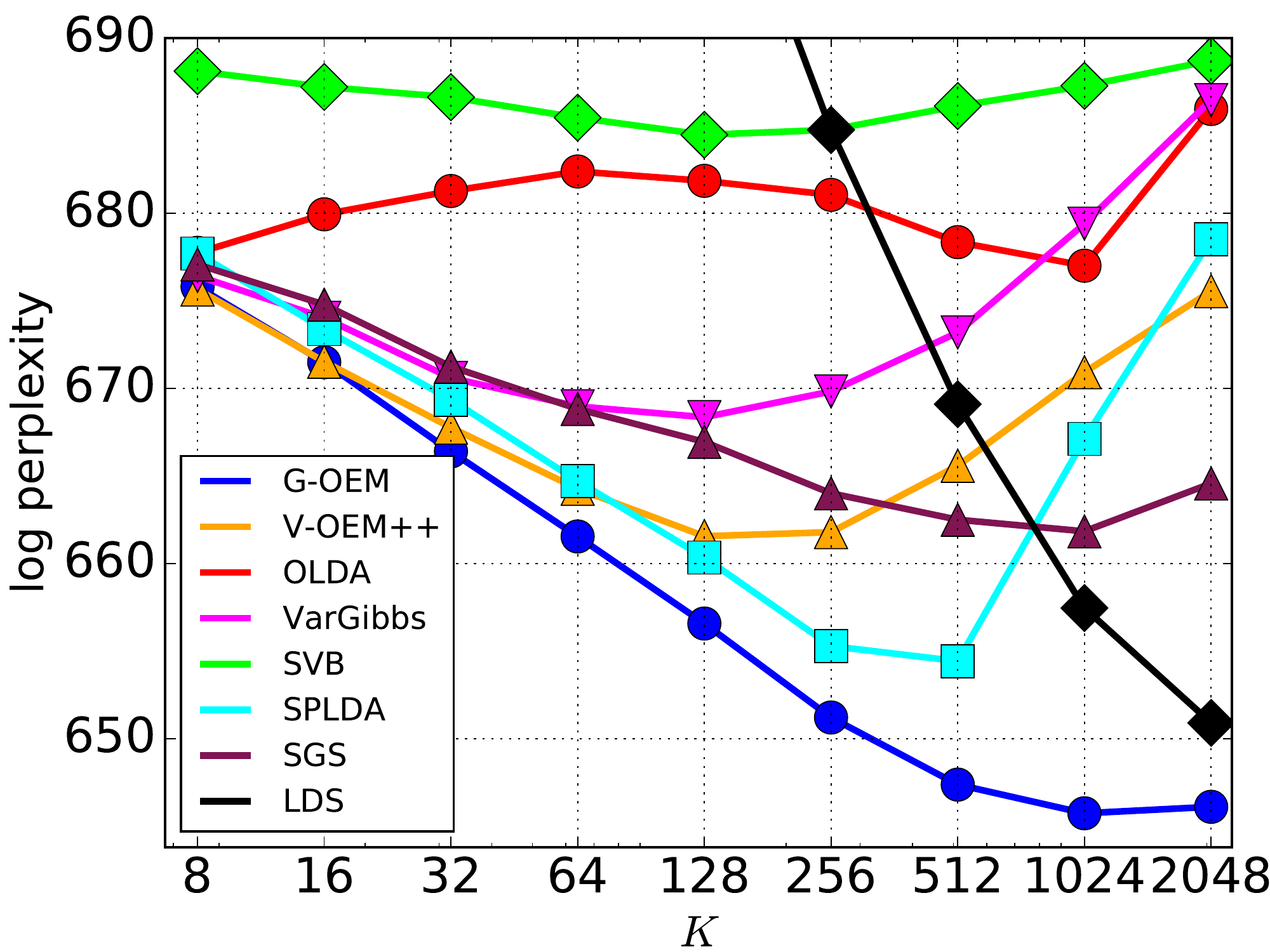}\label{fig:IMDB-Kcomp}}

\subfigure[Dataset: Wikipedia]{\includegraphics[width=0.48\columnwidth]{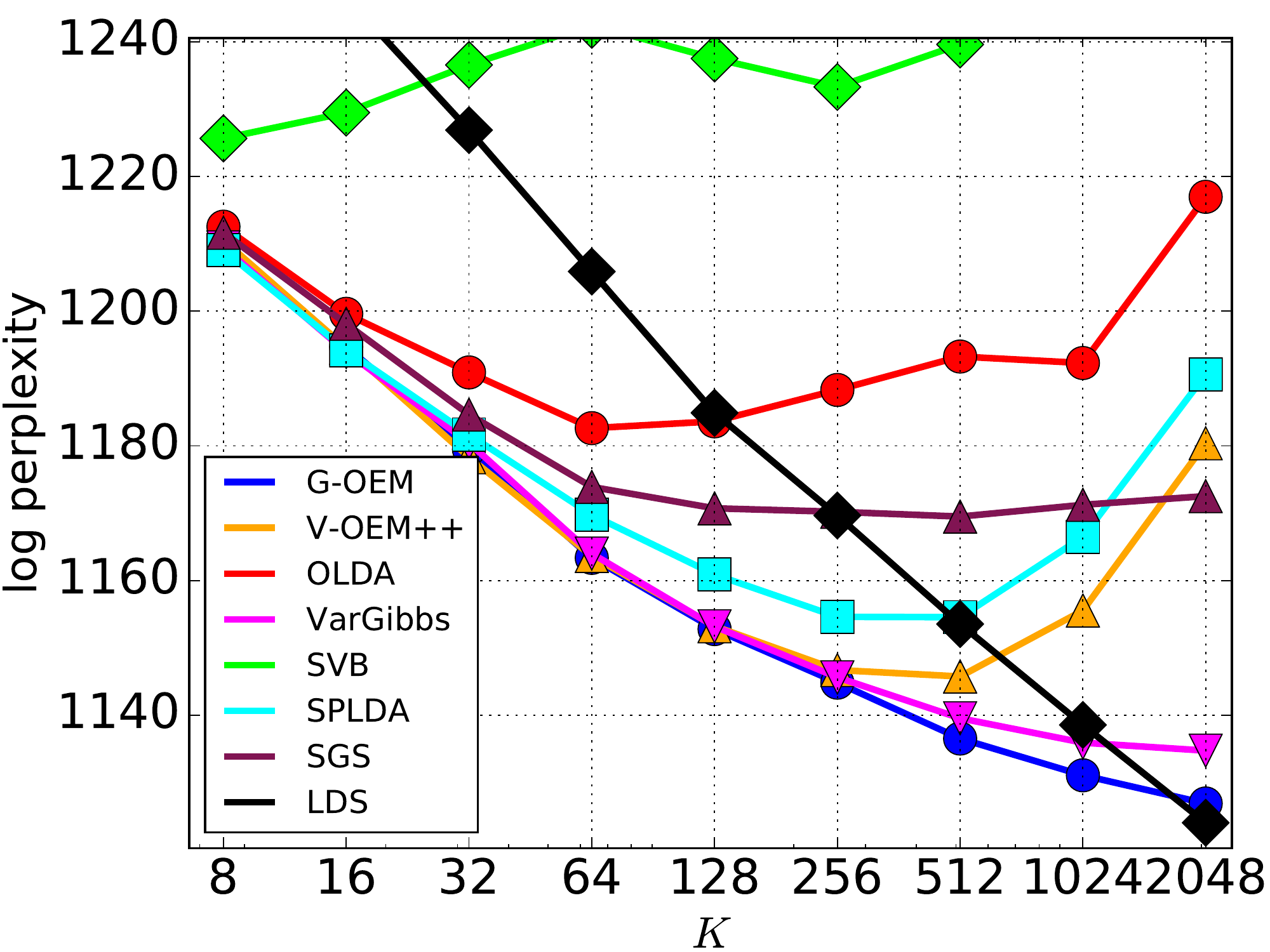}\label{fig:wiki-Kcomp}}
\subfigure[Dataset: New York Times]{\includegraphics[width=0.48\columnwidth]{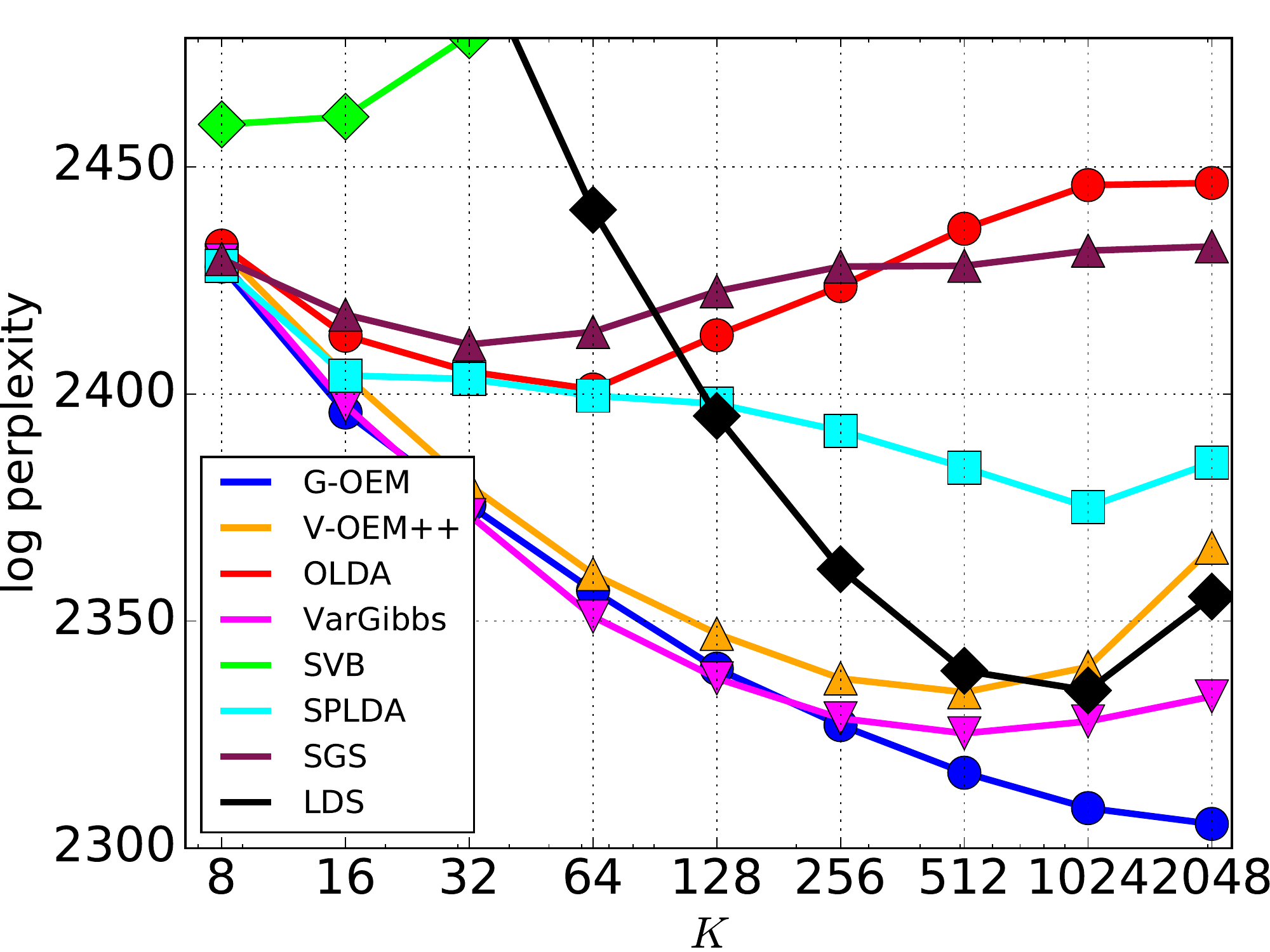}\label{fig:NYT-Kcomp}}

\subfigure[Dataset: Pubmed]{\includegraphics[width=0.48\columnwidth]{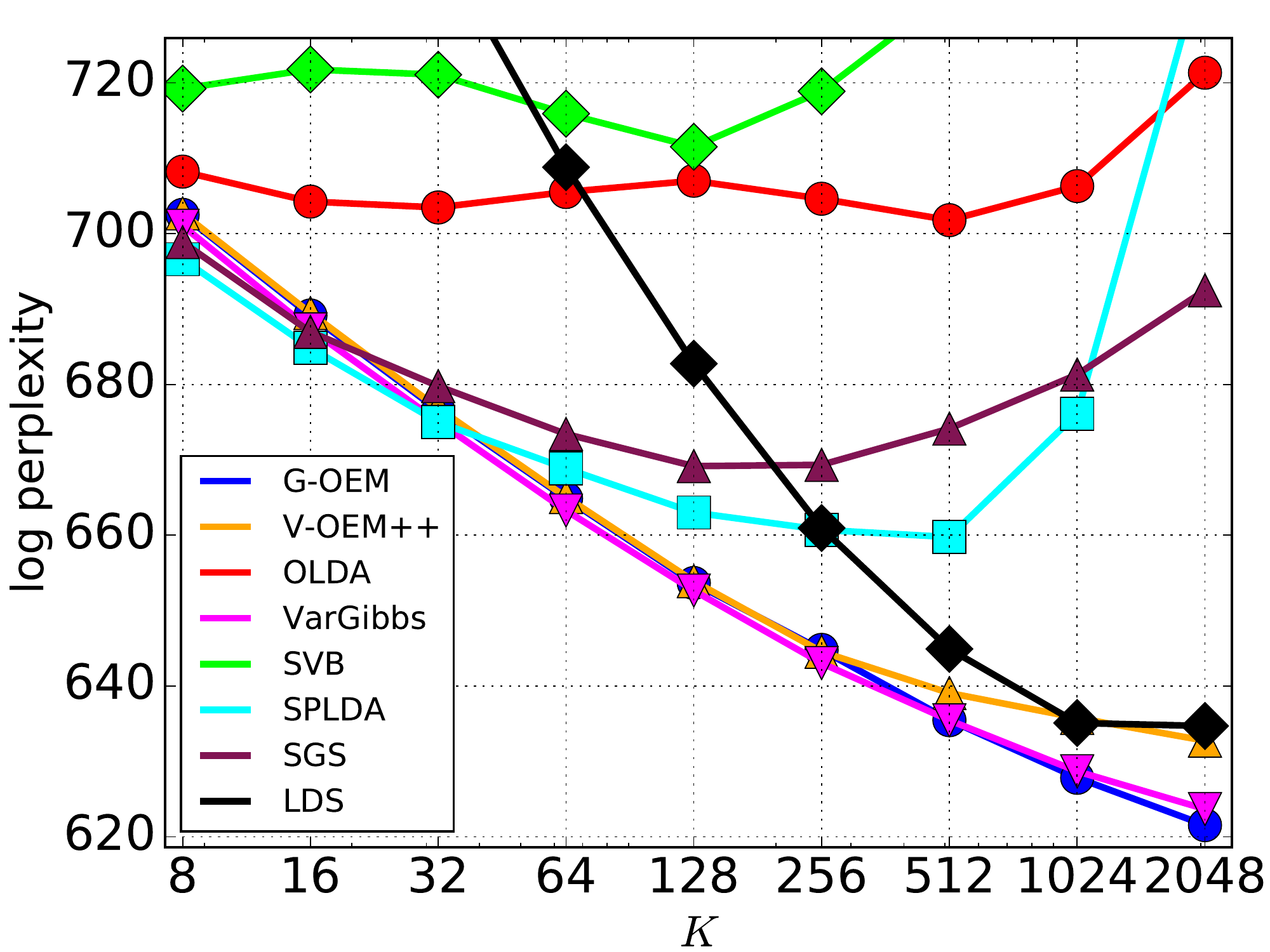}\label{fig:Pubmed-Kcomp}}
\subfigure[Dataset: Amazon movies]{\includegraphics[width=0.48\columnwidth]{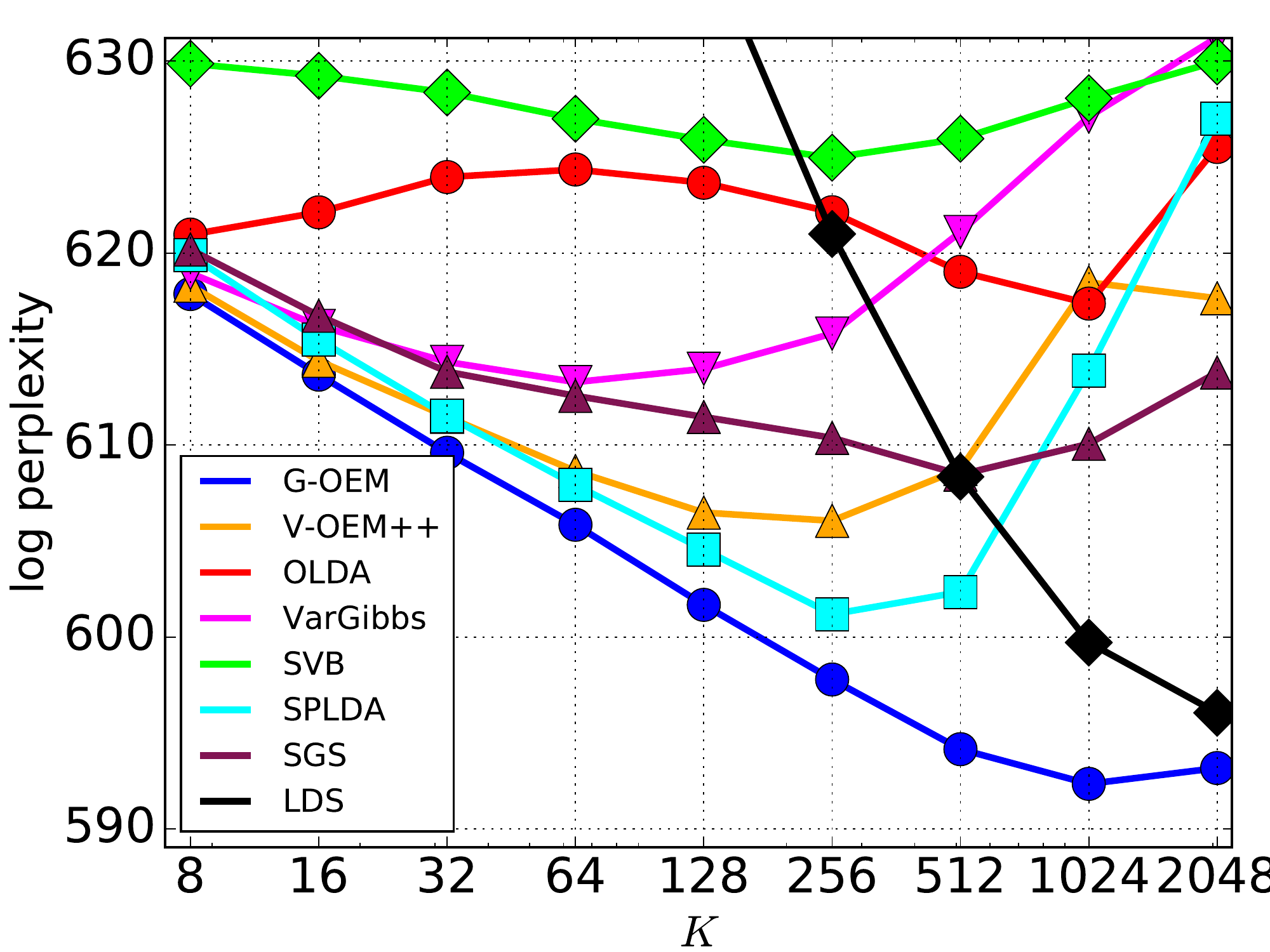}\label{fig:Amazon-Kcomp}}
\caption{Perplexity on different test sets as a function of $K$, the number of topics inferred. Best seen in color.}
\label{fig:Kcomp}
\end{center}
\end{figure}

As shown in Figure~\ref{fig:Kcomp}, for  synthetic data in plot (a), although the true number of topics is $K^*=10$, \texttt{SPLDA}, \texttt{OLDA}, \texttt{VarGibbs}  and \texttt{SGS} perform slightly better with $K=20$, while \texttt{G-OEM} has the better fit for the correct value of $K$; moreover,  \texttt{SVB} has very similar performances for any value of $K$, which highlights the fact that this method does not capture more information with a higher value of $K$. \texttt{LDS} performs very poorly on this dataset---for any value of $K$ the log-perplexity is around 400---and is not displayed in Figure~\ref{fig:Kcomp} (a) for clarity.

On non-synthetic datasets in plots (b)-(f), while the log-perplexity of frequentist methods---\texttt{G-OEM}, \texttt{V-OEM++} and \texttt{SPLDA}---decreases with $K$, the log-perplexity of variational Bayesian methods---\texttt{OLDA} and \texttt{SVB}---does not decrease significantly with $K$. As explained below, our interpretation is that the actual  maximization of  the ELBO does not lead to an improvement in log-likelihood. The hybrid Bayesian method \texttt{VarGibbs}---which uses Gibbs sampling for local updates ($\theta,z$) and variational updates for global parameters $(\beta,\alpha)$---performs much better than the variational Bayesian methods. Our interpretation is that the objective function maximized with \texttt{VarGibbs} is a much better approximation of the log-likelihood than the ELBO. 

In terms of robustness, \texttt{G-OEM} and \texttt{LDS} are the only methods that do not display overfitting on any dataset. However, \texttt{LDS} is only competitive for the highest values of $K$---$K\geq 500$.

\subsubsection{Performance Through Iterations}
\begin{figure}
\begin{center}
\subfigure[\label{fig:Synth_ite}Synthetic, $K=10$]{\includegraphics[width=0.48\columnwidth]{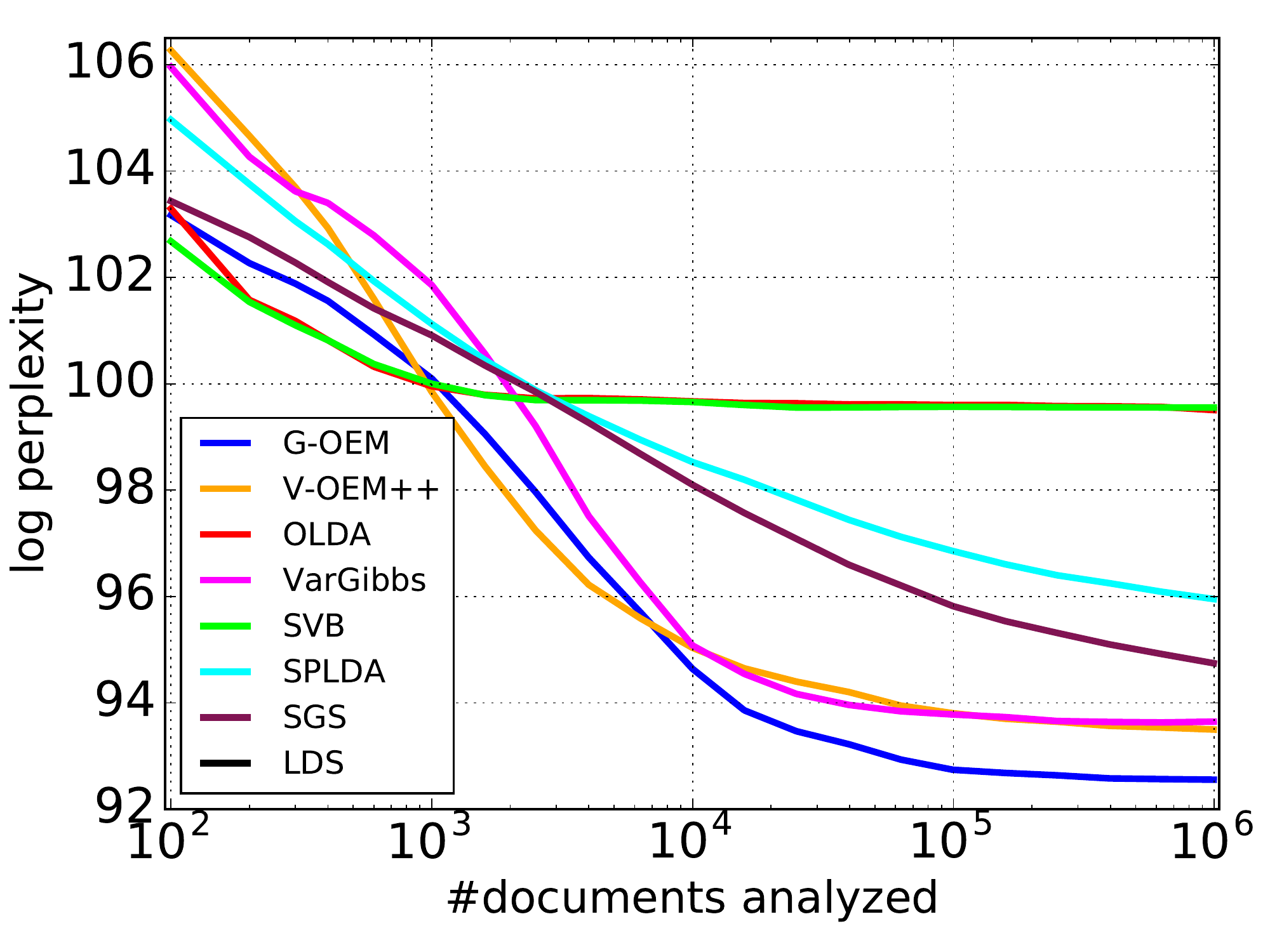}}
\subfigure[\label{fig:IMDB_ite}IMDB, $K=128$]{\includegraphics[width=0.48\columnwidth]{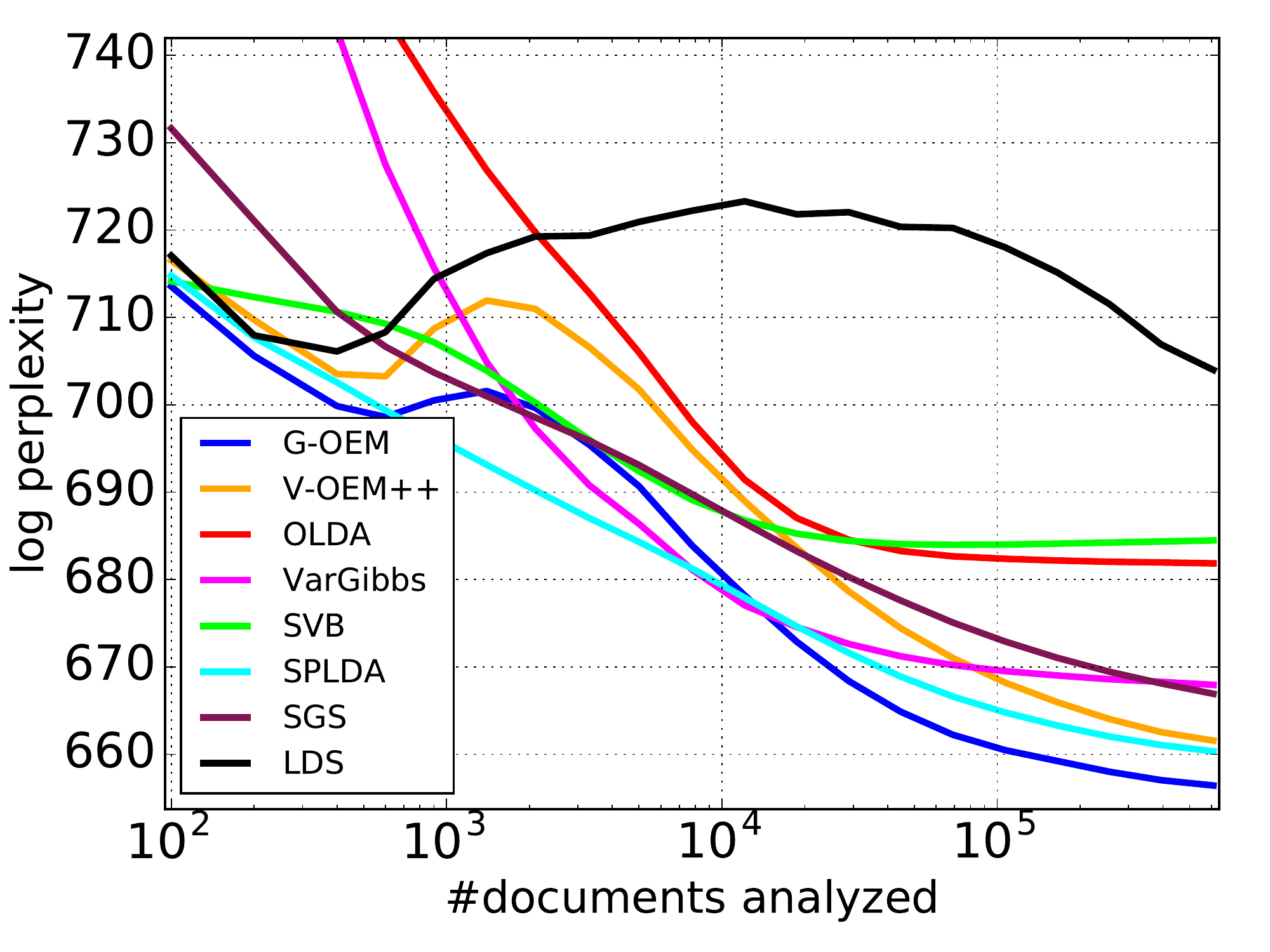}}

\subfigure[\label{fig:wiki_ite}Wikipedia, $K=128$]{\includegraphics[width=0.48\columnwidth]{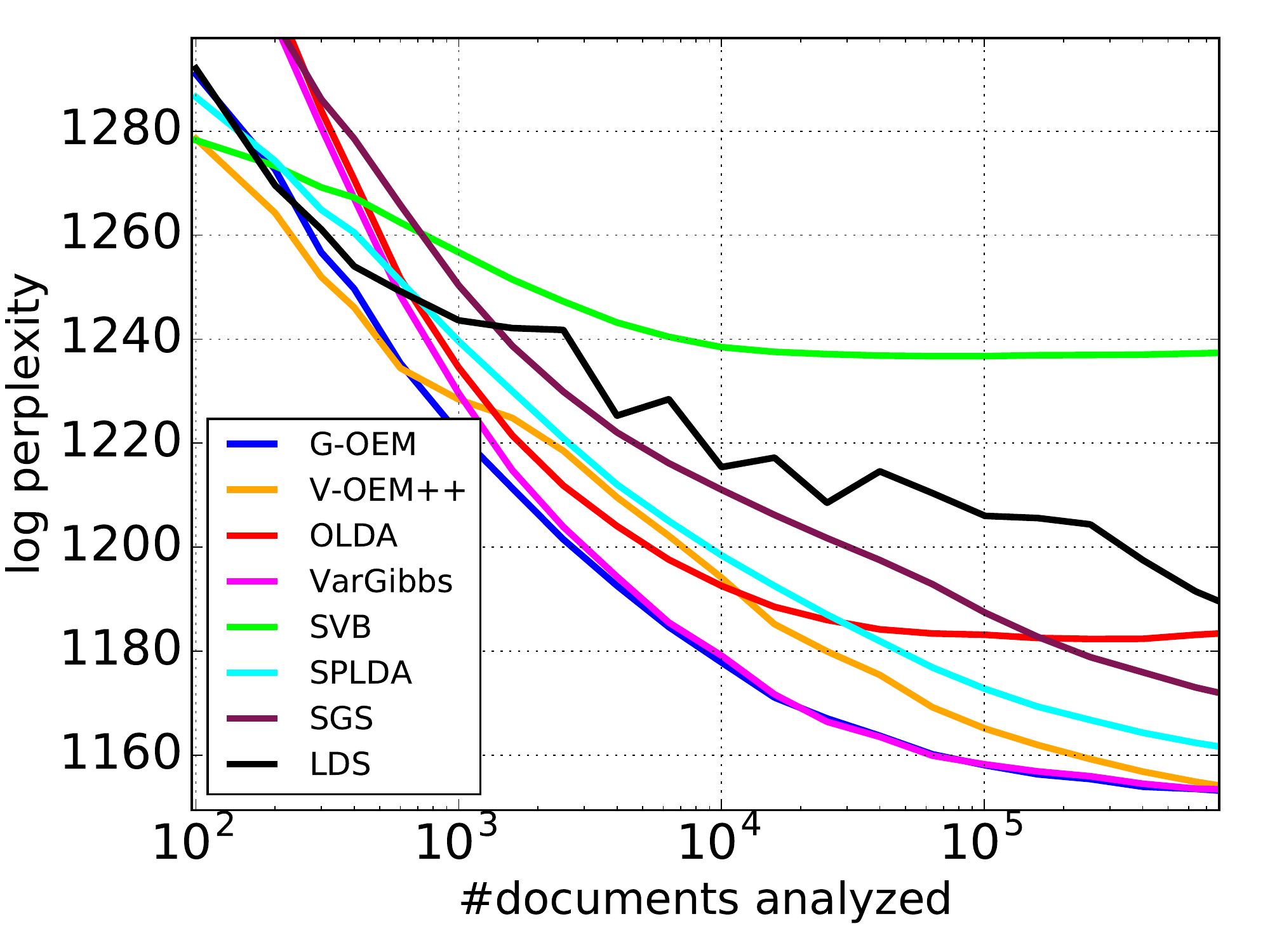}}
\subfigure[\label{fig:NYT_ite}New York Times, $K=128$]{\includegraphics[width=0.48\columnwidth]{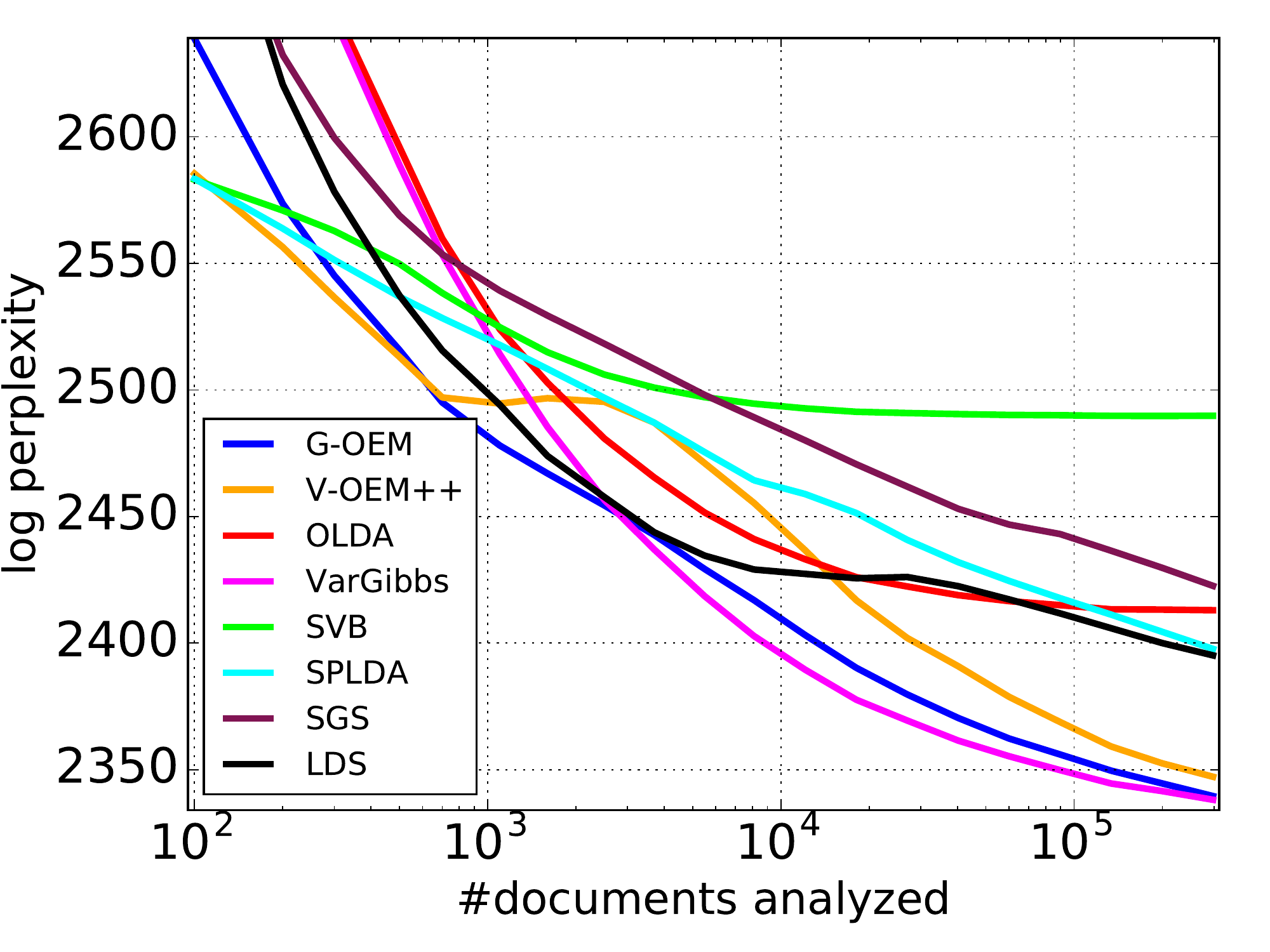}}

\subfigure[\label{fig:Pubmed_ite}Pubmed, $K=128$]{\includegraphics[width=0.48\columnwidth]{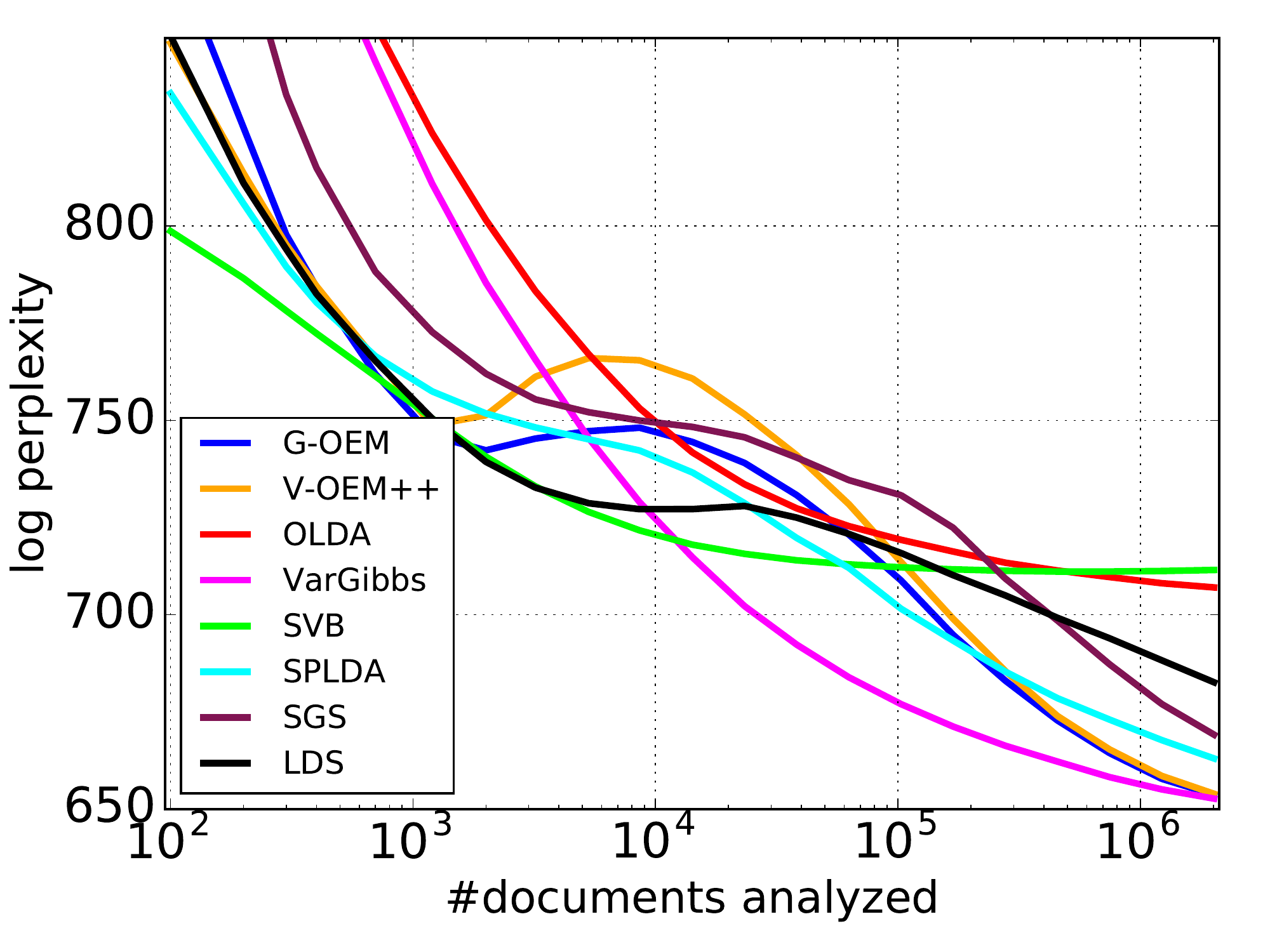}}
\subfigure[\label{fig:Amazon_ite}Amazon movies, $K=128$]{\includegraphics[width=0.48\columnwidth]{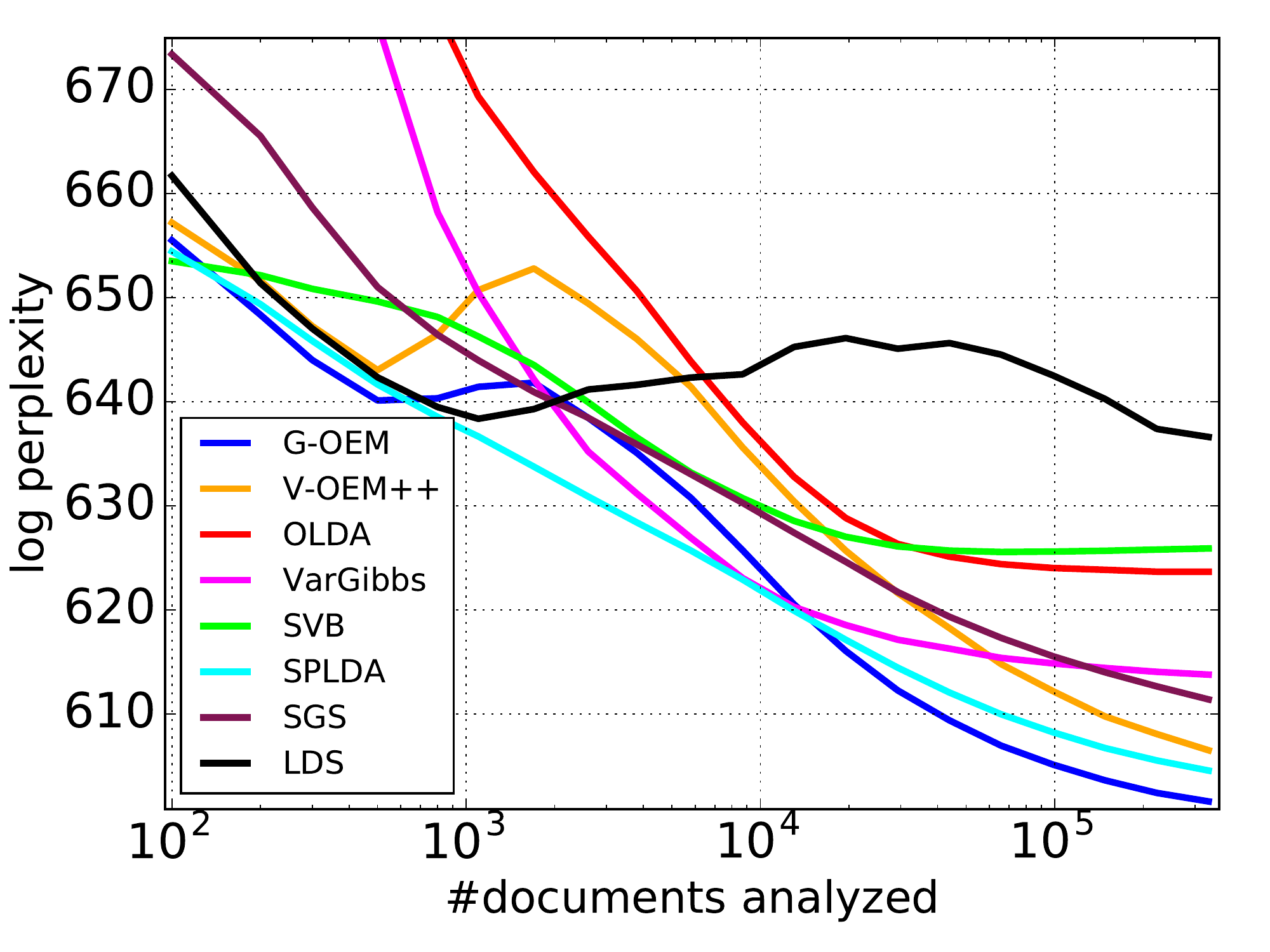}}

\caption{Perplexity through iterations on different test sets with the presented methods. Best seen in color.}
\label{fig:ite}
\end{center}
\end{figure}

As shown in Figure~\ref{fig:ite}, for synthetic data in plot (a), after only few dozens of iterations---few thousands of documents seen---\texttt{G-OEM}, \texttt{V-OEM++} and \texttt{VarGibbs} outperform the other presented methods. Variational Bayesian methods again do converge but to a worse parameter value.
On real datasets in plots (b)-(f), \texttt{G-OEM} and \texttt{VarGibbs} are significantly faster; we can indeed still observe that after around 100 iterations---10,000 documents seen---\texttt{G-OEM} and \texttt{VarGibbs} perform better than other methods on all the datasets except Pubmed, where the performances of \texttt{G-OEM}, \texttt{V-OEM++}, \texttt{VarGibbs} and \texttt{SPLDA} are  similar.
Note that

\subsubsection{Variational vs.~Sampling} Our method \texttt{G-OEM} directly optimizes the likelihood with a consistent approximation, and  performs  better than its variational counterparts \texttt{SPLDA} and \texttt{V-OEM++} in all experiments. The hybrid method \texttt{VarGibbs} is less robust than \texttt{G-OEM} as it performs either similarly to \texttt{G-OEM}---for the datasets Wikipedia, New York Times and Pubmed---or worse than \texttt{G-OEM} and its variational counterparts \texttt{SPLDA} and \texttt{V-OEM++}---for the datasets IMDB and Amazon.

\subsubsection{Frequentist vs.~Bayesian}
\begin{figure}[t]
\begin{center}
\subfigure[Synthetic dataset.]{\includegraphics[width=0.49\columnwidth]{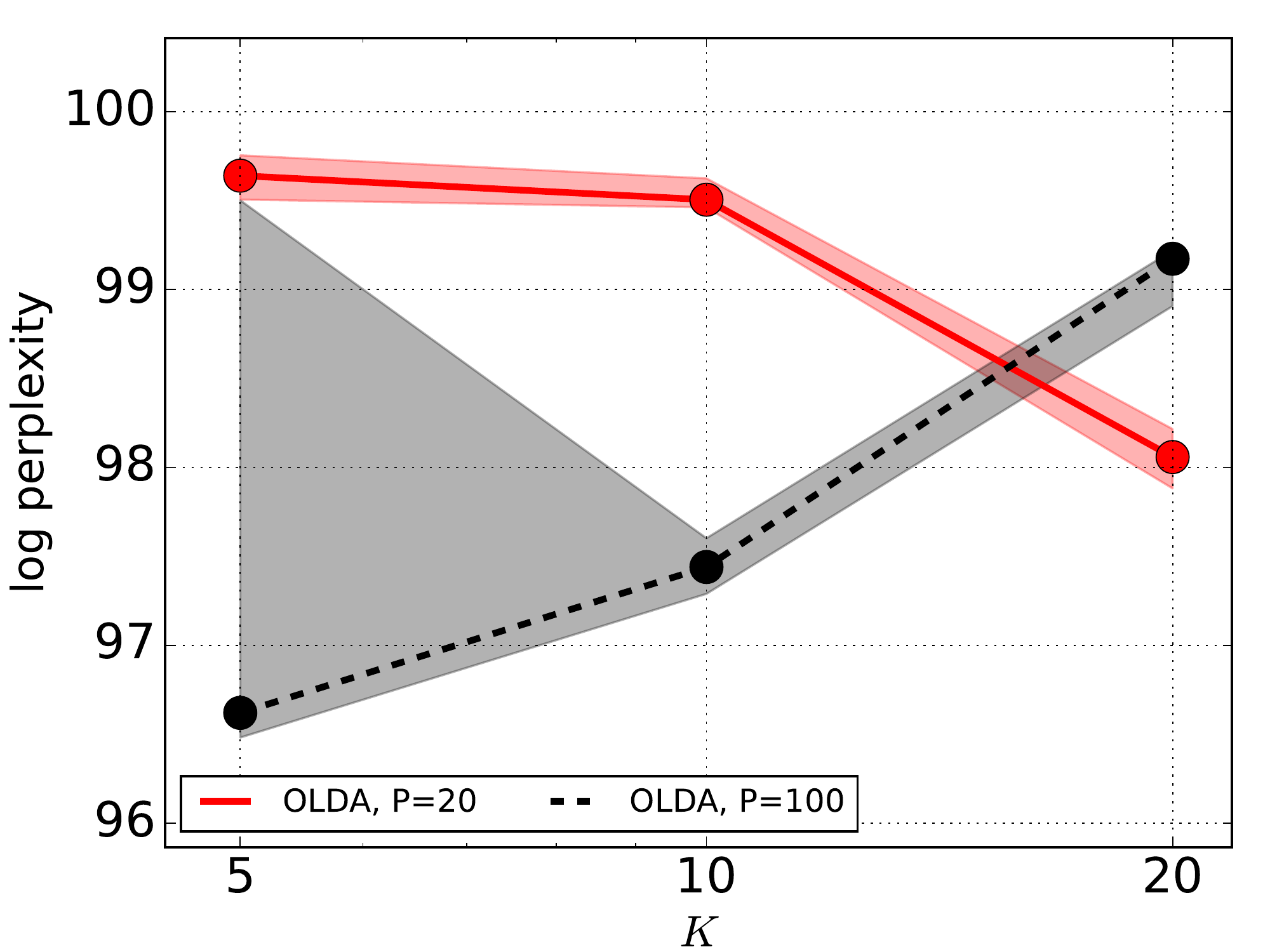}}
\subfigure[IMDB.]{\includegraphics[width=0.49\columnwidth]{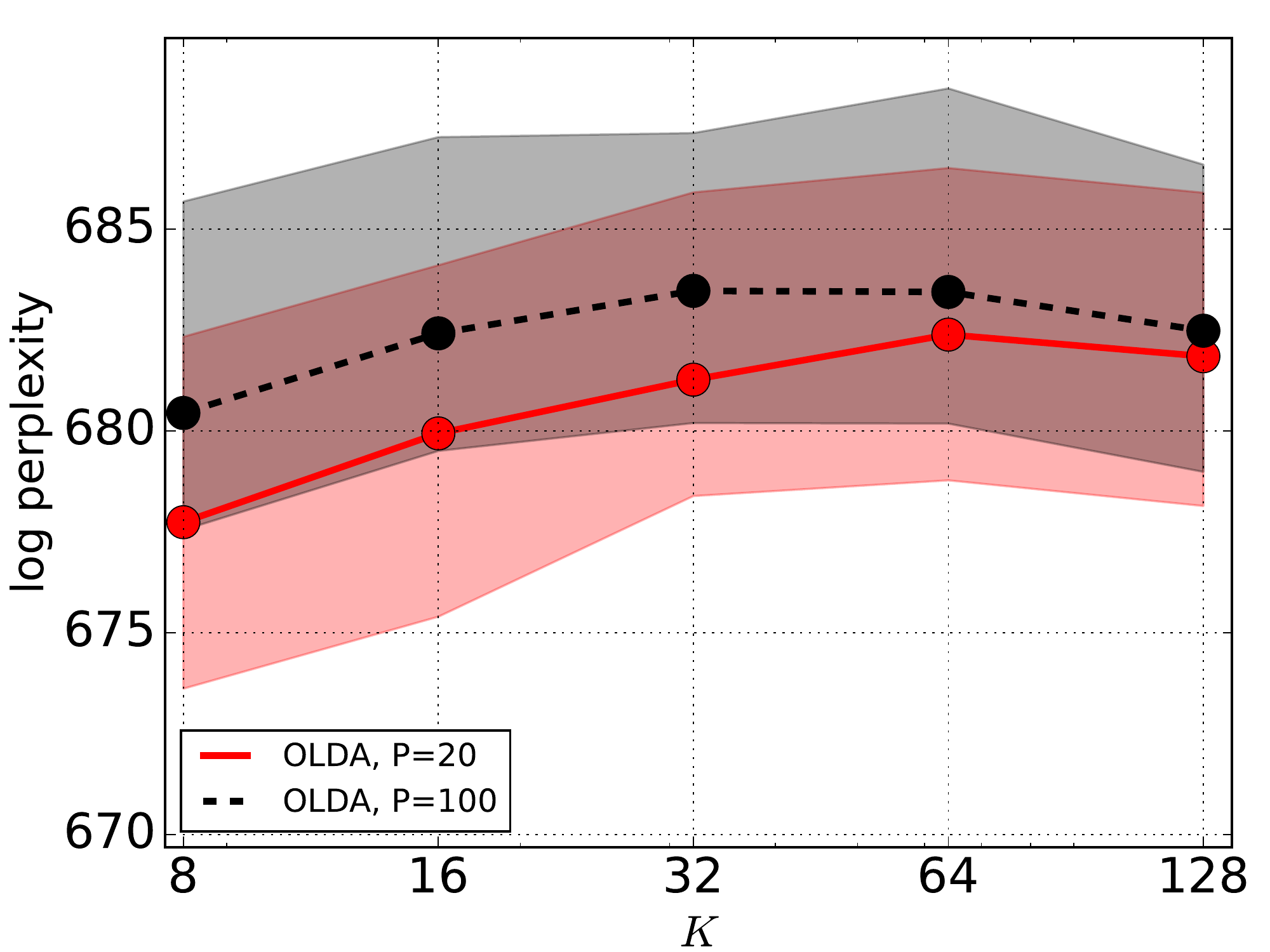}}
\caption{OLDA. Perplexity on different test sets as a function of $K$ for \texttt{OLDA} with $P=10$ (red) and $P=100$ (black) internal updates. }
\label{fig:Kcomp_P100}
\end{center}
\end{figure}

\begin{figure}[t]
\begin{center}
\subfigure{\includegraphics[width=0.49\columnwidth]{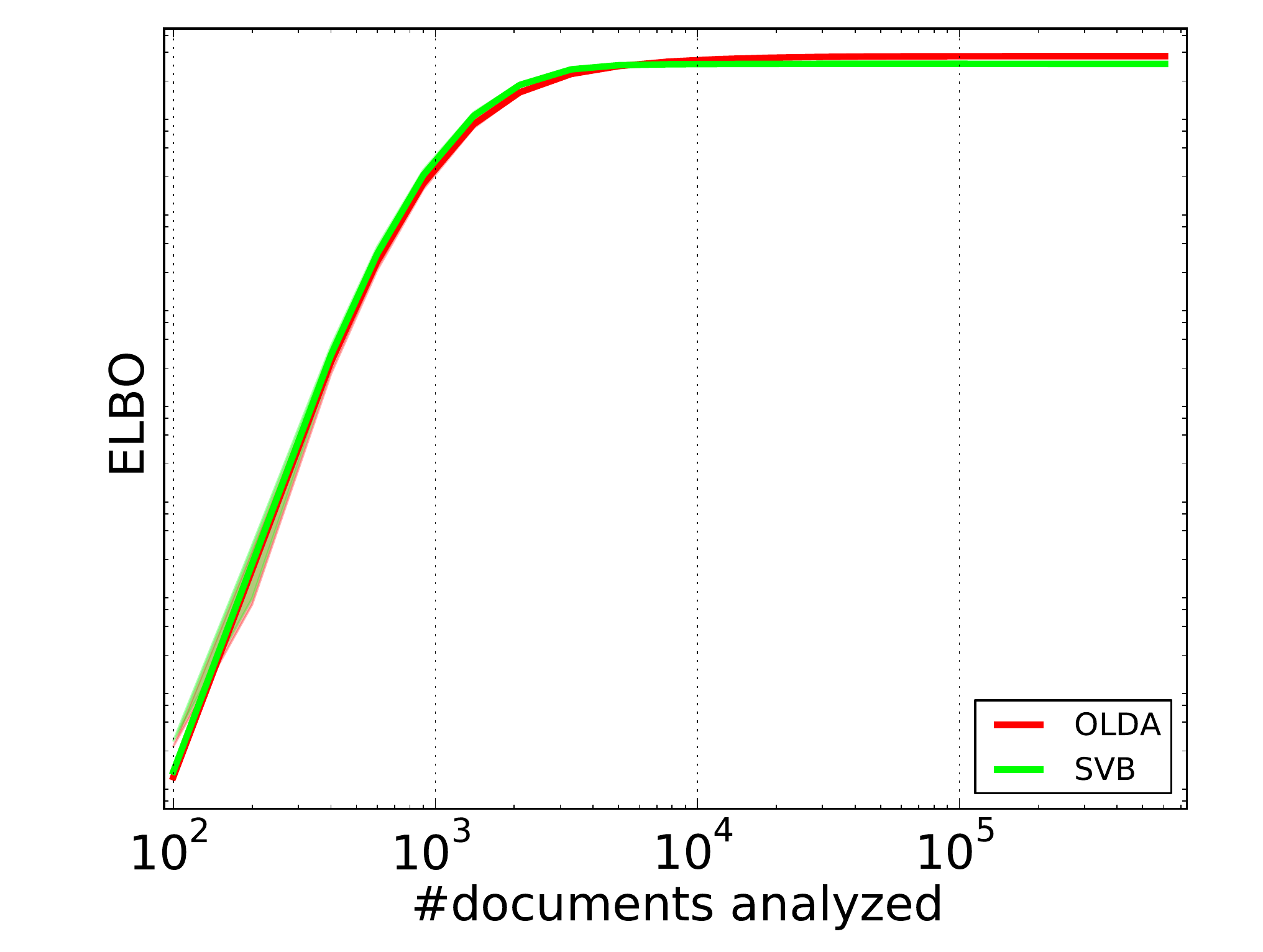}}
\subfigure{\includegraphics[width=0.49\columnwidth]{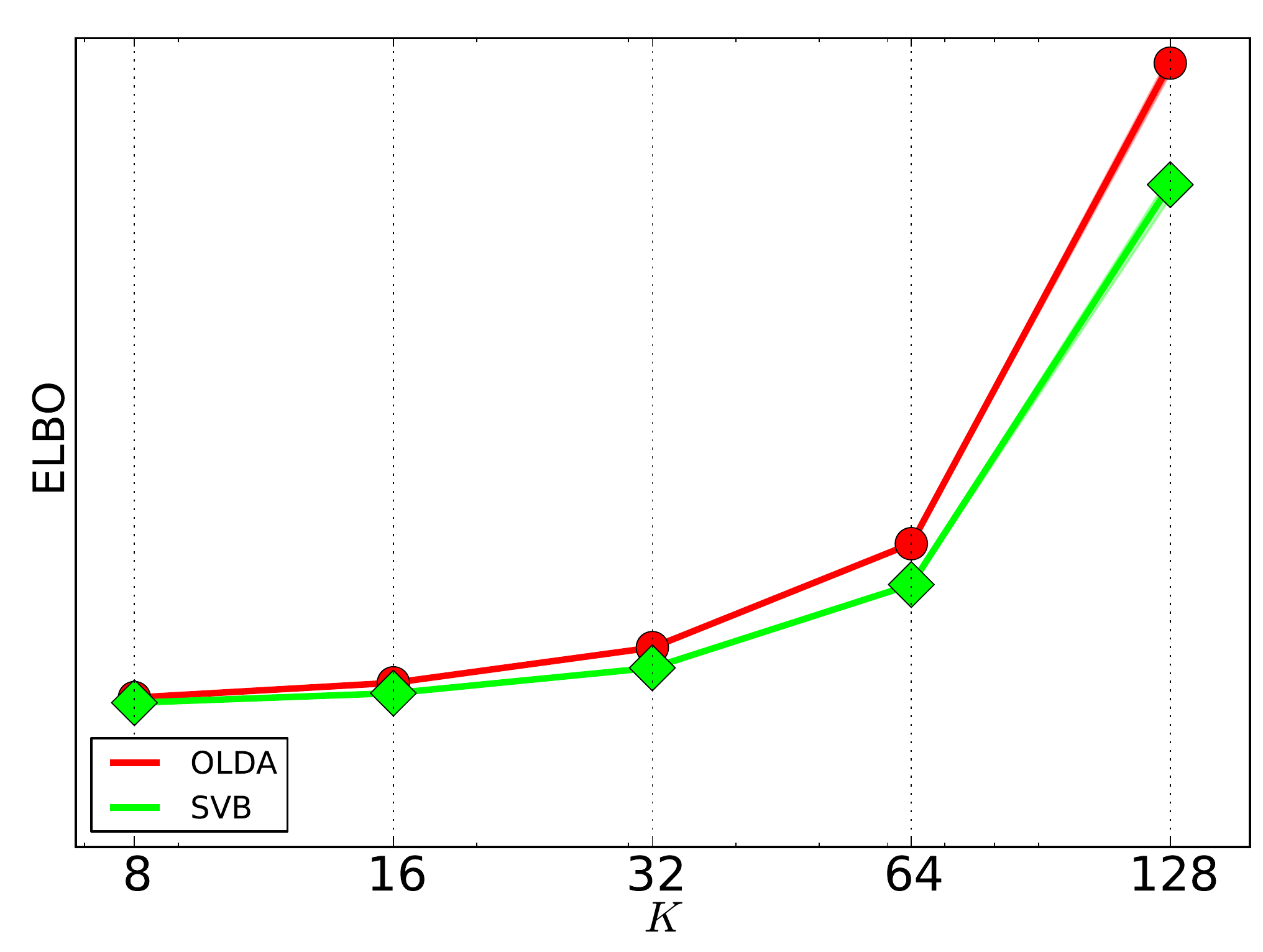}}
\caption{Dataset: IMDB. Evidence Lower BOund (ELBO) computed on test sets (20 internal iterations and 1 pass over the dataset). \textit{Left}: ELBO through iterations with $K=128$. \textit{Right}: ELBO as a function of the number of topics $K$. }
\label{fig:ELBO1}
\end{center}
\end{figure}

In all our experiments we observe that frequentist methods---\texttt{G-OEM}, \texttt{V-OEM++} and \texttt{SPLDA}---outperform variational Bayesian methods---\texttt{OLDA} and \texttt{SVB}. As described in Section~\ref{sec:VB}, variational Bayesian methods maximize the ELBO, which makes additional strong independence assumptions and here leads to poor results. For example, as the number $K$ of topics increases, the log-likelihood goes down for some datasets. In order to investigate if this is an issue of slow convergence, we show on Figure~\ref{fig:Kcomp_P100} (dotted black line) that running $P=100$ internal updates in \texttt{OLDA} to get a finer estimate of the ELBO for each document may deteriorate the performance. Moreover, Figure~\ref{fig:ELBO1} presents the evolution of the ELBO, which does always increase when $K$ increases, showing that the online methods do optimize correctly the ELBO (while not improving the true log-likelihood). See Appendix~\ref{app:ELBO} for additional results on the convergence of the ELBO.
The results are mitigated for the hybrid Bayesian method \texttt{VarGibbs}. The performance of this method is either similar to \texttt{G-OEM} and \texttt{V-OEM++} or significantly worse than both \texttt{G-OEM} and \texttt{V-OEM++}.
 
\subsubsection{Small Step-sizes vs. Large Step-sizes}
\texttt{SPLDA} is also a variational method which is equivalent to \texttt{V-OEM++}, but with a step-size $1/t$, which is often slower than bigger step-sizes~\citep{incrementalMM2014}, which we do observe---see Appendix~\ref{app:rho} for a further analysis on the effect of the choice of step-sizes as $1/i^\kappa$ on \texttt{G-OEM}. Note that we run all the methods on a fixed (finite) number of observations. If we were to extend to infinite datasets, the difference between the step-sizes should be the speed of convergence. However, even if the number of observations is large, the gap between the step-sizes is still significant to justify the use of $1/\sqrt{t}$ for the step-size. Indeed, when considering large datasets, the contribution of each iteration at the end of the pass over the data is squeezed by the step-size in~$1/t$. When the number of observations is large enough to prevent the use of batch algorithms
but still insufficient for an online algorithm to converge in one pass, a possible solution could be to consider constant step-sizes in order to converge even faster to a local maxima. As proposed, we do not have any guarantee for our methods to converge with constant step-sizes, but previous works have shown the benefits of using constant step-sizes under certain assumptions (e.g., \citet{NewtonSteps2013})

\subsection{Empirical Analysis}
\label{app:topics}
{
In this section we provide a qualitative empirical analysis on the topics extracted with the different methods. We note this is clearly a subjective analysis but it stresses the benefits of a ``better'' inference mechanism in terms of log-likelihood \citep{TeaLeaf2009}. Examples of eight topics extracted with \texttt{G-OEM} and \texttt{OLDA} on the IMDB dataset of movie reviews are presented in Table~\ref{tab:topics} page~\pageref{tab:topics}. 

We first compute the KL divergence between the $K=128$ topics extracted with \texttt{G-OEM} and the $K=128$ topics extracted with \texttt{OLDA}. We run the Hungarian algorithm on the resulting distance matrix to assign each topic extracted with \texttt{G-OEM} to a single topic of \texttt{OLDA}.
We choose manually eight topics extracted with \texttt{G-OEM} that are representative of the usual behavior, and display the eight corresponding topics of \texttt{OLDA} assigned with the above method.

We observe that the topics extracted with \texttt{G-OEM} are more consistent than topics extracted with \texttt{OLDA}: topics of \texttt{G-OEM} precisely describe only one aspect of the reviews while the topics of \texttt{OLDA} tend to mix several aspects in each topic. For instance, the words of topic~1 extracted with \texttt{G-OEM} are related to \textit{horror} movies. The words of the corresponding topic extracted with \texttt{OLDA} mix \textit{horror} movies---e.g., \textit{horror, scary}---and \textit{ghost} movies---e.g., \textit{ghost, haunt}. In this \texttt{OLDA} topic 1, we can also observe less relevant words, like \textit{effective, mysterious}, which are not directly linked with \textit{horror} and \textit{ghost} vocabularies. We can make the same remarks with topic~2 and topic~3, respectively related to \textit{comedy} movies and \textit{romantic comedy} movies. In topic~2 extracted with \texttt{G-OEM}, the least related words to \textit{comedy} are names of characters/actors---i.e., \textit{steve} and \textit{seth}---while the words not related to \textit{comedy} in topic~25 of \texttt{OLDA} are more general, belonging to a different lexical field---e.g., \textit{sport, site, progress, brave, definition}. In topic~3 of \texttt{G-OEM}, all the presented words are related to \textit{romantic comedy} while in topic~3 of \texttt{OLDA}, the words \textit{old, hard} and \textit{review} are not related to this genre.

We also observe that \texttt{G-OEM} extracts strongly ``qualitative'' topics---topic~4 and topic~5---which is not done with \texttt{OLDA}. Indeed, it is difficult to group the top words of topic~4 or topic~5 of \texttt{OLDA} in the same lexical field. Except \textit{dialogue} and \textit{suppose}, all the top words of topic~4 of \texttt{G-OEM} are negative words. These two words may appear in a lot of negative sentences, leading to a high weight in this topic. In topic~5 of \texttt{G-OEM}, the words \textit{absolutely} and \textit{visual} are non strictly positive words while the thirteen other words in this topic convey a positive opinion. The word \textit{absolutely} is an adverb much more employed in positive sentences than negative or neutral sentences, which can explain its high weight in topic~5.

The topic~6 of both \texttt{G-OEM} and \texttt{OLDA} can be considered as a ``junk'' topic, as for both method, most of its top words are contractions of modal verbs or frequent words---e.g., \textit{didn't, isn't, wait, bad}. The contractions are not filtered when removing the stop words as they are not included in the list of words removed\footnote{See NLTK toolbox \citep{NLTK} for the exhaustive list of stop words.}.

For both \texttt{G-OEM} and \texttt{OLDA}, the top words of topic~7 are general words about movies. These words are usually employed to describe a movie as a whole---e.g., \textit{narrative, filmmaker}.

Finally, the top words of topic~8 of \texttt{G-OEM} are related to the situation of the scenes. We could not find such topic in the other presented methods and we can see that the top words of topic~8 of \texttt{OLDA}---supposedly close to topic~8 of \texttt{G-OEM}---are related to \textit{family} movies. Each word of topic~8 of \texttt{G-OEM}---except \textit{group} and \textit{beautiful}---are related to a \textit{spatial location}, and may help answer the question ``\textit{where does the scene take place?}''.
 
}

\begin{sidewaystable*}[!t]
\caption{Comparison of topics extracted on IMDB dataset, $K=128$---15 top words of eight topics extracted with \texttt{G-OEM} and \texttt{OLDA}.}
\label{tab:topics}
\vskip 0.15in
\begin{center}
\begin{small}
\begin{sc}
\begin{tabular}{cllllllll}
\hline
\multicolumn{9}{c}{\texttt{G-OEM}}\\
\hline
\# & Topic 1 & Topic 2 & Topic 3 & Topic 4 & Topic 5 & Topic 6 & Topic 7 & Topic 8 \\
\hline
1 & violence & comedy & romantic & bad & brilliant & didn't & narrative & town \\
2 & violent & funny & comedy & worst & perfect & i've & ultimately & local \\
3 & disturbing & laugh & love & waste & beautiful & wasn't & cinematic & mountain \\
4 & brutal & joke & romance & boring & masterpiece & isn't & approach & village \\
5 & murder & hilarious & funny & awful & amazing & we're & protagonist & location \\
6 & graphic & comic & charming & poor & superb & i'll & seemingly & road \\
7 & killer & comedic & chemistry & dialogue & stunning & couldn't & nature & journey \\
8 & torture & steve & sweet & worse & wonderful & wouldn't & tone & group \\
9 & victim & amusing & enjoy & dull & absolutely & pretty & filmmaker & travel \\
10 & rape & fun & heart & fail & best & bad & whose & country \\
11 & kill & gag & nice & mess & incredible & haven't & craft & landscape \\
12 & horror & seth & charm & ridiculous & brilliantly & guess & contemporary & land \\
13 & bloody & funniest & great & suppose & beautifully & aren't & manner & beautiful \\
14 & revenge & cameo & fun & terrible & visual & enjoy & serve & area \\
15 & blood & situation & wonderful & unfortunately & perfectly & review & material & trip \\
\hline

\hline
\multicolumn{9}{c}{\texttt{OLDA}}\\
\hline
\# & Topic 1 & Topic 2 & Topic 3 & Topic 4 & Topic 5 & Topic 6 & Topic 7 & Topic 8 \\
\hline
1 & horror & hilarious & love & bad & great & wrong & visual & young \\
2 & night & romance & enjoy & action & best & didn't & focus & family \\
3 & dead & clever & pretty & interesting & star & i've & reality & child \\
4 & twist & smart & old & original & long & wait & difficult & father \\
5 & scary & intriguing & funny & far & john & catch & filmmaker & son \\
6 & effective & comedic & comedy & special & excellent & exactly & image & age \\
7 & mysterious & funniest & fun & fight & classic & wasn't & narrative & whose \\
8 & bloody & progress & hard & hero & beautiful & huge & intelligent & tale \\
9 & ghost & sport & perfect & entire & drama & i'll & accept & discover \\
10 & haunt & site & laugh & half & wonderful & choice & impression & easily \\
11 & fear & brave & entertaining & save & michael & etc & extreme & dream \\
12 & evil & dreadful & worth & dialogue & heart & seriously & central & introduce \\
13 & nightmare & shoulder & nice & full & forget & notice & maintain & marry \\
14 & gory & gimmick & favorite & violence & robert & ridiculous & deny & raise \\
15 & mask & definition & review & example & early & answer & nail & rule \\
\hline

\end{tabular}
\end{sc}
\end{small}
\end{center}
\vskip -0.1in
\end{sidewaystable*}

\subsection{Results on HDP}
\begin{figure}
\begin{center}
\subfigure[IMDB.]{\includegraphics[width=0.48\columnwidth]{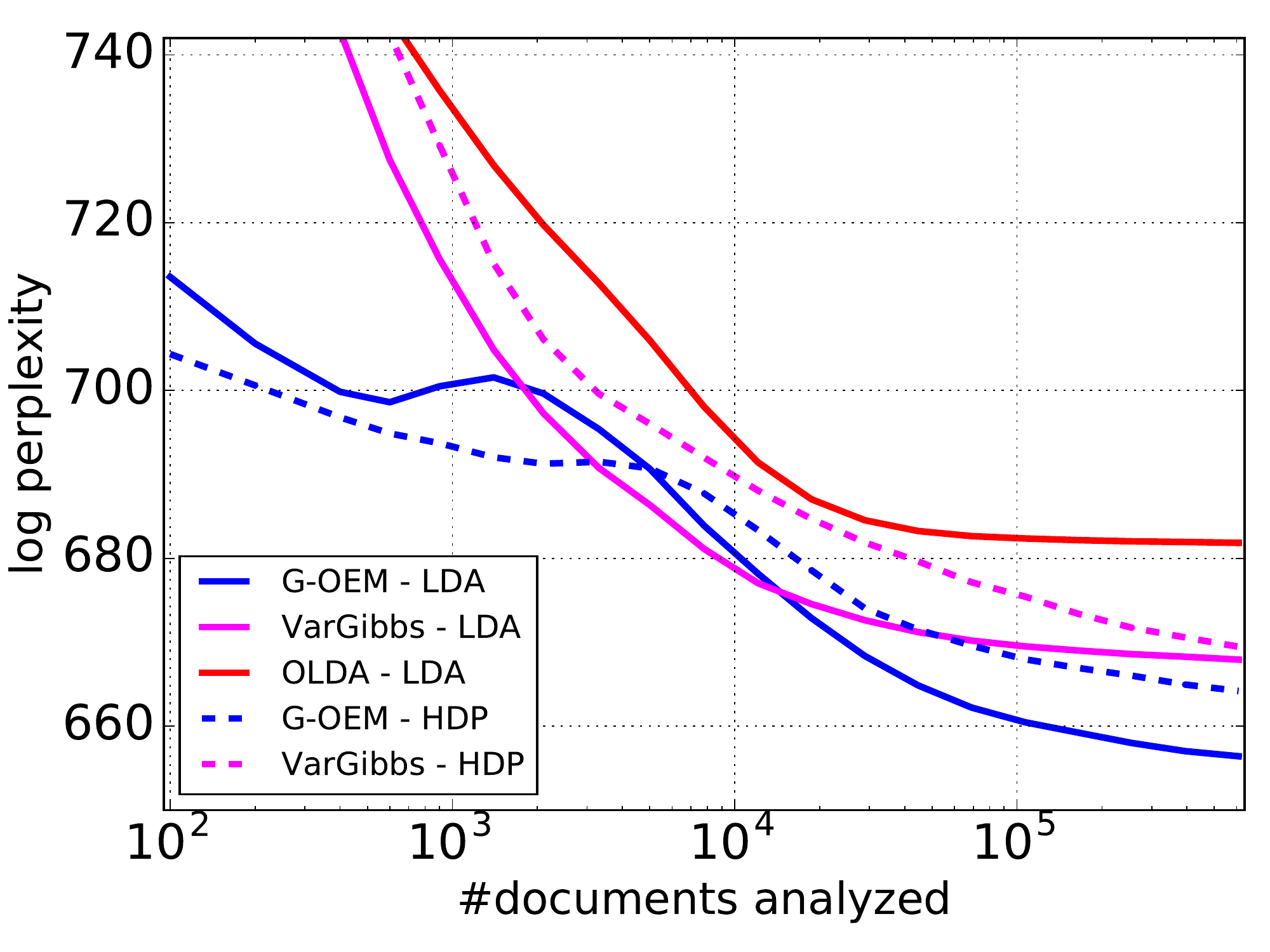}}
\subfigure[Wikipedia]{\includegraphics[width=0.48\columnwidth]{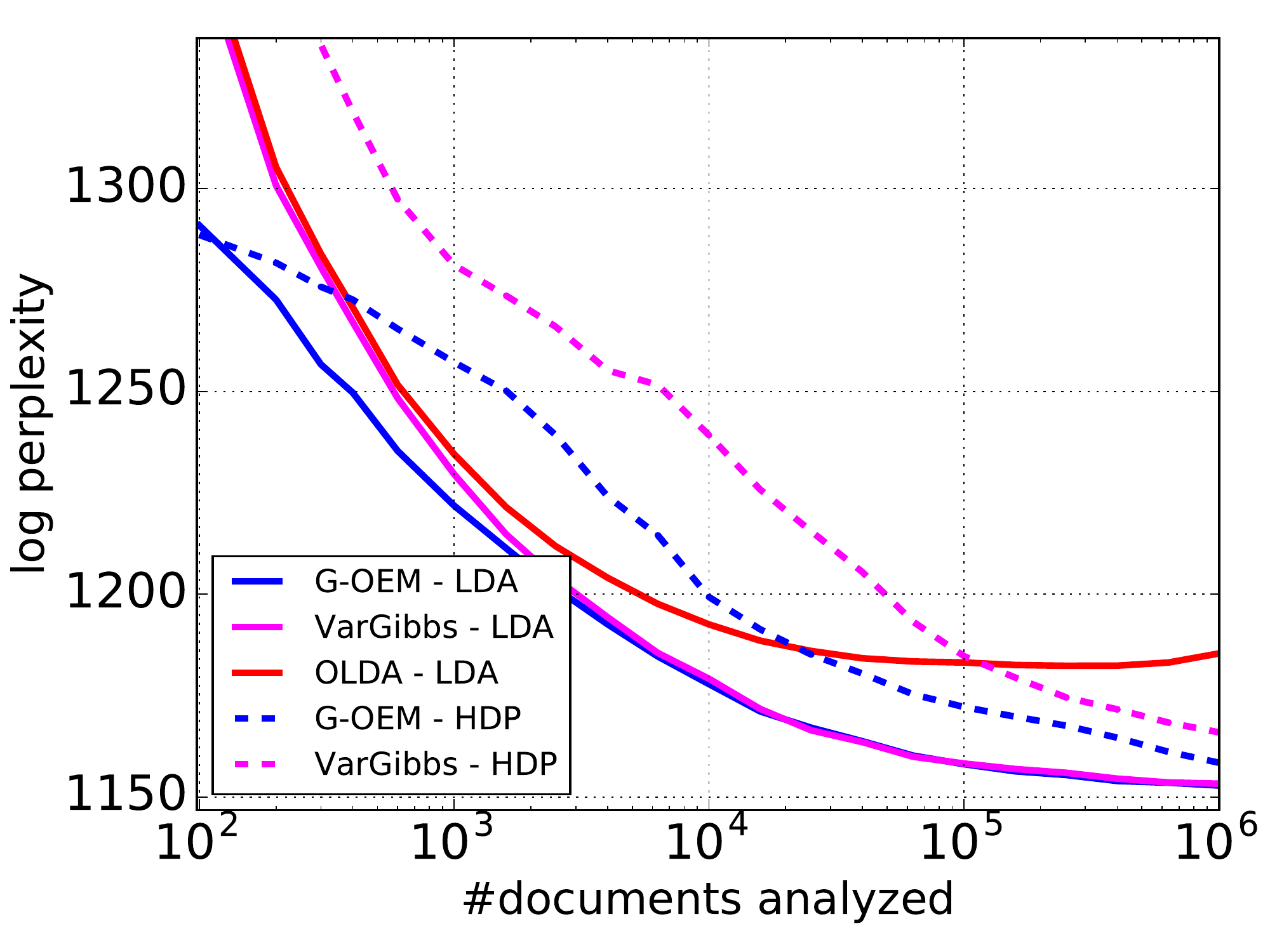}}
\caption{Perplexity through iterations on different test sets with \texttt{G-OEM} and \texttt{VarGibbs} applied to both LDA and HDP. Best seen in color.}
\label{fig:HDP_novar}
\end{center}
\vskip -0.8cm
\end{figure}

For the HDP model, we compare our \texttt{G-OEM} method to the Bayesian \texttt{VarGibbs} \citep{BNP_HDP} method. We set the initial number of topics to $T=2$. We present in Figure~\ref{fig:HDP_novar} results obtained with \texttt{G-OEM} and \texttt{VarGibbs} applied to both LDA and HDP. Results with error bars are presented in Appendix~\ref{app:hdp}. For both LDA and HDP, \texttt{G-OEM} outperforms the Bayesian method \texttt{VarGibbs}.

\section{Conclusion}
We have developed an online inference scheme to handle intractable conditional distributions of latent variables, with a proper use of local Gibbs sampling within online EM, that leads to significant improvements over variational methods and Bayesian estimation procedures. Note that all methods for the same problem are similar (in fact a few characters away from each other); ours is based on a proper stochastic approximation maximum likelihood framework and is empirically the most robust. 
It would be interesting to explore distributed large-scale settings~\citep{StreamingVB,GPUGibbsLDA2009,StreamingGibbs2016} and potentially larger (e.g., constant) step-sizes that have proved efficient in supervised learning~\citep{NewtonSteps2013}.

\vskip 0.5cm
\acks{
We would like to thank David Blei, Olivier Capp\'e, Nicolas Flammarion and John Canny for helpful discussions related to this work.}

\clearpage

\appendix

\section{Gibbs/Variational Online EM Analysis}
\label{app:Gibbssettings}
In this section we evaluate the proposed methods \texttt{G-OEM} and \texttt{V-OEM} with different settings in terms of step-sizes, averaging outputs and boosting internal updates.
\subsection{Effect of Inference Boosting on \texttt{G-OEM} and \texttt{V-OEM}}
\label{app:boostinf}
\begin{figure}[!h]
\begin{center}
\subfigure[Synthetic dataset.]{\includegraphics[width=0.49\columnwidth]{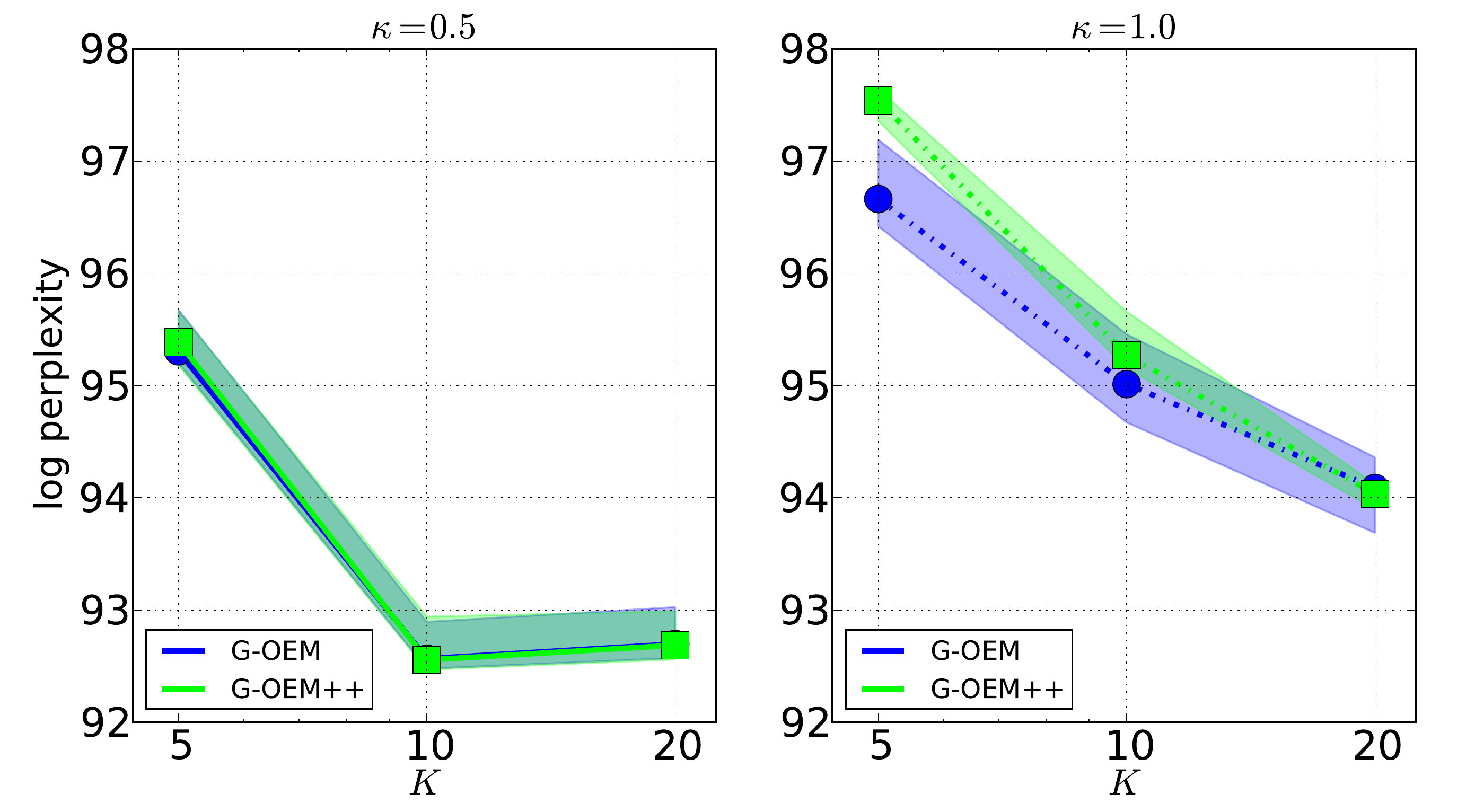}}
\subfigure[IMDB.]{\includegraphics[width=0.49\columnwidth]{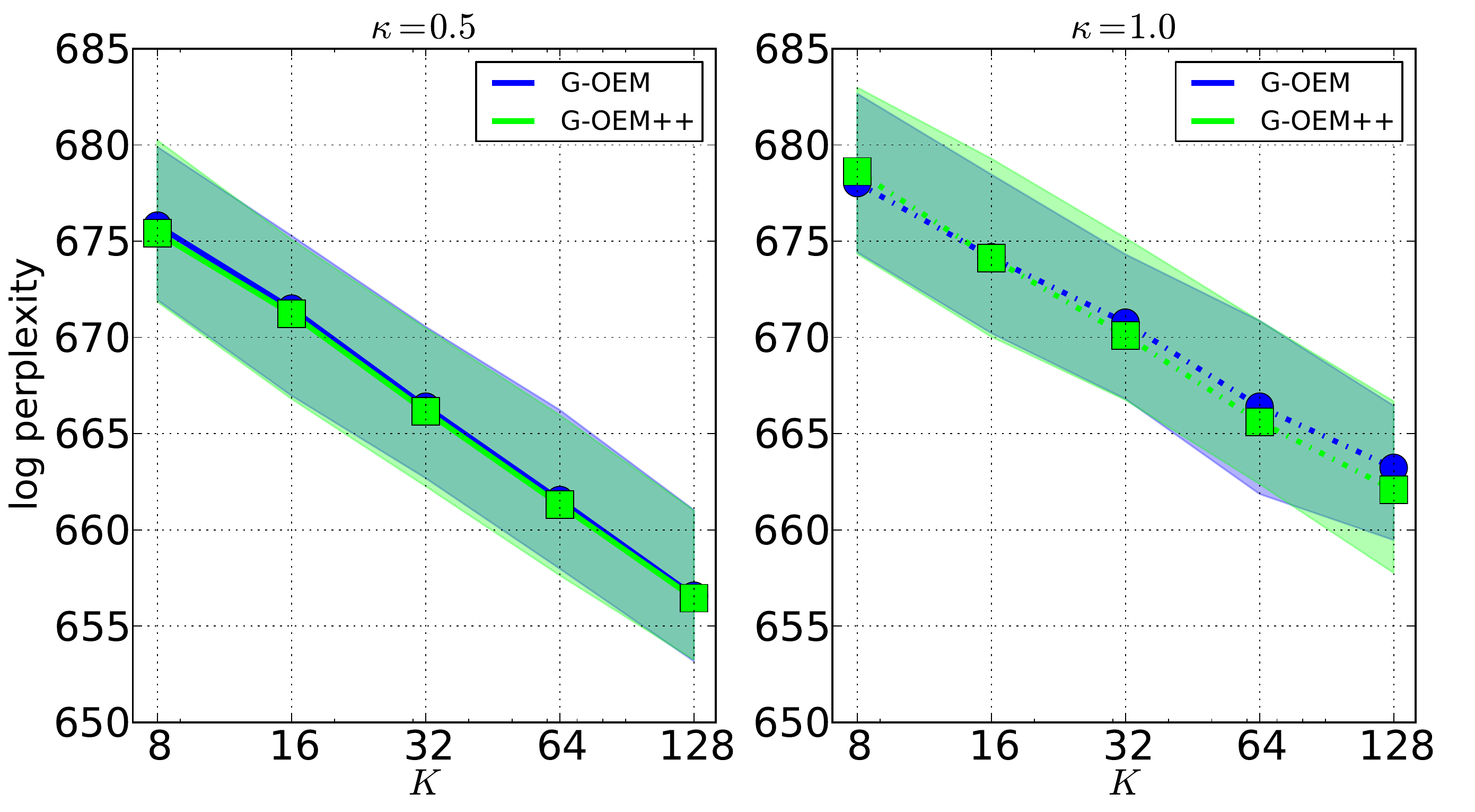}}
\subfigure[Wikipedia.]{\includegraphics[width=0.49\columnwidth]{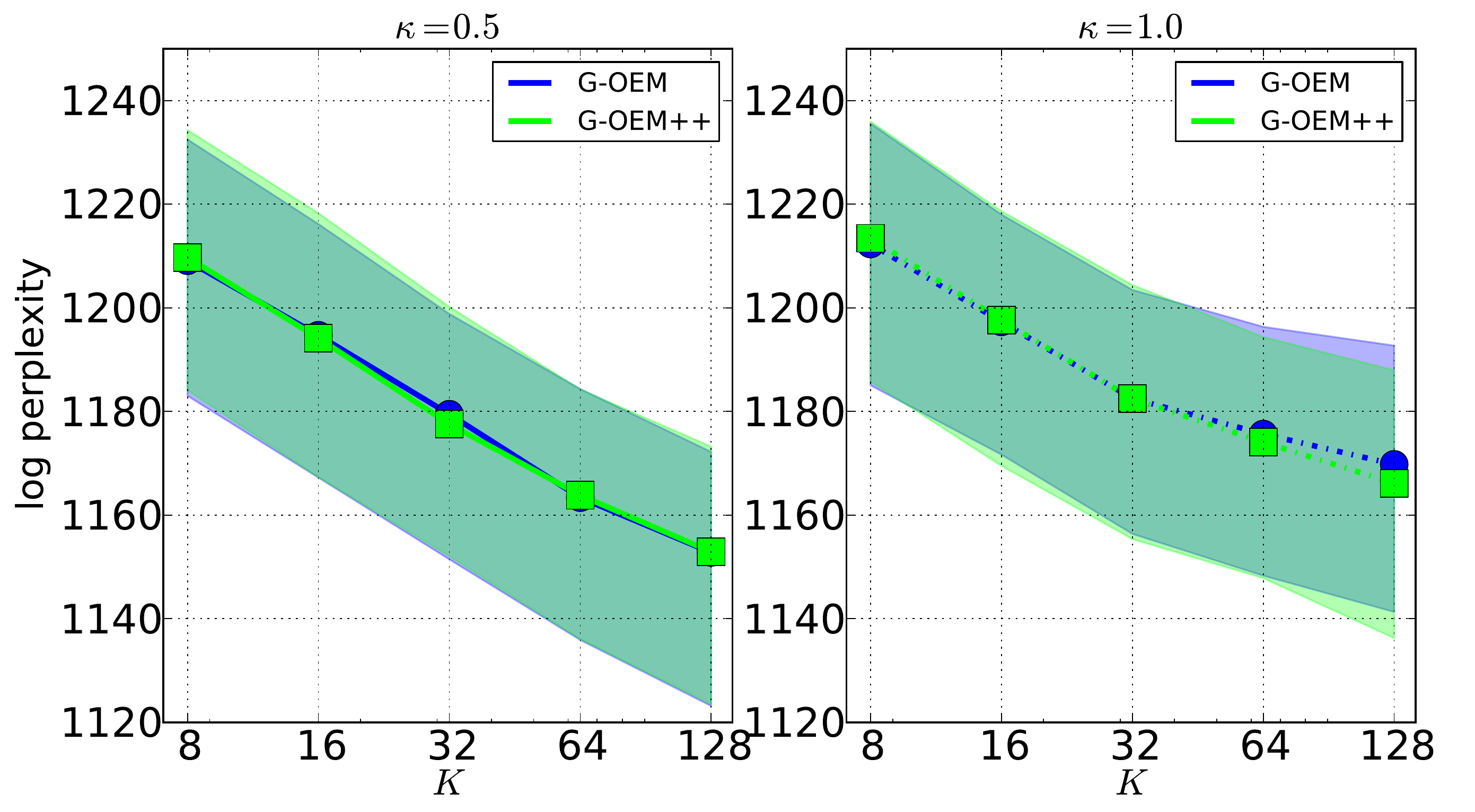}}
\subfigure[New York Times.]{\includegraphics[width=0.49\columnwidth]{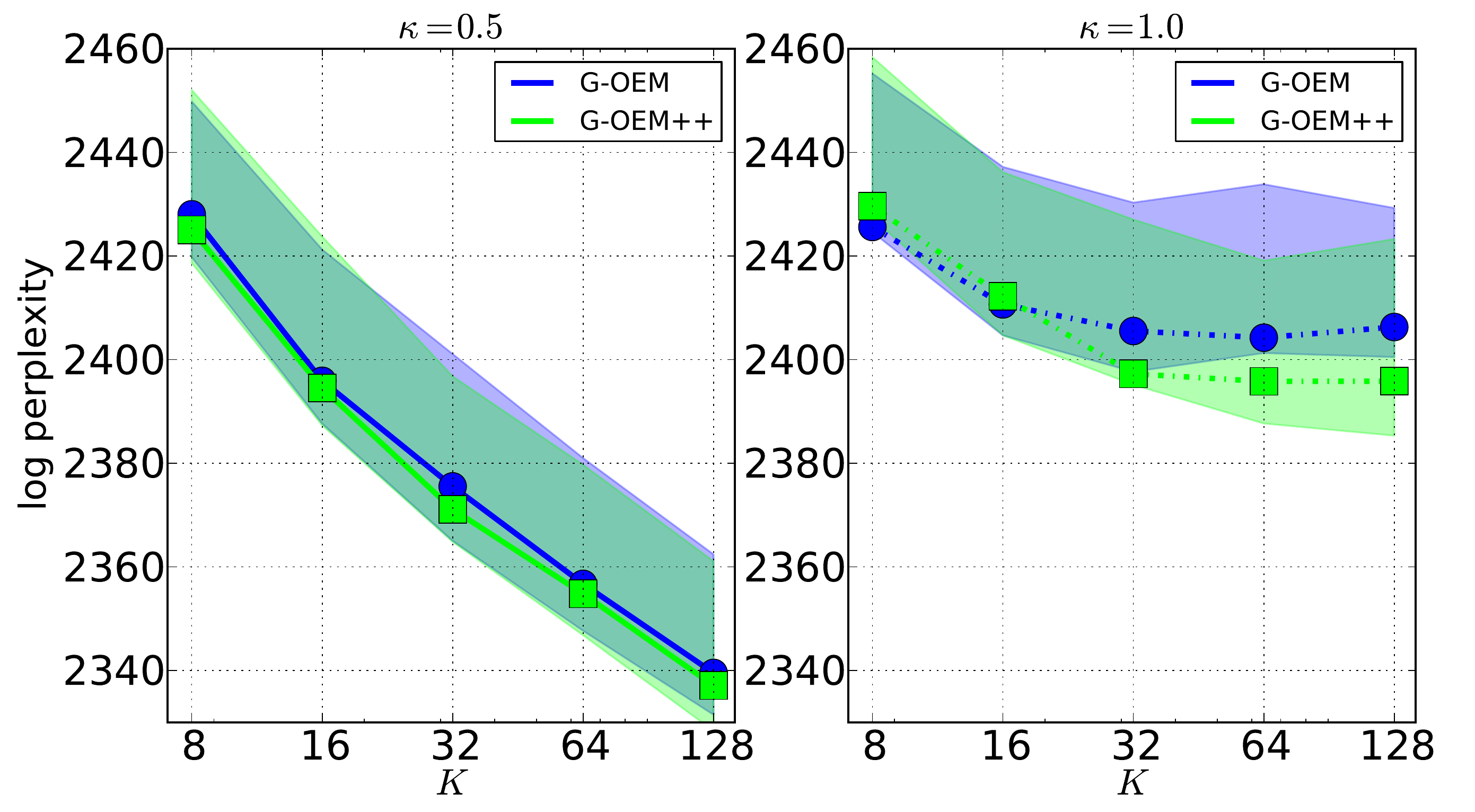}}\\
\subfigure[Pubmed.]{\includegraphics[width=0.49\columnwidth]{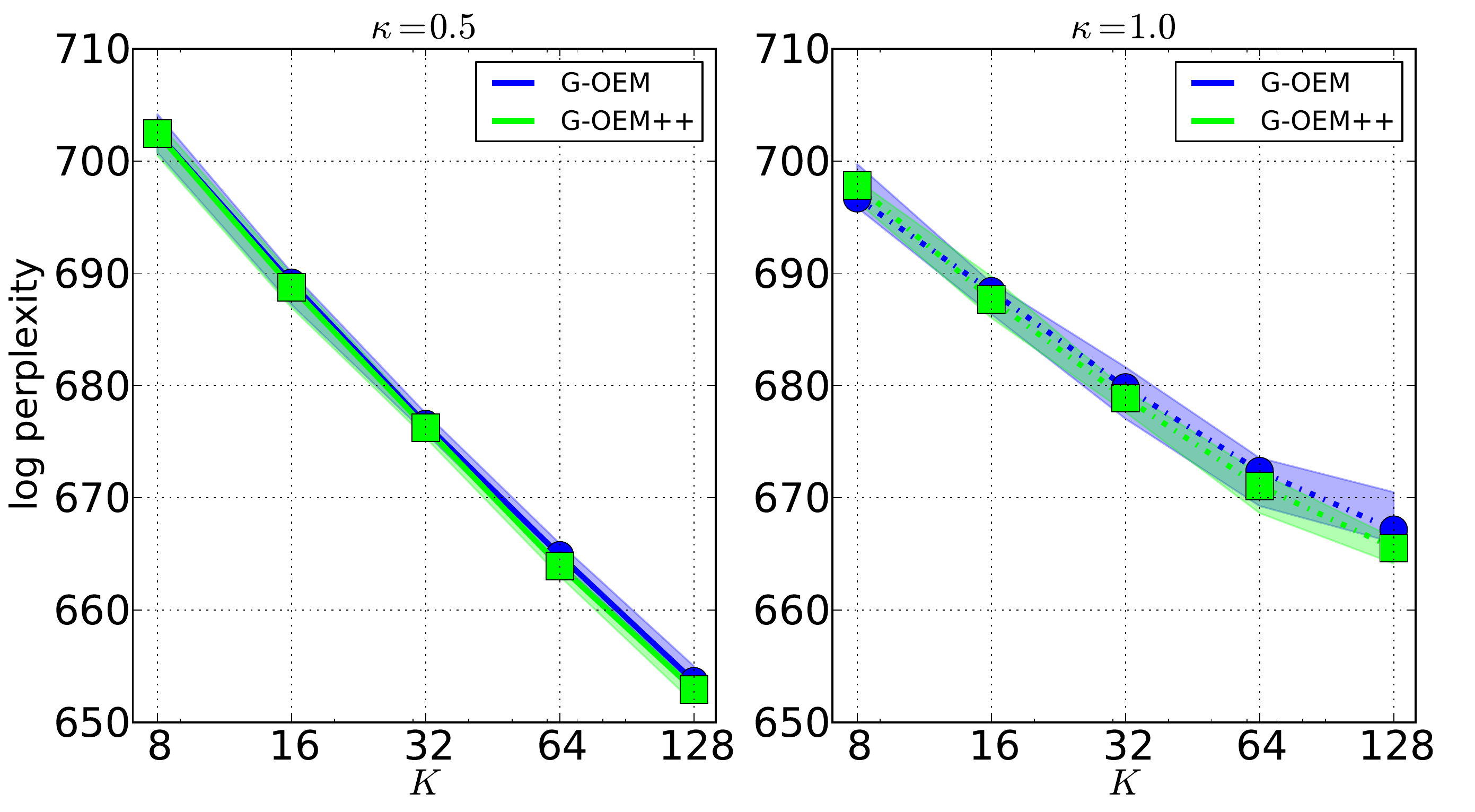}}
\subfigure[Amazon movies.]{\includegraphics[width=0.49\columnwidth]{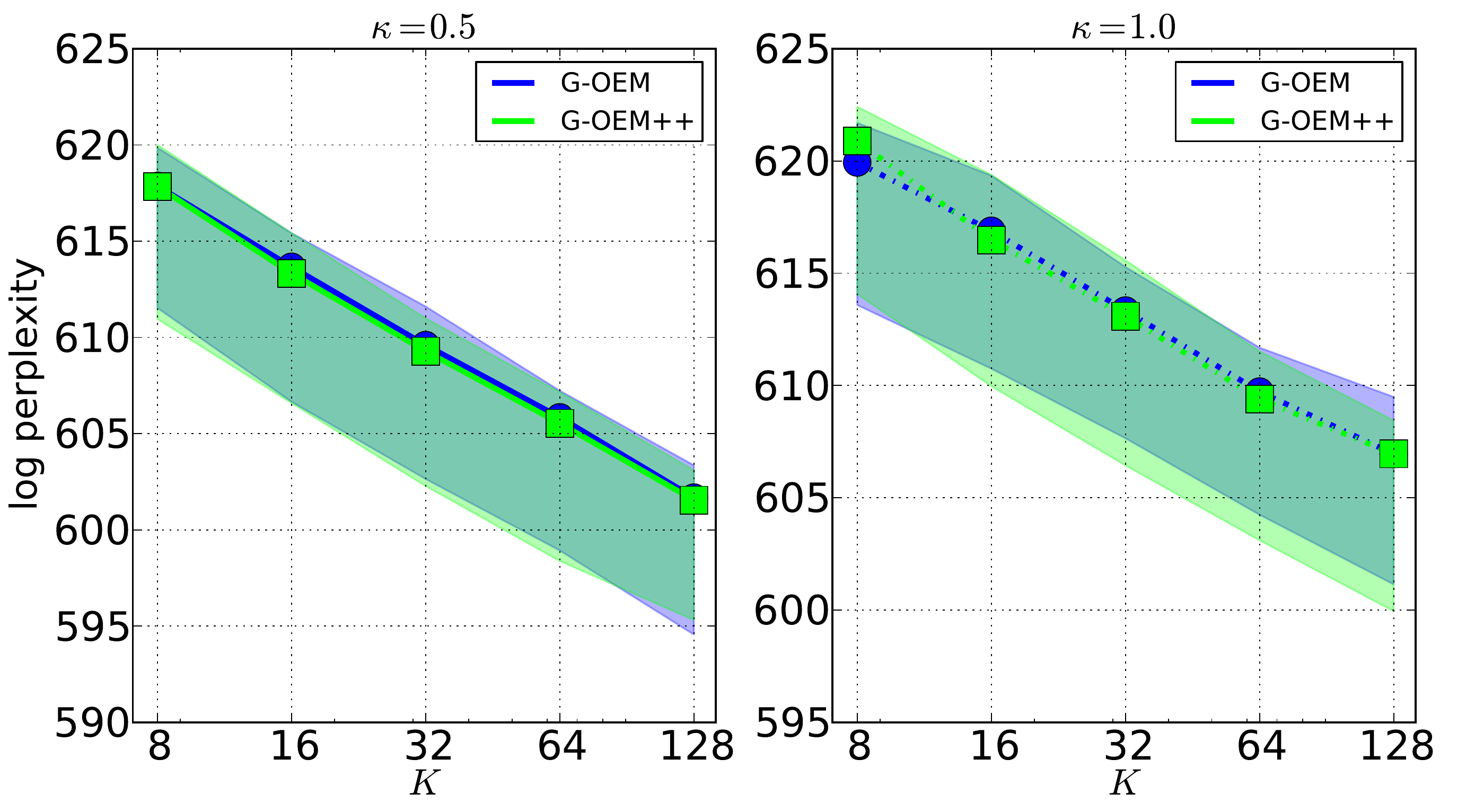}}
\caption{\texttt{G-OEM}. Perplexity on different test sets as a function of the number of topics $K$ for regular EM and boosted EM (\texttt{++}).  We observe that for almost all datasets, there is no significant improvement when boosting the inference. Our interpretation is that each Gibbs sample is noisy and does not provide a stable boost. Best seen in color.}
\label{fig:BoostGOEM}
\end{center}
\vskip -0.2in
\end{figure}

\begin{figure}[!h]
\begin{center}
\subfigure[Synthetic dataset.]{\includegraphics[width=0.49\columnwidth]{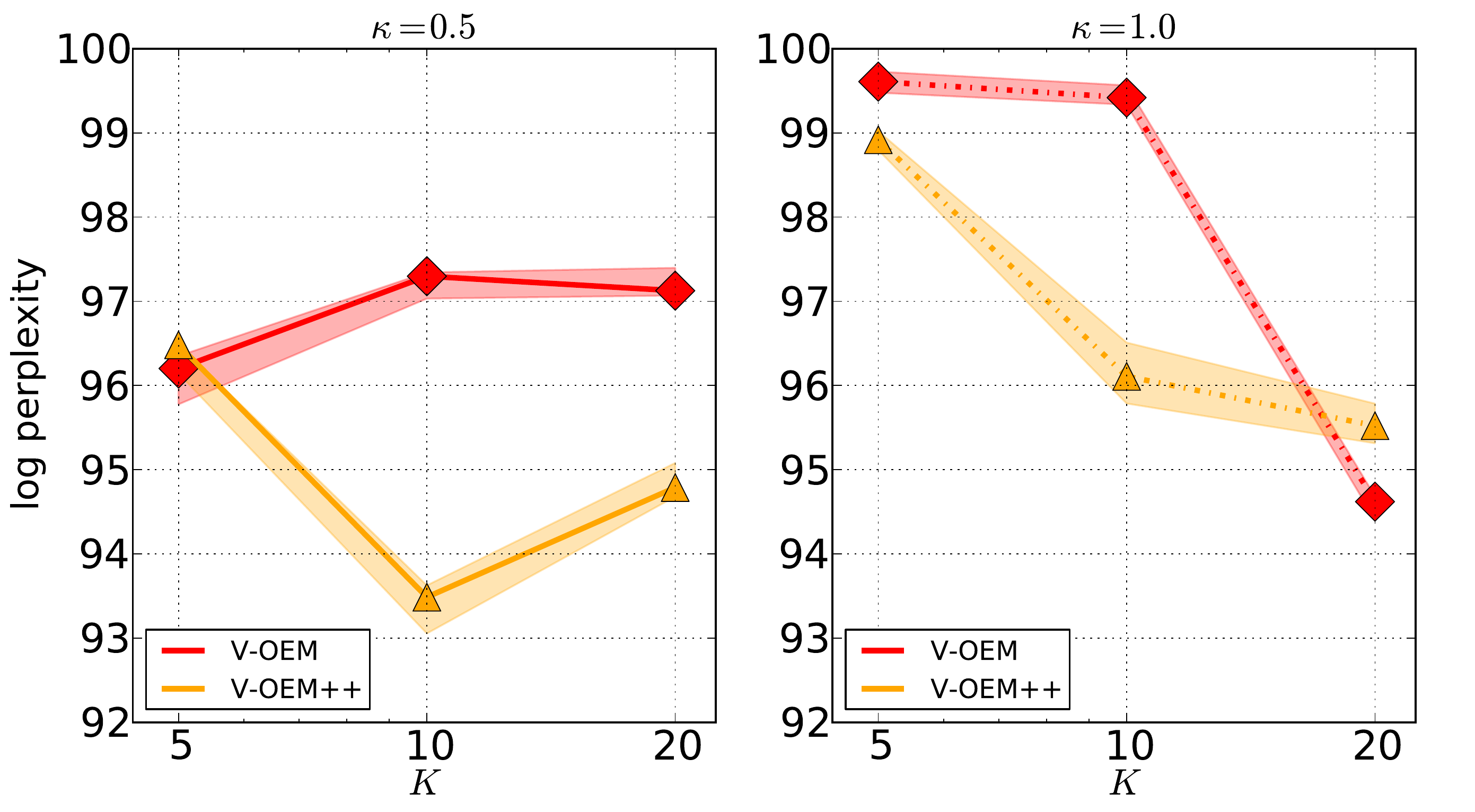}}
\subfigure[IMDB dataset.]{\includegraphics[width=0.49\columnwidth]{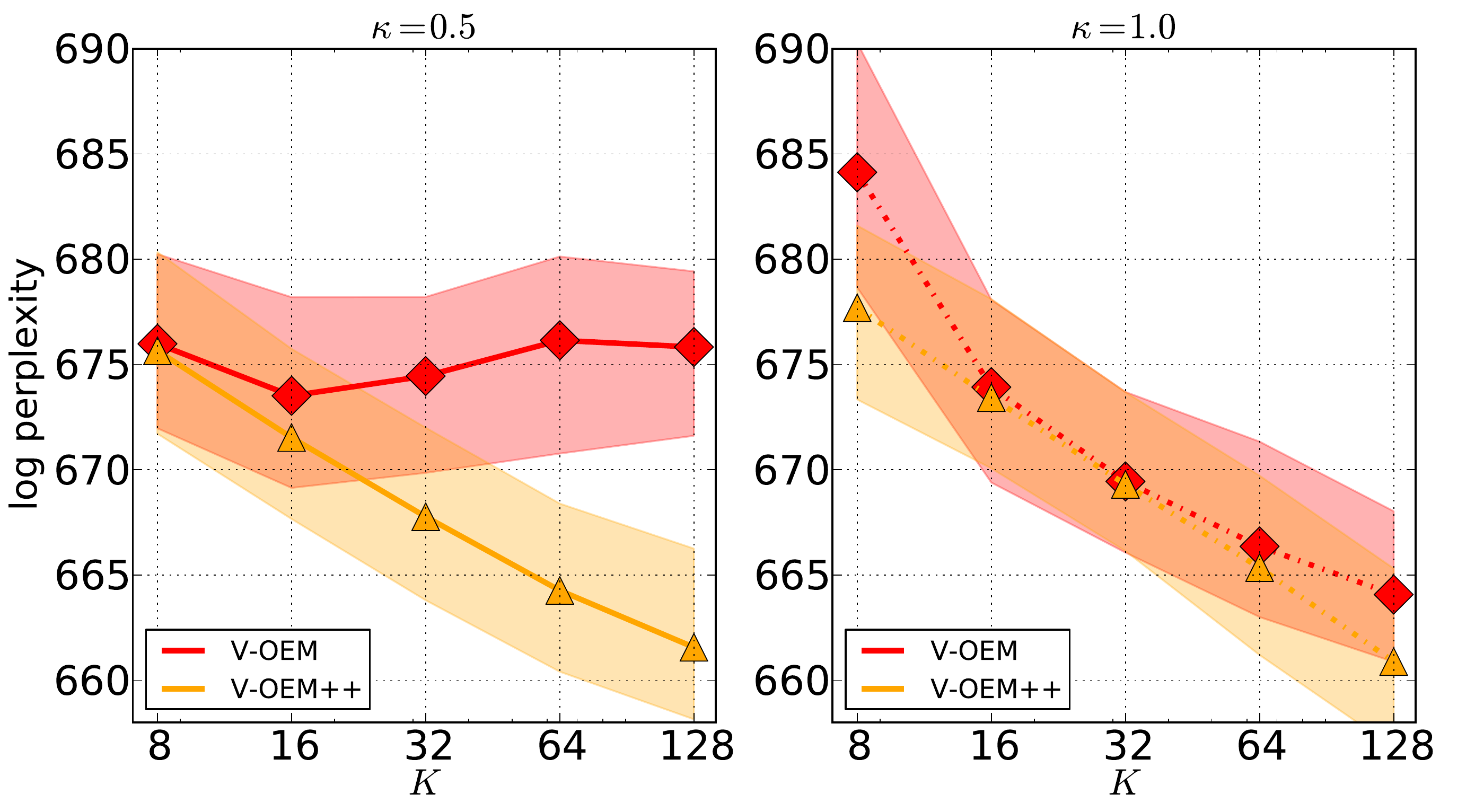}}
\subfigure[Wikipedia.]{\includegraphics[width=0.49\columnwidth]{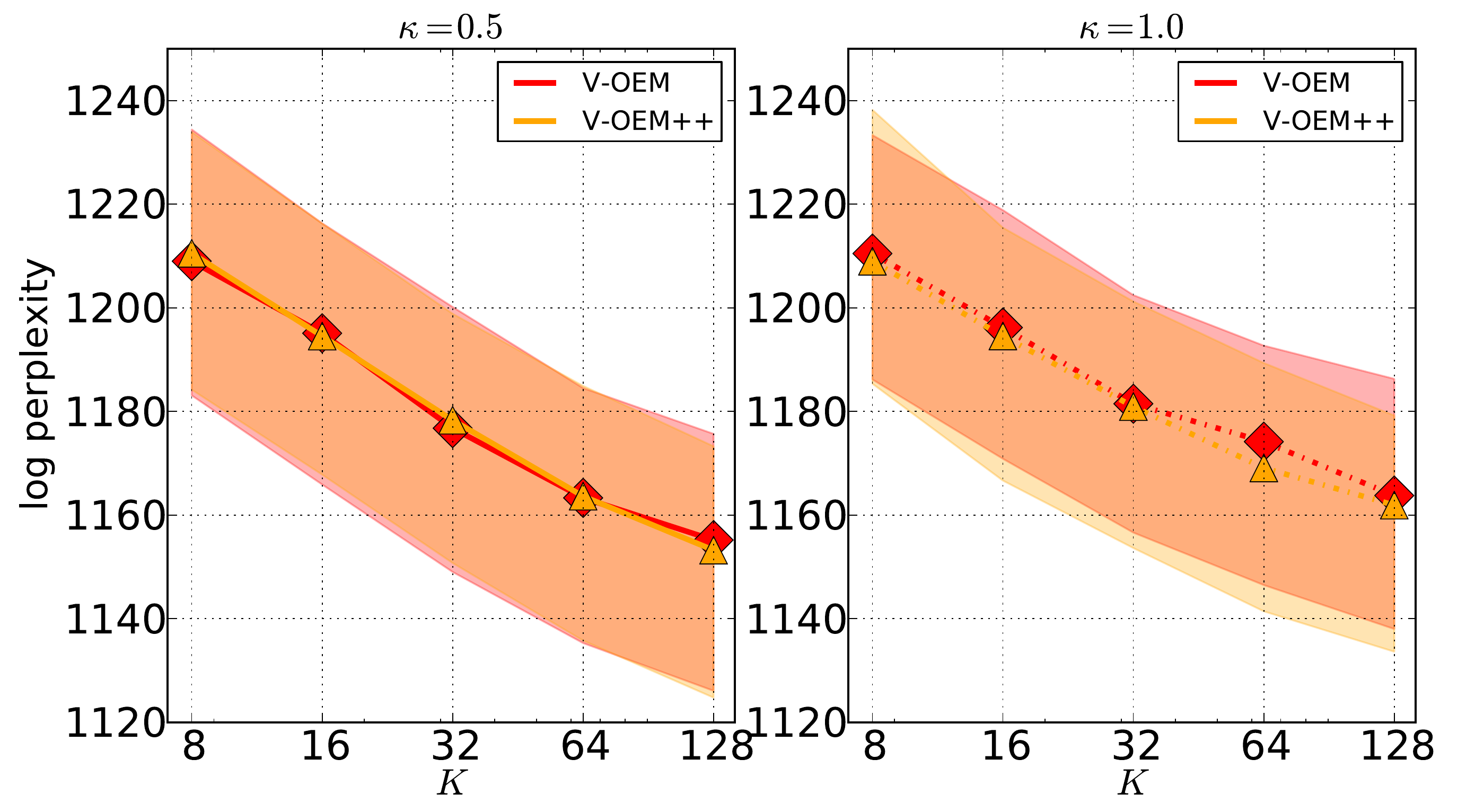}}
\subfigure[New York Times.]{\includegraphics[width=0.49\columnwidth]{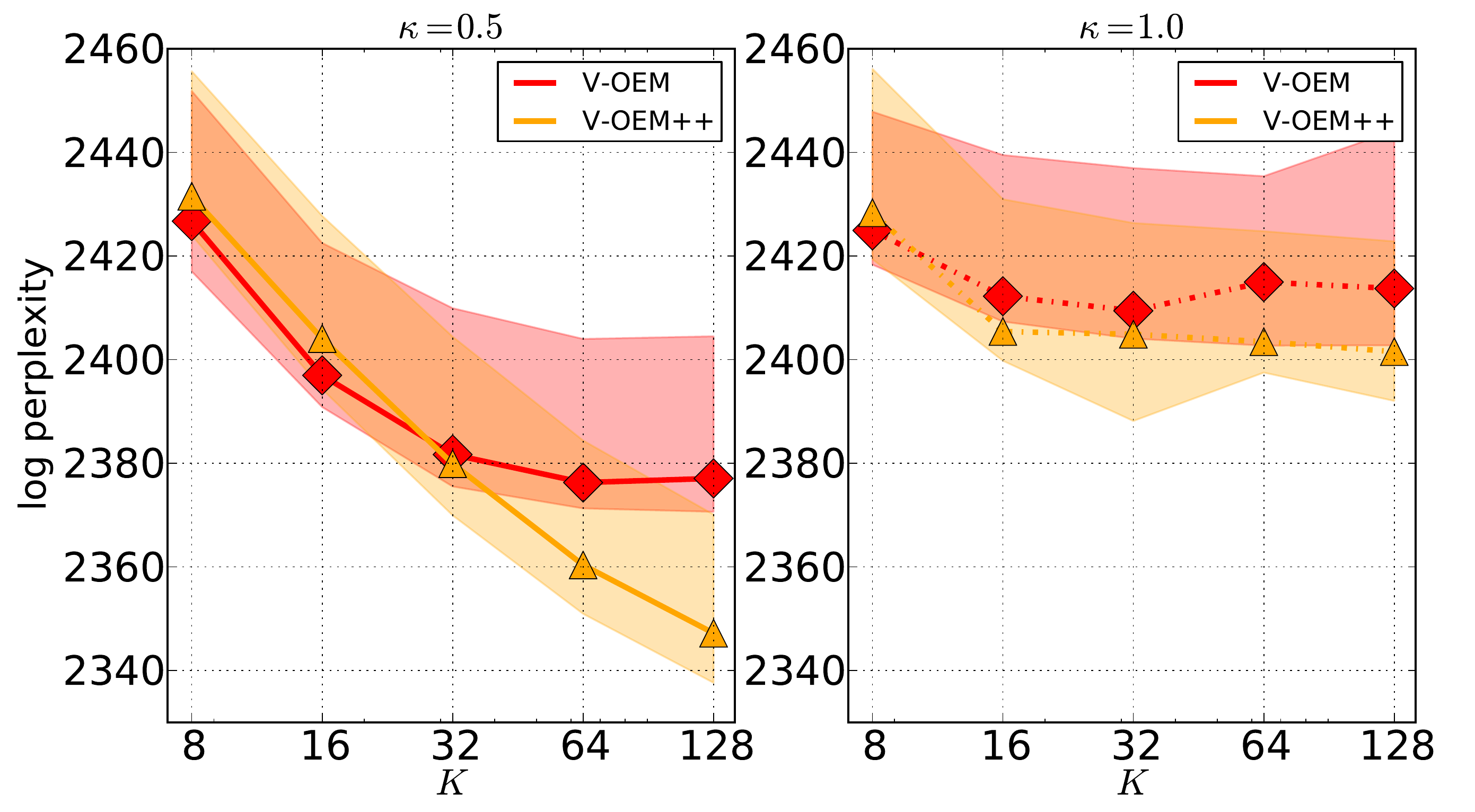}}\\
\subfigure[Pubmed.]{\includegraphics[width=0.49\columnwidth]{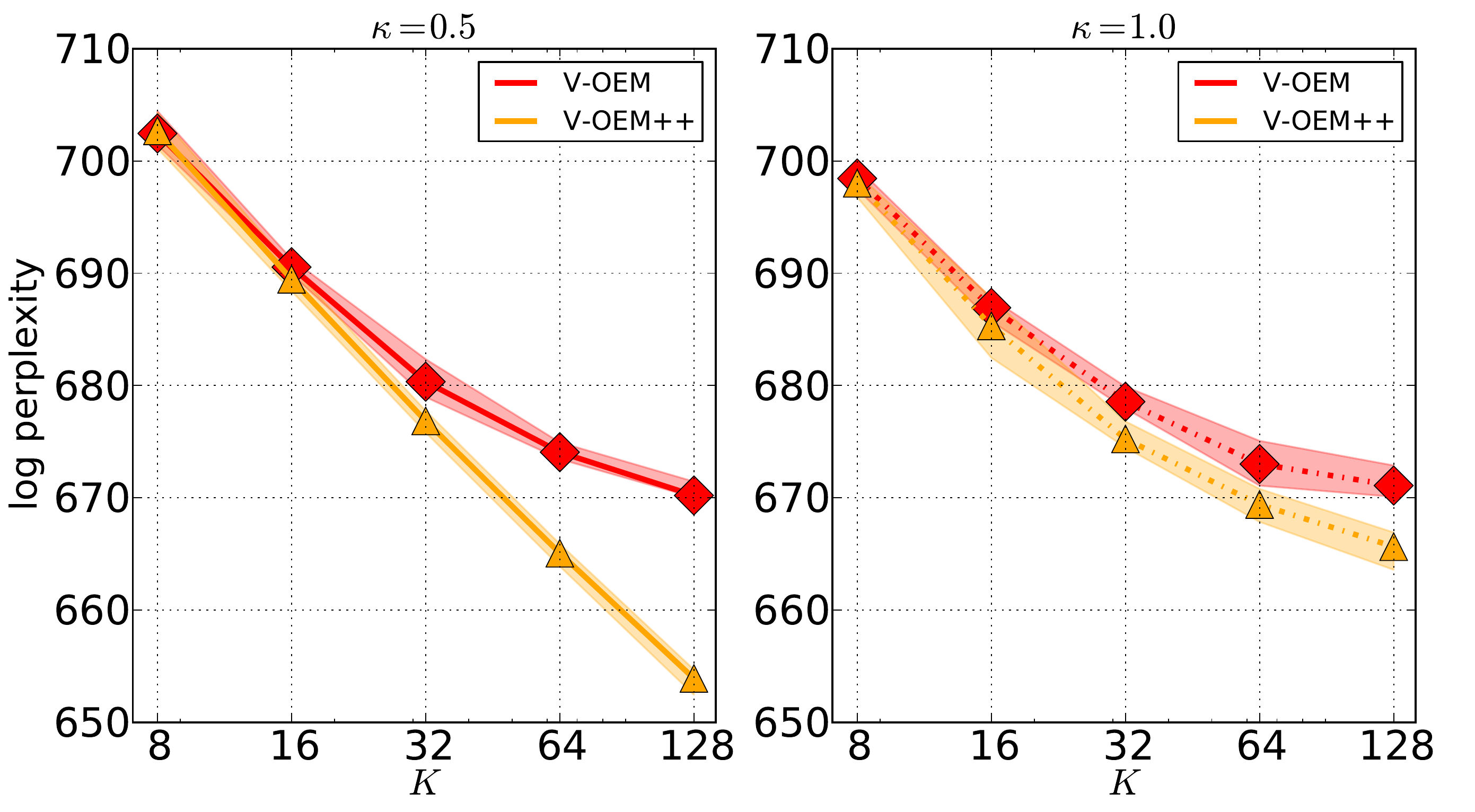}}
\subfigure[Amazon movies.]{\includegraphics[width=0.49\columnwidth]{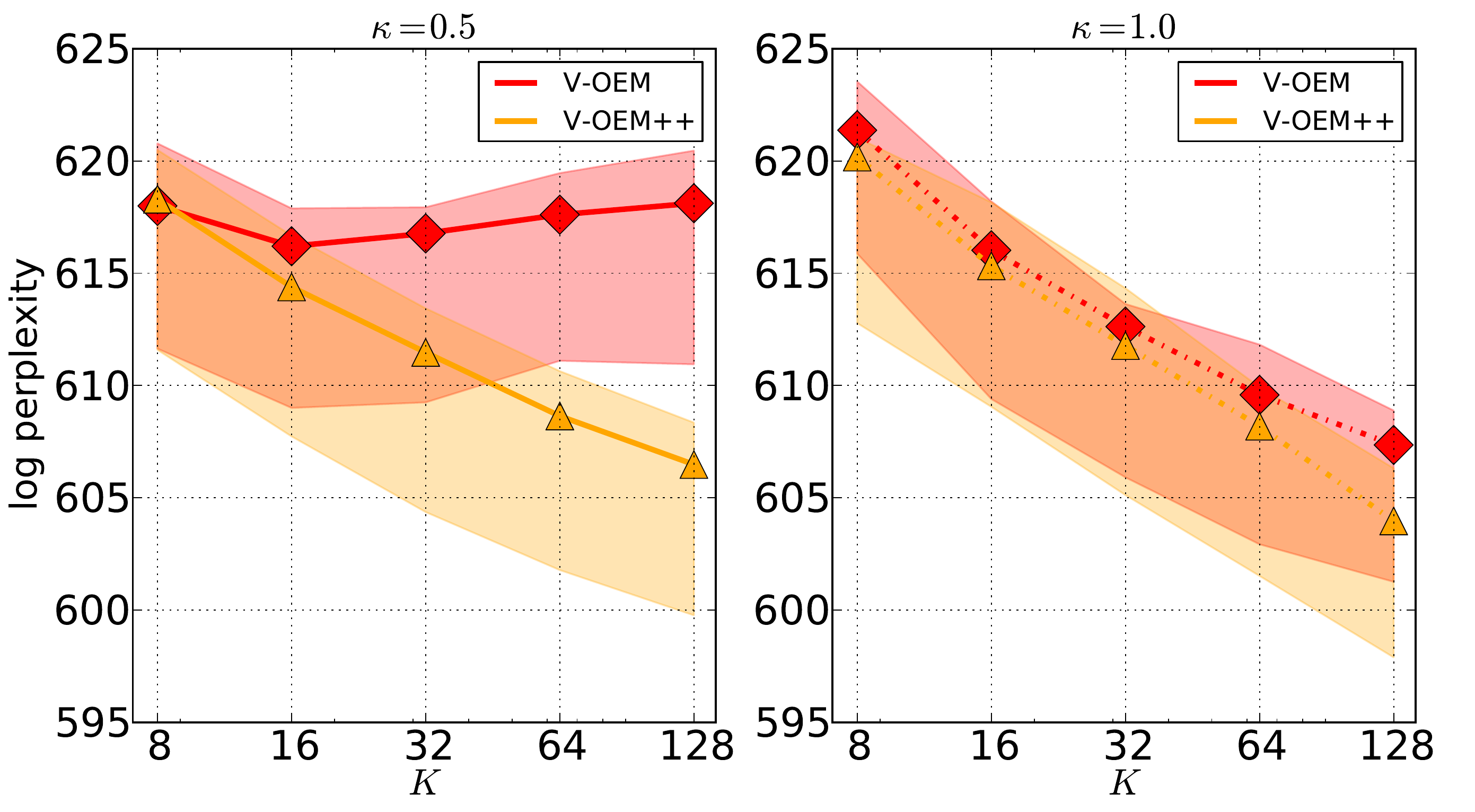}}
\caption{\texttt{V-OEM}. Perplexity on different test sets as a function of the number of topics $K$
for regular EM and boosted EM (\texttt{++}).  We observe that boosting inference improves significantly the results on all the datasets excepted on Wikipedia where \texttt{V-OEM} and \texttt{V-OEM++} have similar performances. The variational estimation of the posterior is finer and finer through iterations. When updating the parameters at each iteration of the posterior estimation, the inference is indeed boosted. Best seen in color.} 
\label{fig:BoostVOEM}
\end{center}
\vskip -0.2in
\end{figure}

The effect of the inference boost as described in Section~\ref{sec:boost} on  \texttt{G-OEM} and \texttt{V-OEM} with synthetic and IMDB datasets is presented in Figure~\ref{fig:BoostGOEM} and in Figure~\ref{fig:BoostVOEM}. It leads to a minor improvement for 
\texttt{G-OEM++} and a significant one for \texttt{V-OEM++}.

\subsection{Step-sizes and Averaging}
\label{app:rho}
We apply \texttt{G-OEM} with different stepsizes $\rho_i=\frac{1}{i^\kappa}$. Note that because we average sufficient statistics, there is no needed proportionality constants. We first compare the performance of the last iterate $\eta_N$ (without averaging) and the average of the iterates ${\bar{\eta}_N =\frac{1}{N}\sum_{i=0}^{N} \eta_i}$ (with averaging) for different values of $\kappa$. 

\begin{figure}[!h]
\begin{center}
\subfigure[\emph{With} averaging]{\includegraphics[width=0.49\columnwidth]{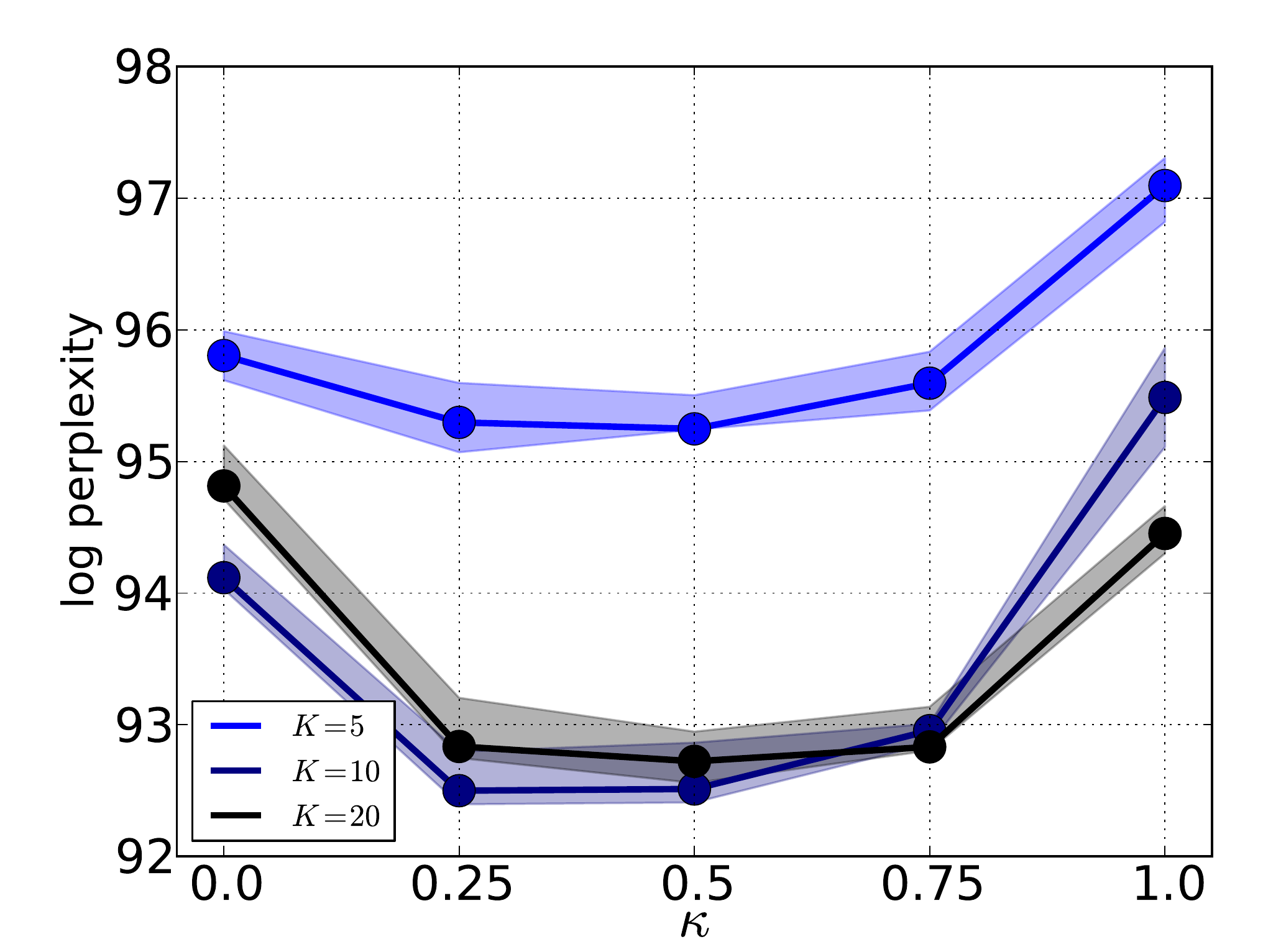}}
\subfigure[\emph{Without} averaging]{\includegraphics[width=0.49\columnwidth]{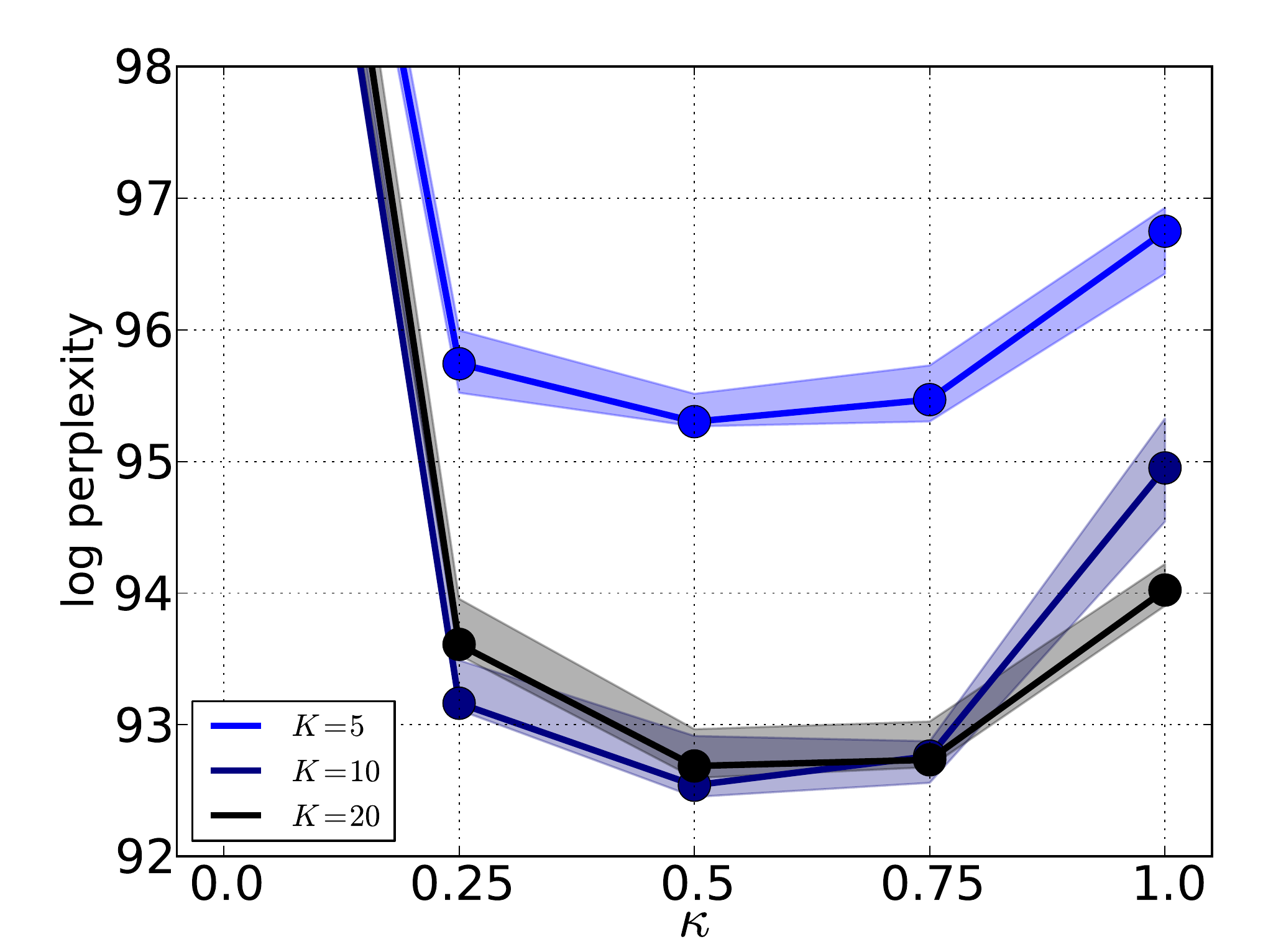}}
\caption{Dataset: synthetic. Perplexity on different test sets as a function of the exponent $\kappa$---the corresponding stepsize is $\rho_i=\frac{1}{i^\kappa}$---for \texttt{G-OEM} with averaging (left) and without averaging (right). The number of topics inferred $K$ goes from 5 (the lightest) to 20 (the darkest). Best seen in color.} 
\label{fig:Synth-Acomp}
\end{center}
\vskip -0.2in
\end{figure}

\begin{figure}[!h]
\begin{center}
\subfigure[\emph{With} averaging]{\includegraphics[width=0.49\columnwidth]{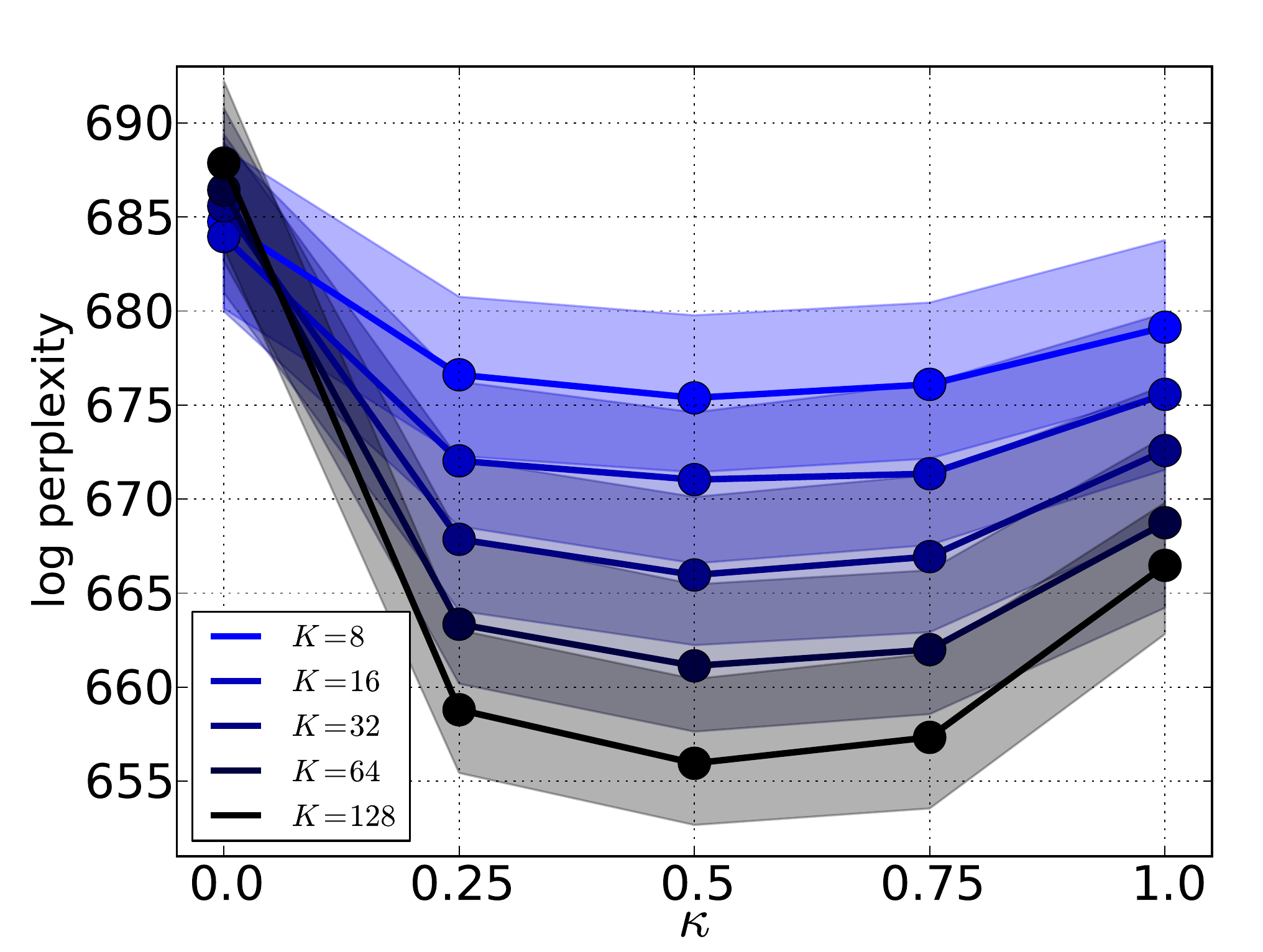}}
\subfigure[\emph{Without} averaging]{\includegraphics[width=0.49\columnwidth]{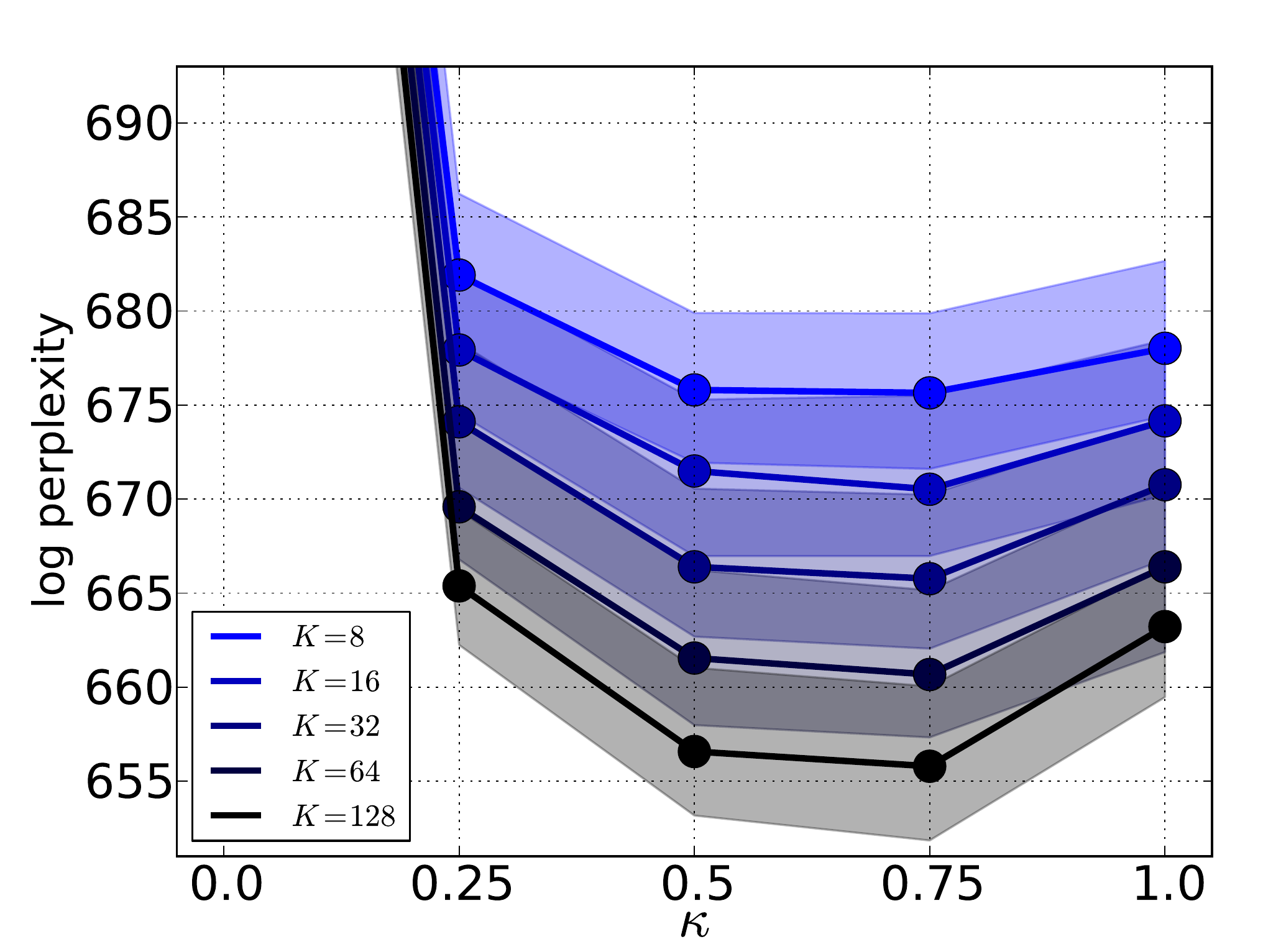}}
\caption{Dataset: IMDB. Perplexity on different test sets as a function of the exponent $\kappa$---the corresponding stepsize is $\rho_i=\frac{1}{i^\kappa}$---for \texttt{G-OEM} with averaging (left) and without averaging (right). The number of topics inferred $K$ goes from 8 (the lightest) to 128 (the darkest). Best seen in color.} \label{fig:IMDB-Acomp}
\end{center}
\vskip -0.2in
\end{figure}

Results are presented in Figure~\ref{fig:Synth-Acomp} on the synthetic data and in Figure~\ref{fig:IMDB-Acomp} on the IMDB dataset.
For ${\kappa\in\left[0,\frac{1}{2}\right[}$, averaging improves the performance while for ${\kappa\in\left]\frac{1}{2},1\right]}$, averaging deteriorates the performance. For ${\kappa=\frac{1}{2}}$, averaging is only slightly beneficial on IMDB dataset.
For constant stepsizes $\kappa=0$ the averaging improves significantly the performance, as the iterates do not converge and tend to oscillate around a local optimum \citep{NewtonSteps2013}. We can expect the same effect for $\kappa\in\left[0,\frac{1}{2}\right[$ as the function $n\mapsto \frac{1}{n^\kappa} $ deacreases slowly for such values of $\kappa$.
For ${\kappa\in\left]\frac{1}{2},1\right]}$, the stochastic gradient ascent scheme is guaranteed to converge to a local optimum \citep{OnlineLearning1998}. The averaging then deteriorates the performance as it incorporates the first iterates, which gets the last iterate away from local optimum. However, the stepsize $1/i$ (${\kappa=1}$) is not competitive. {The performance with ${\kappa=0.75}$ is only slightly better on IMDB dataset.} The setting ${\kappa=\frac{1}{2}}$ represents a good balance between first and last iterates. For this step-size, performances with or without averaging are similar but results without averaging seem to be more stable, hence our choice for all our other simulations.

\begin{figure}[!h]
\begin{center}
\subfigure[IMDB.]{\includegraphics[width=0.49\columnwidth]{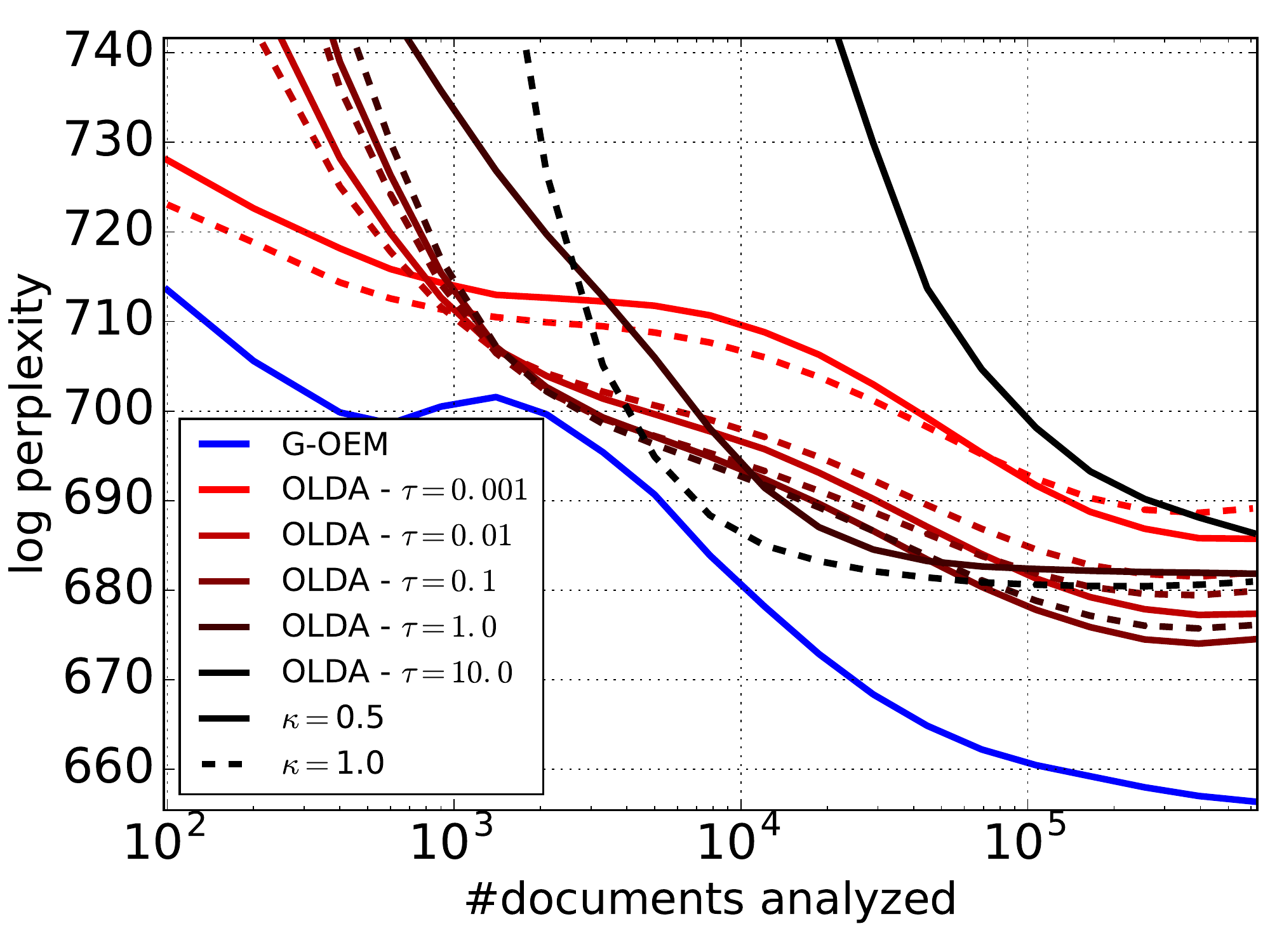}}
\subfigure[Wikipedia.]{\includegraphics[width=0.49\columnwidth]{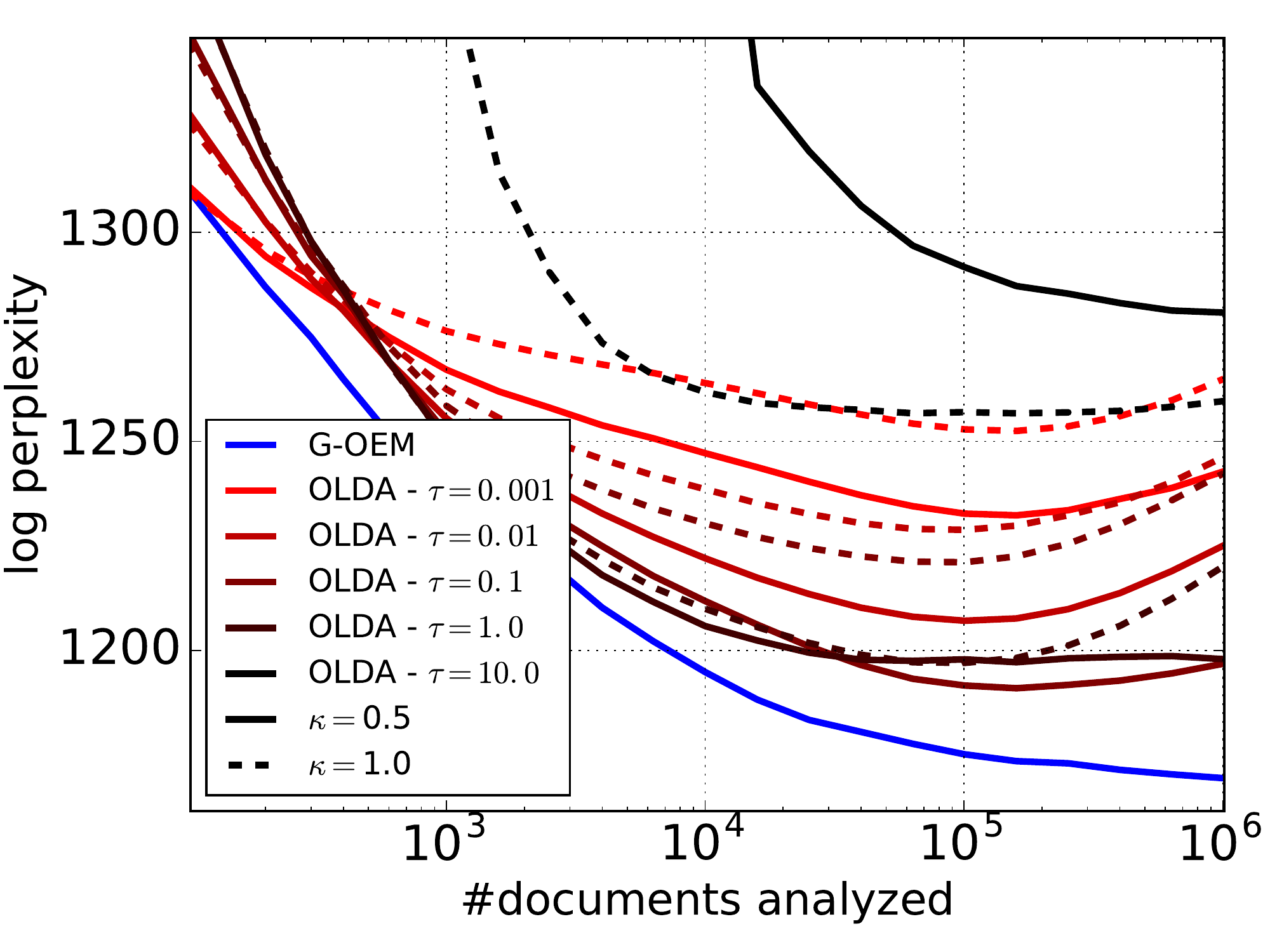}}
\caption{Evolution of perplexity on different test sets as a function of the number of documents analyzed. For \texttt{OLDA}, we compare the performance with different step-sizes $\rho_t=\tau/t^\kappa$ for different values of $\tau,\kappa$. Solid line: $\kappa=1/2$; Dashed line: $\kappa=1$.} 
\label{fig:OLDAsteps}
\end{center}
\vskip -0.2in
\end{figure}

\begin{figure}[!h]
\begin{center}
\subfigure[IMDB.]{\includegraphics[width=0.49\columnwidth]{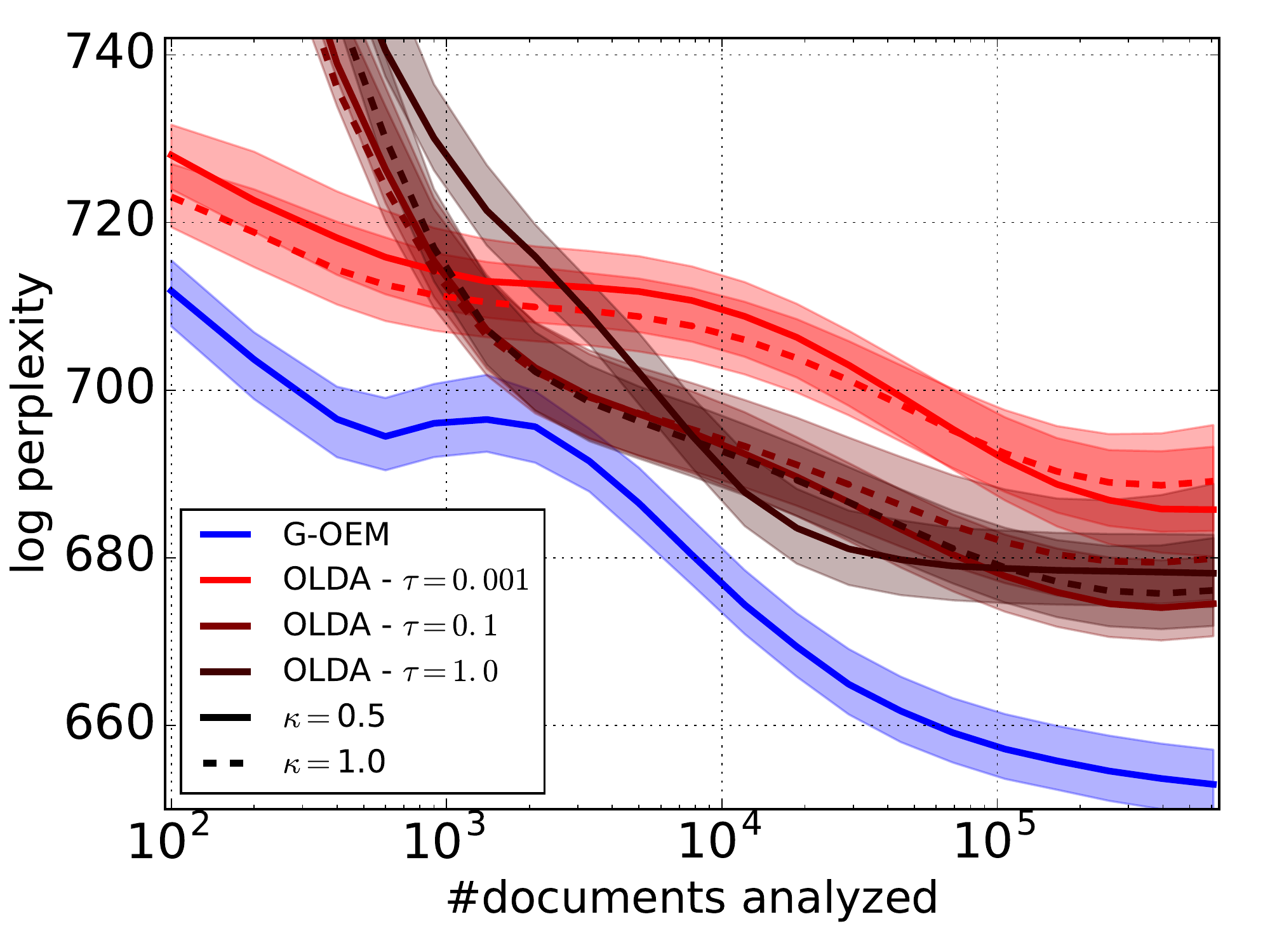}}
\subfigure[Wikipedia.]{\includegraphics[width=0.49\columnwidth]{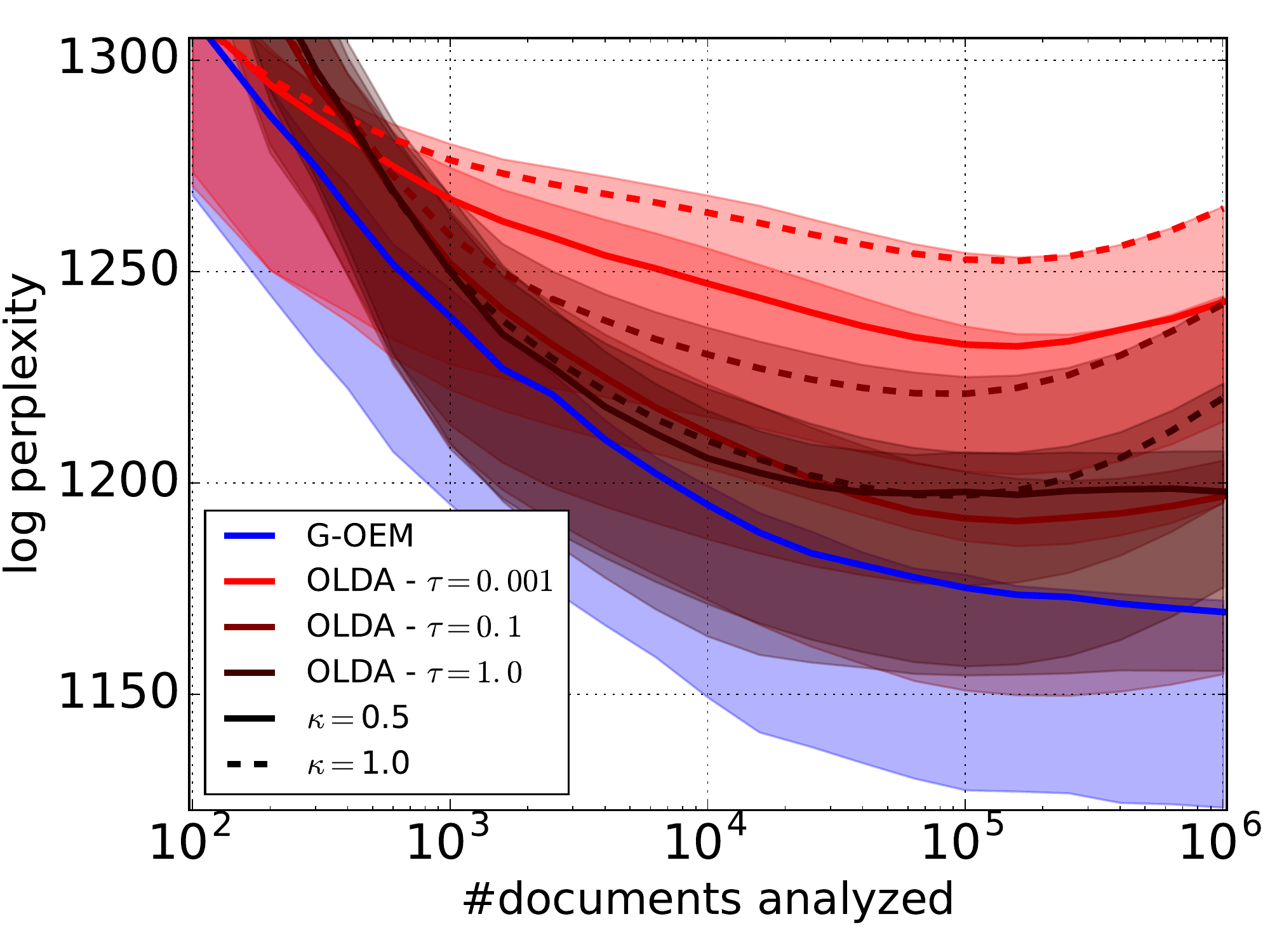}}
\caption{Evolution of perplexity on different test sets as a function of the number of documents analyzed, with error bars. For \texttt{OLDA}, we compare the performance with different step-sizes $\rho_t=\tau/t^\kappa$ for different values of $\tau,\kappa$. Solid line: $\kappa=1/2$; Dashed line: $\kappa=1$.} 
\label{fig:OLDAsteps_var}
\end{center}
\vskip -0.2in
\end{figure}

We also apply \texttt{OLDA} with different step-sizes $\rho_t=\tau/t^\kappa$ for different values of $\tau,\kappa$. Results are presented in Figure~\ref{fig:OLDAsteps} without error bars and in Figure~\ref{fig:OLDAsteps_var} with error bars. For \texttt{OLDA}, results are very similar for any step-size.

\clearpage

\clearpage

\clearpage

\section{Evolution of the ELBO}
\label{app:ELBO}
\begin{figure}[!ht]
\begin{center}
\subfigure[IMDB, $K=128$]{\includegraphics[width=0.49\columnwidth]{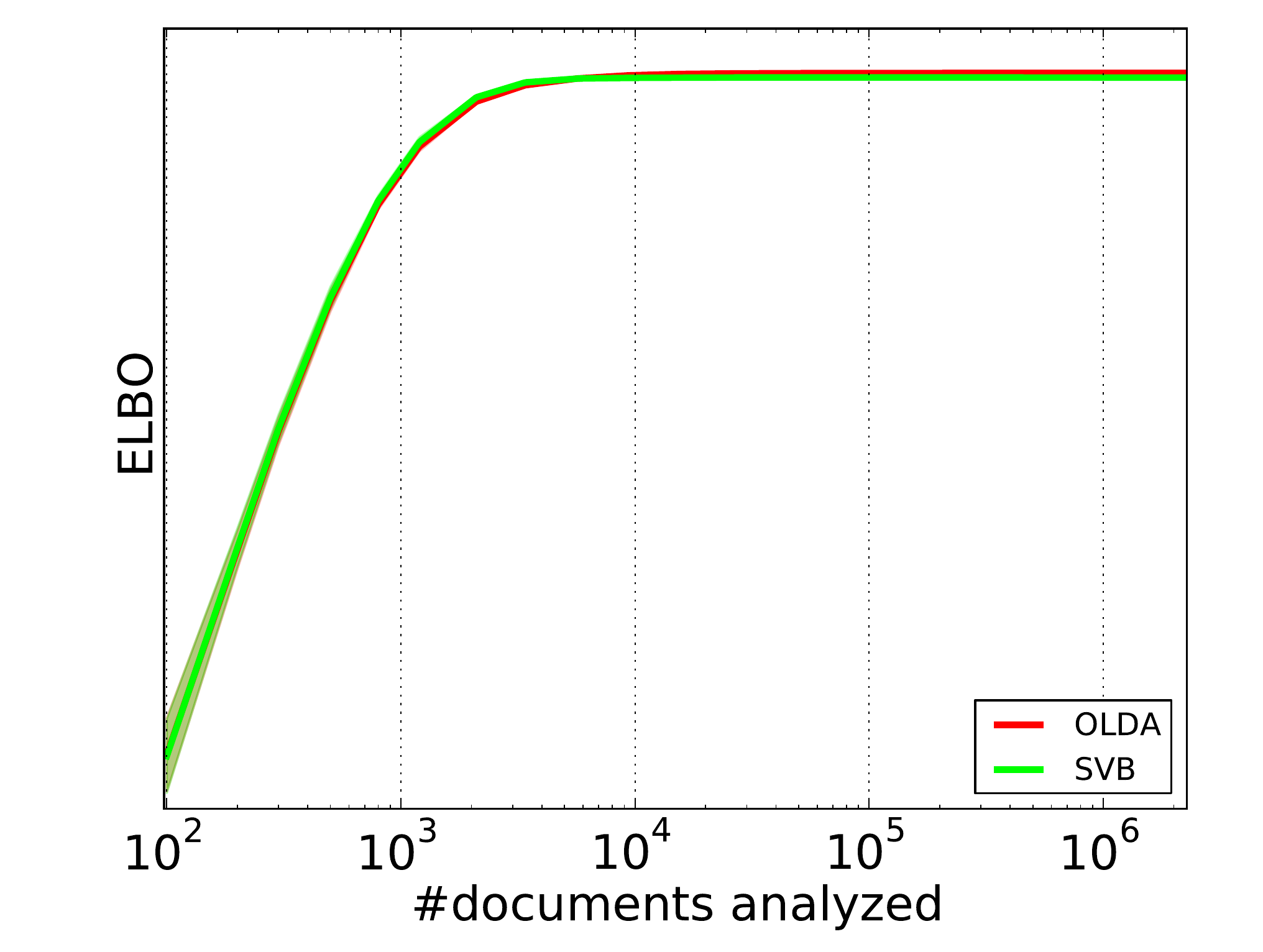}}
\subfigure[Wikipedia, $K=128$]{\includegraphics[width=0.49\columnwidth]{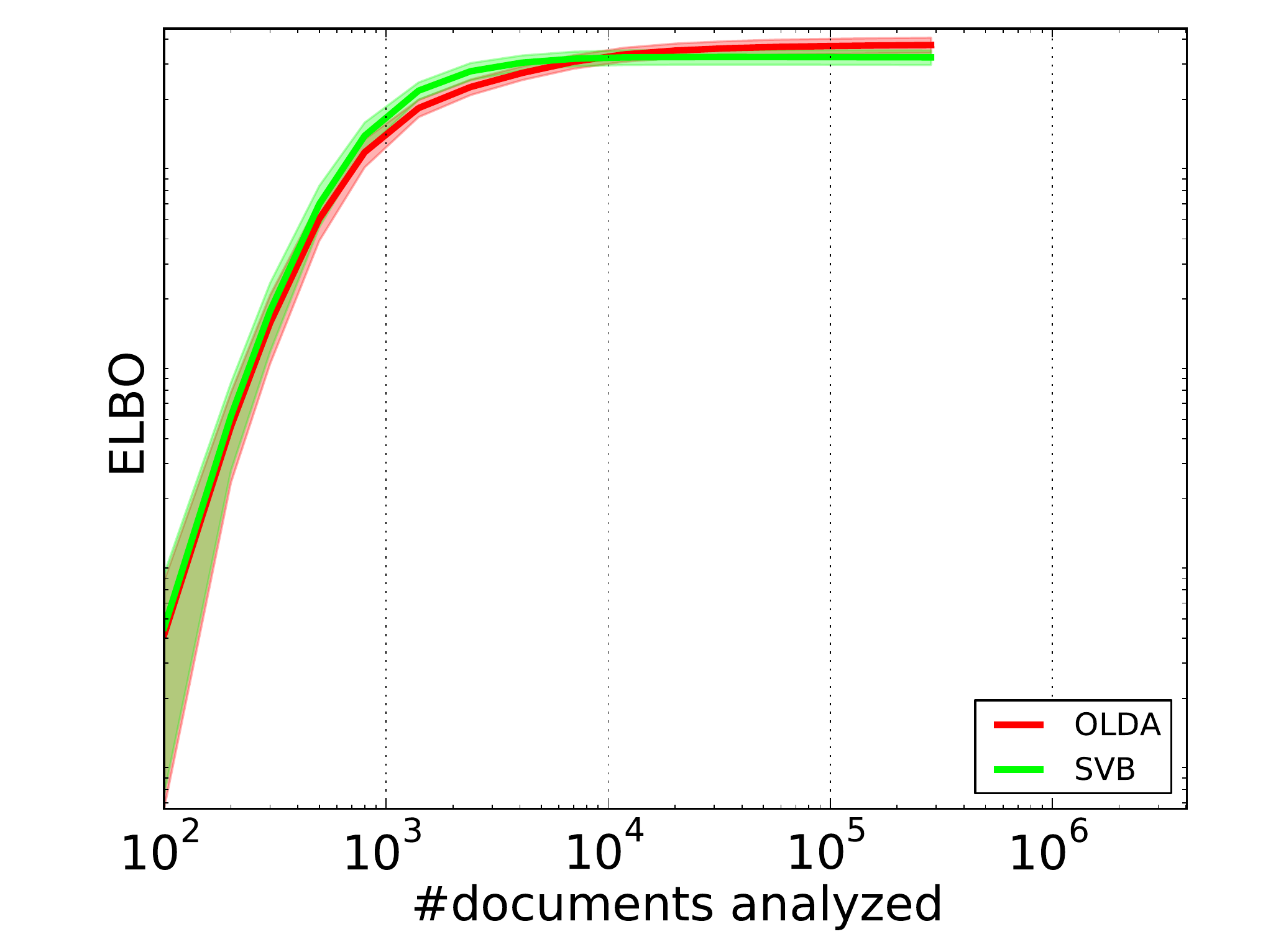}}

\subfigure[IMDB]{\includegraphics[width=0.49\columnwidth]{IMDB_ELBO_Kcomp.pdf}}
\subfigure[Wikipedia]{\includegraphics[width=0.49\columnwidth]{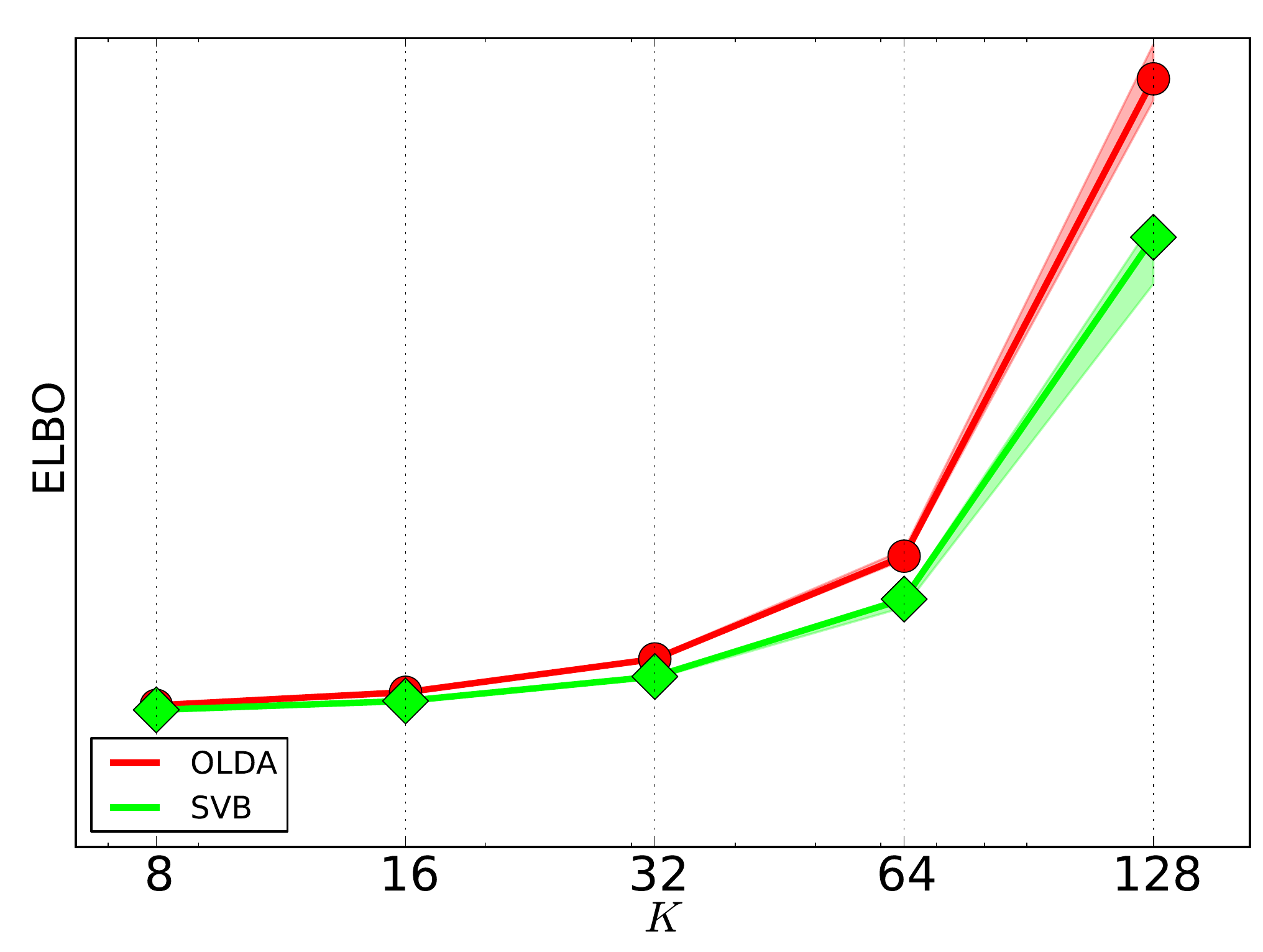}}
\caption{Evidence Lower BOund (ELBO) computed on different test sets. {Top: ELBO through iterations, with 4 passes over each dataset and 200 internal iterations. Bottom: ELBO as a function of the number of topics $K$, with 20 internal iterations and 1 pass over each dataset. Best seen in color.} }
\label{fig:ELBO}
\end{center}
\end{figure}

Figure~\ref{fig:ELBO} presents the evolution of the ELBO for online LDA (\texttt{OLDA}) and \texttt{SVB} on different test sets. We compute the ELBO on test documents as described by \citet{OnlineLDA}. This plot helps us to observe that even if the ELBO reaches a local maximum (i.e., it stabilizes), the quality of the model in terms of perplexity is not controllable. We can also see in Figure~\ref{fig:ELBO} that the ELBO is much better optimized with $K=128$  than with other values of $K$ for both \texttt{SVB} and \texttt{OLDA}, that is, as expected, latent variables of higher dimensionality lead to better fits for the cost function which is optimized.
However, for several datasets the performance in terms of perplexity is better with low values of $K$ ($K=8$ or $K=16$) {than with high dimensional variables ($K=64$ or $K=128$)}.

In order to check if more internal iterations could help variational Bayesian methods, we present in Table~\ref{tab:perpl} the values of perplexity reached by \texttt{OLDA} when running 4 passes over each dataset with $P=200$ internal iterations 1 pass over each dataset with $P=20$ internal iterations. We observe that the ELBO converges quickly to a local optimum and doing ten times more internal iterations does not change significantly the final performance.

\begin{table}[!ht]
\vskip 0.1cm
\begin{center}
\begin{tabular}{|lcc|}
\hline
 & $P=200$, 4 passes & $P=20$, 1 pass \\
\hline
OLDA   & 682.6$\pm$3.7  & 681.9$\pm$3.9 \\
SVB & 683.8$\pm$3.8 & 684.5$\pm$3.8  \\
\hline
\end{tabular}
\end{center}
\caption{Comparison of log-perplexity levels reached with \texttt{OLDA} and \texttt{SVB} on IMDB dataset.}
\label{tab:perpl}
\end{table}

\section{Updates in $\alpha$}
\label{app:alpha}
\begin{figure}[!t]
\begin{center}
\subfigure[\texttt{G-OEM}]{\includegraphics[width=0.49\columnwidth]{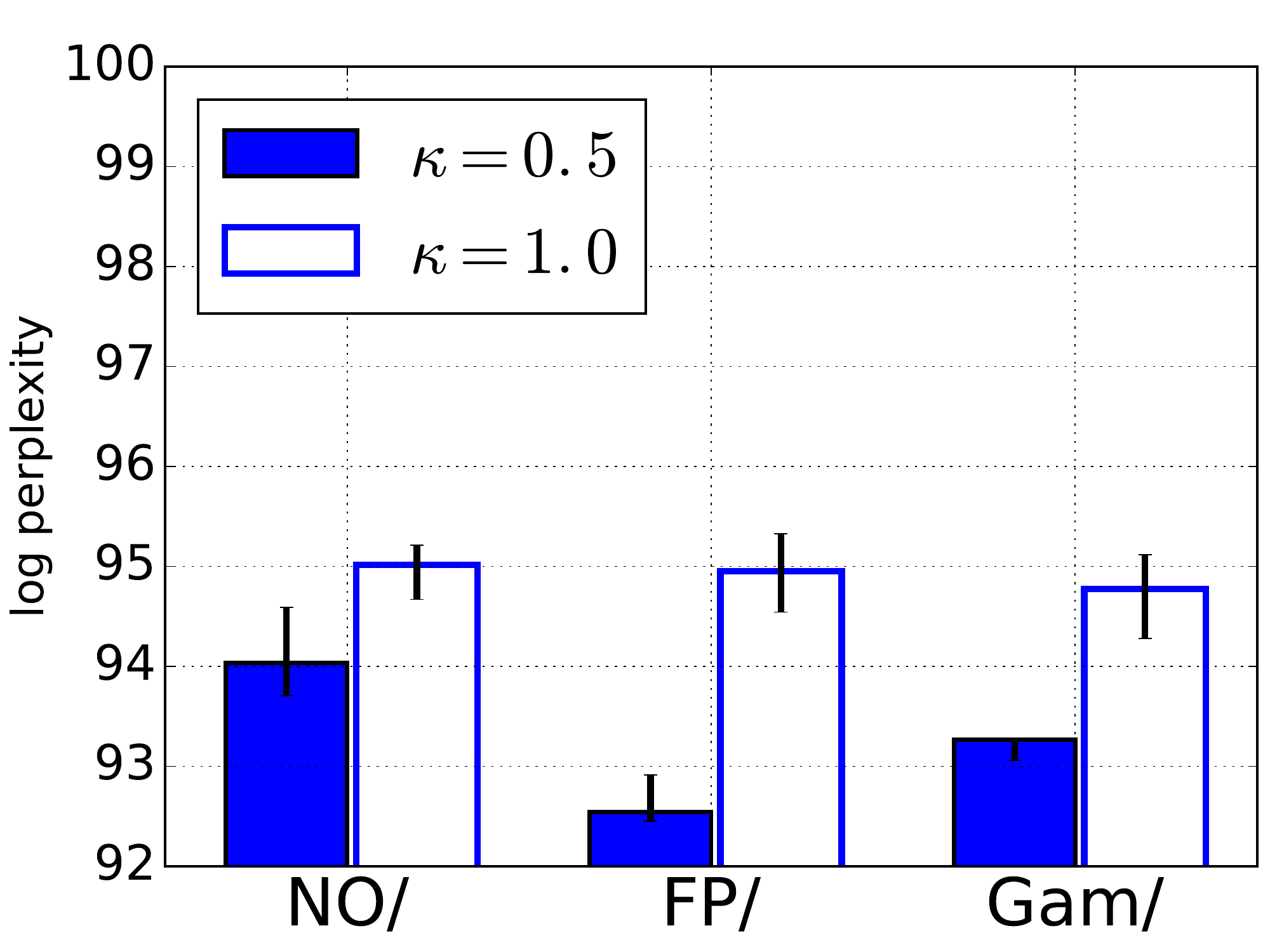}}
\subfigure[\texttt{G-OEM++}]{\includegraphics[width=0.49\columnwidth]{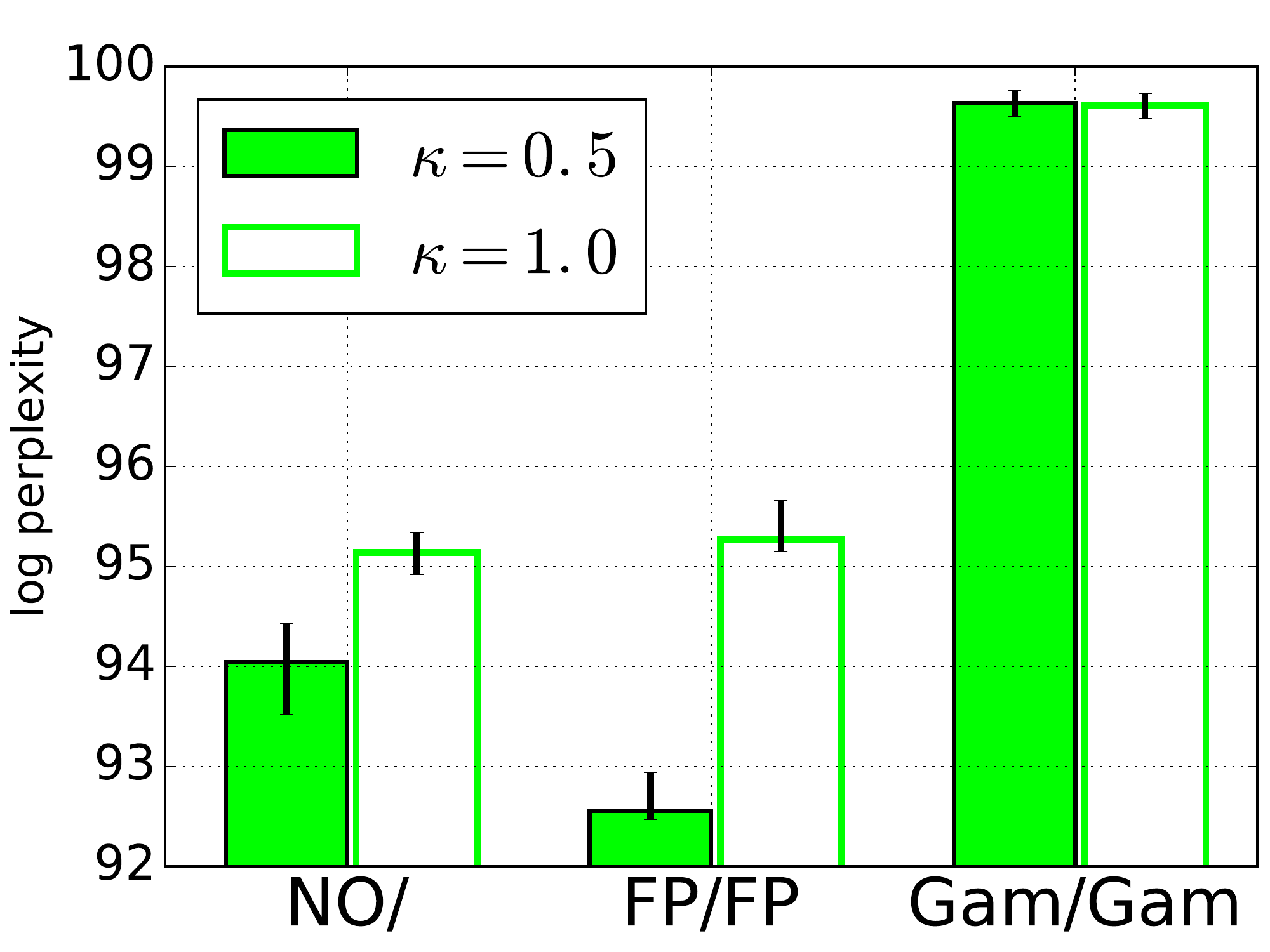}}\\
\subfigure[\texttt{V-OEM}]{\includegraphics[width=0.49\columnwidth]{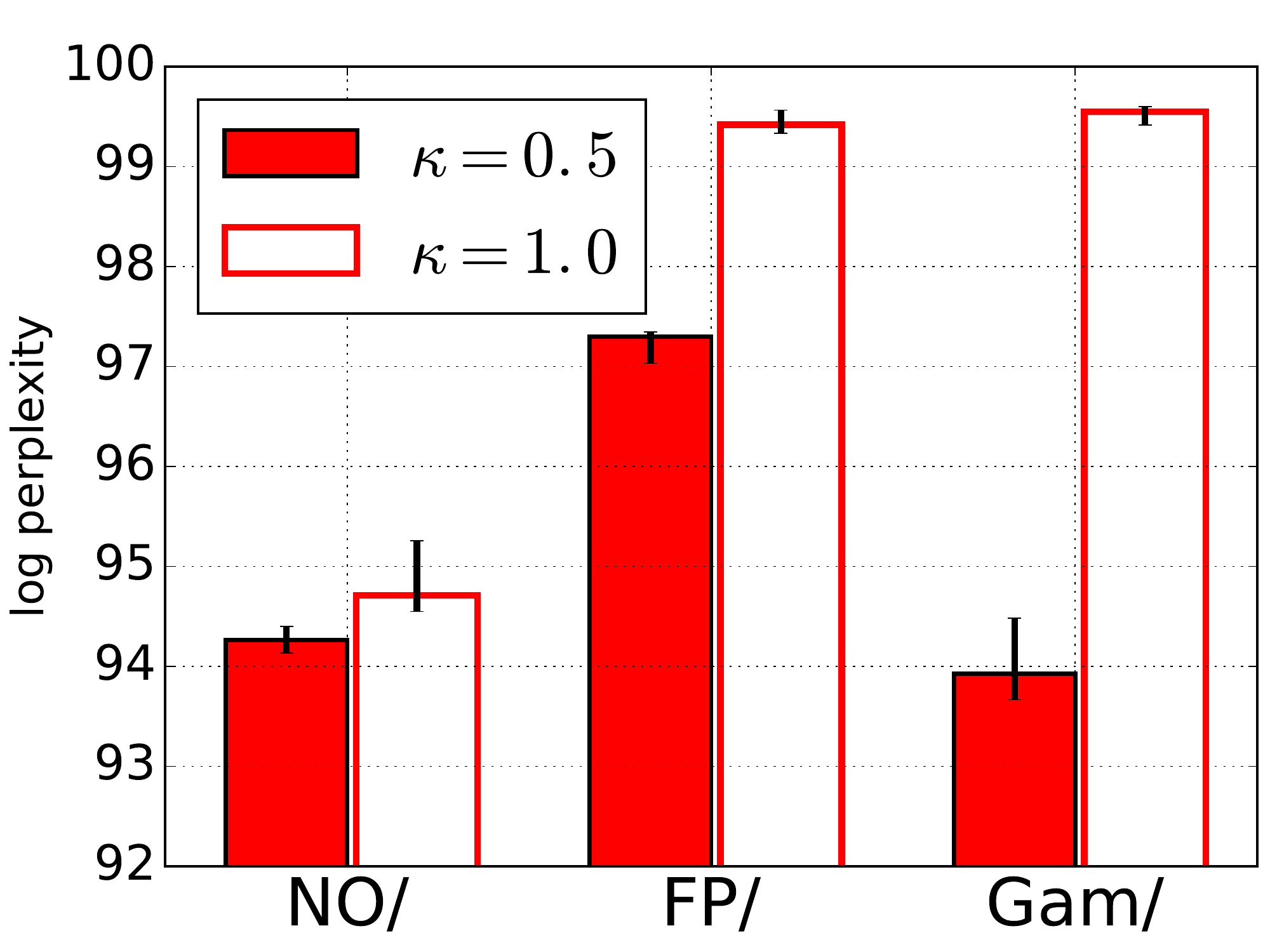}}
\subfigure[\texttt{V-OEM++}]{\includegraphics[width=0.49\columnwidth]{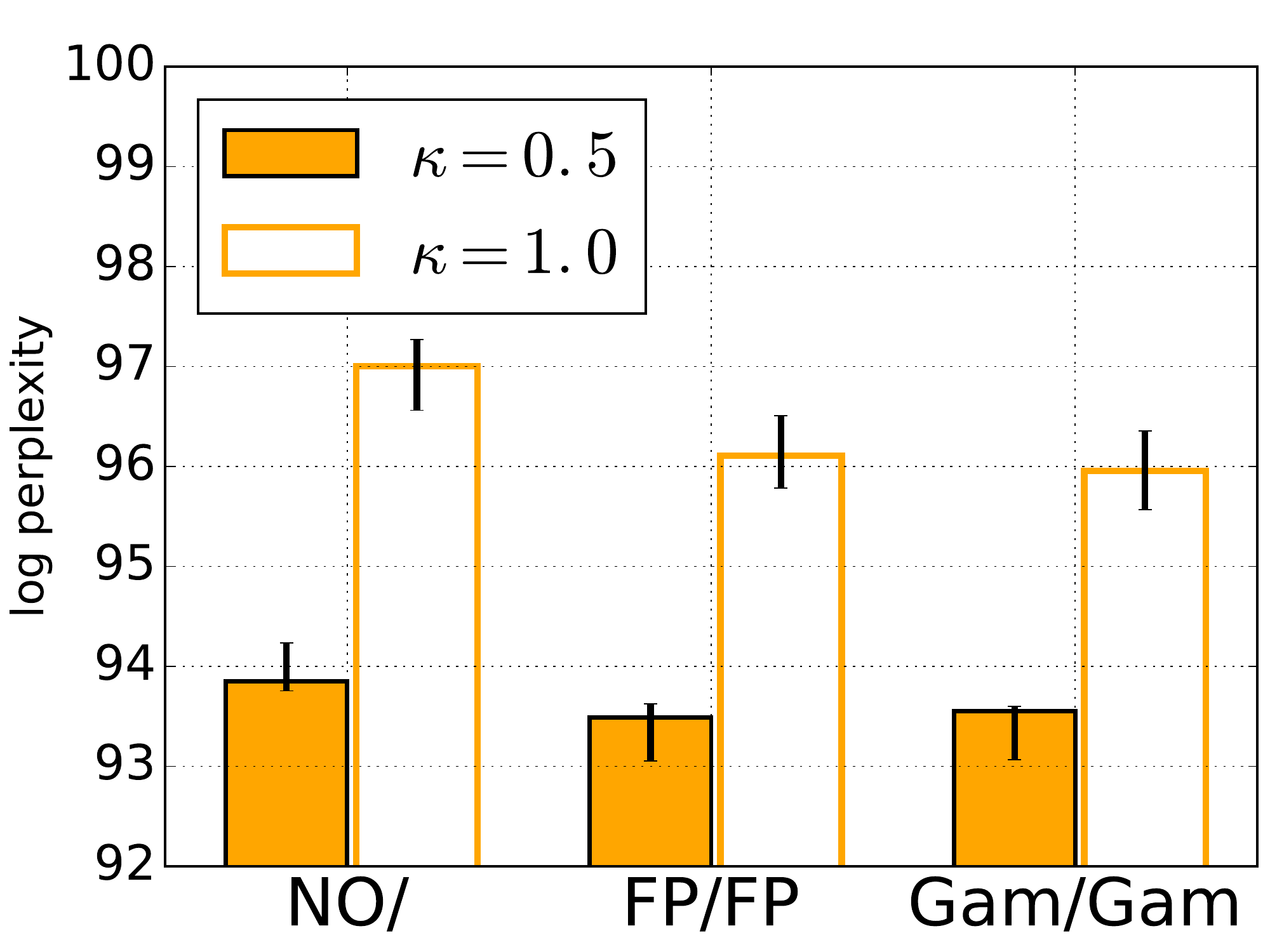}}

\caption{Dataset: Synthetic, $K=10$. Perplexity on different test sets for different types of  updates for $\alpha$; for boosted methods, we use the same inference for $\alpha$ for local and global updates. \texttt{NO}: $\alpha$ is fixed and set to $\alpha_{true}$ that generated the data; \texttt{FP}: fixed point iteration; \texttt{Gam}: gamma prior on $\alpha$ \citep{SinglePassLDA}. Best seen in color.}
\label{fig:Synth-Alph}
\end{center}
\vskip -0.2in
\end{figure}

In this section we compare the different types of updates for~$\alpha$. Figure~\ref{fig:Synth-Alph} presents results obtained on synthetic dataset for fixed point iteration algorithm \citep{EstimateDirichlet} and by putting a gamma prior on $\alpha$ \citep{SinglePassLDA}. We observe that the fixed point method leads to better performance for \texttt{G-OEM}  and \texttt{G-OEM++}. For \texttt{V-OEM}, the gamma updates better perform for $\kappa=\frac{1}{2}$. The performances of the gamma updates and the fixed point method are very similar for \texttt{V-OEM++}. Note that the algorithm \texttt{V-OEM++} with $\kappa=1$ and gamma updates on $\alpha$ is exactly equivalent to \texttt{SPLDA} \citep{SinglePassLDA}. The performance of this method can be improved by setting $\kappa=\frac{1}{2}$ with any update on $\alpha$.

We also observe that fixing $\alpha$ to $\alpha_{true}$ that generated the data does not necessarily lead to better performance. 

\clearpage

\section{Performance with Different $K$, with Error Bars}
\label{app:Kcomp}
\begin{figure}[!t]
\begin{center}
\subfigure[Dataset: Synthetic]{\includegraphics[width=0.49\textwidth]{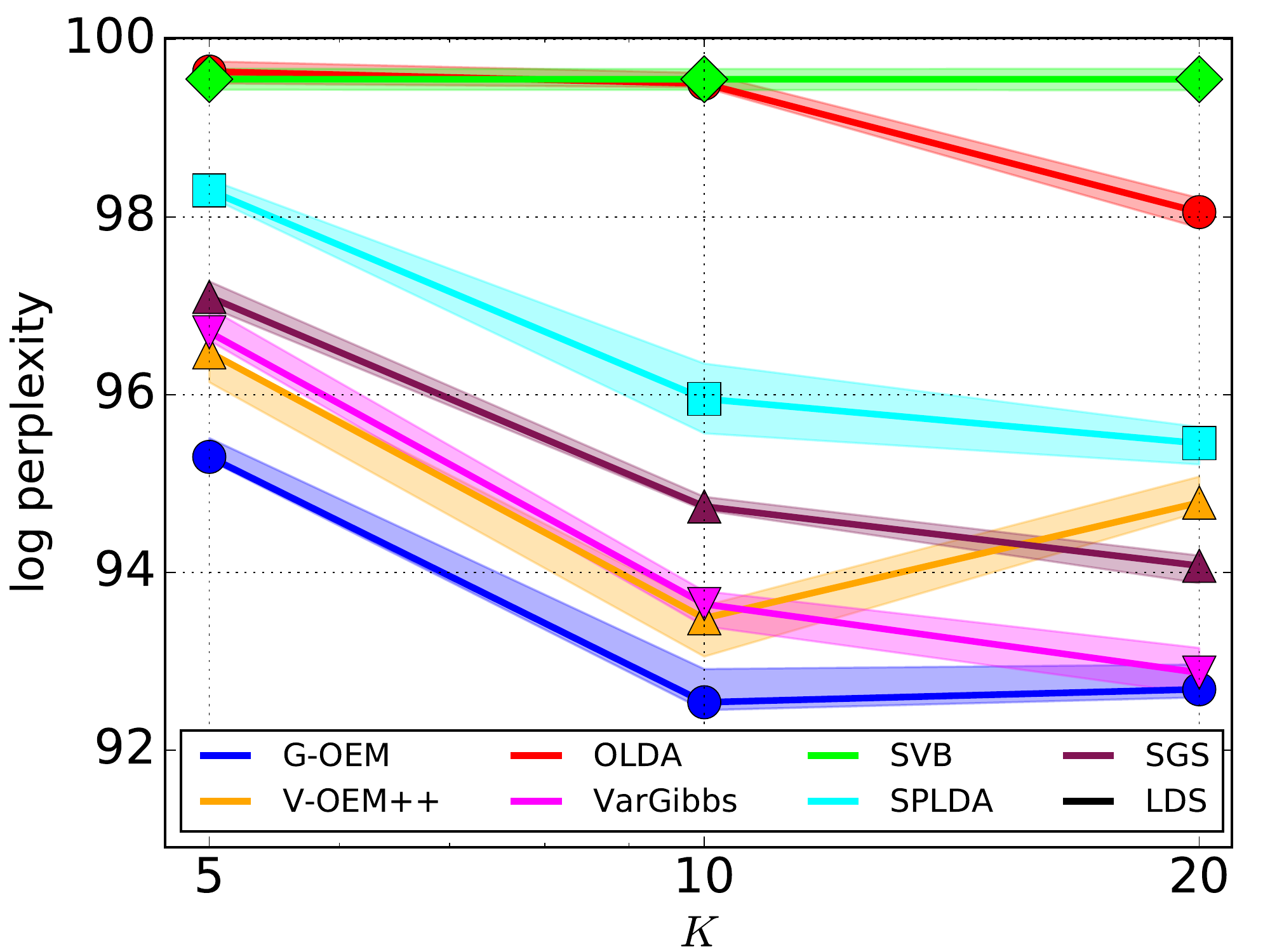}\label{fig:Synth-Kcomp_var}}
\subfigure[Dataset: IMDB]{\includegraphics[width=0.49\textwidth]{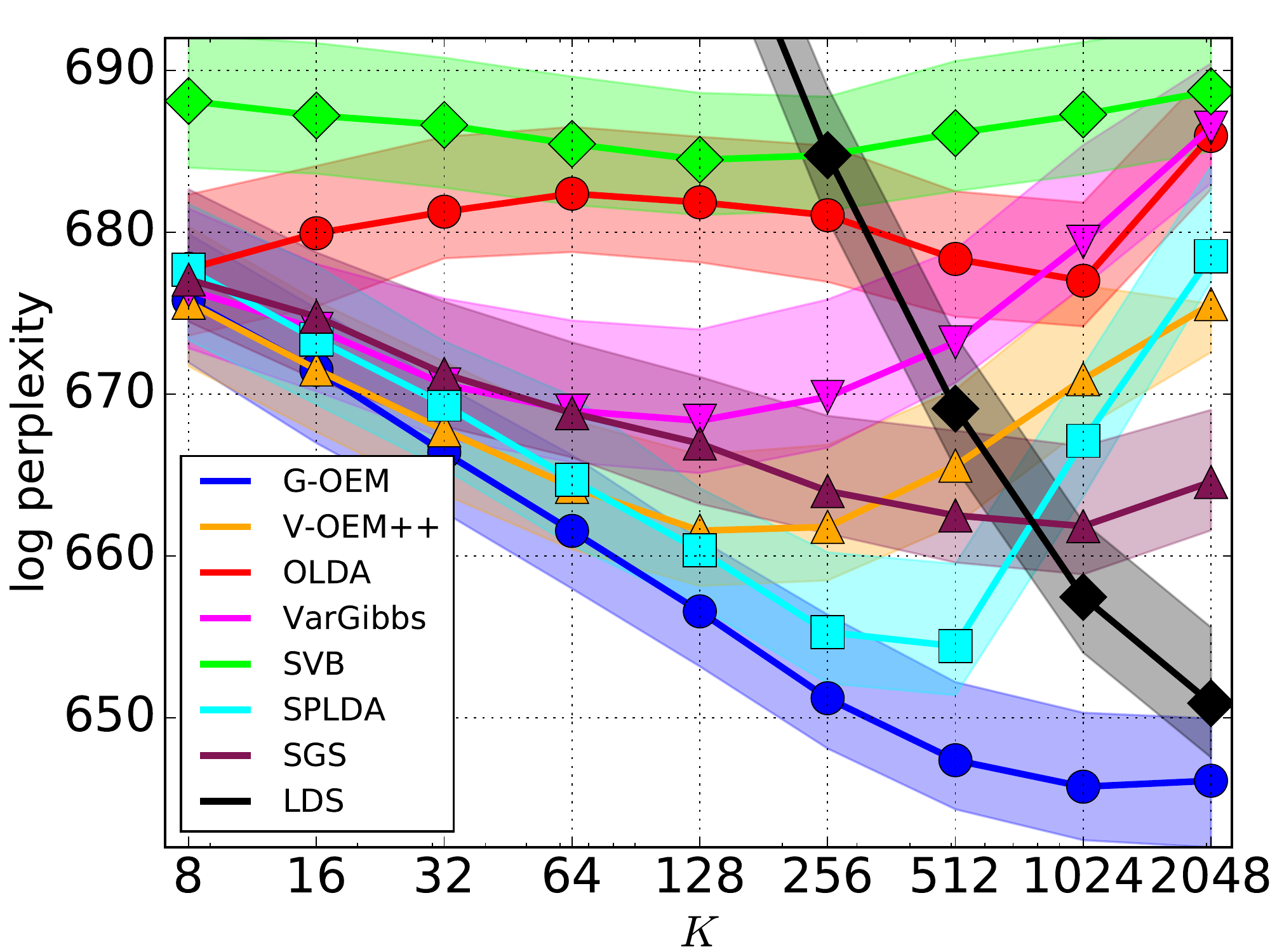}\label{fig:IMDB-Kcomp_var}}

\subfigure[Dataset: Wikipedia]{\includegraphics[width=0.49\textwidth]{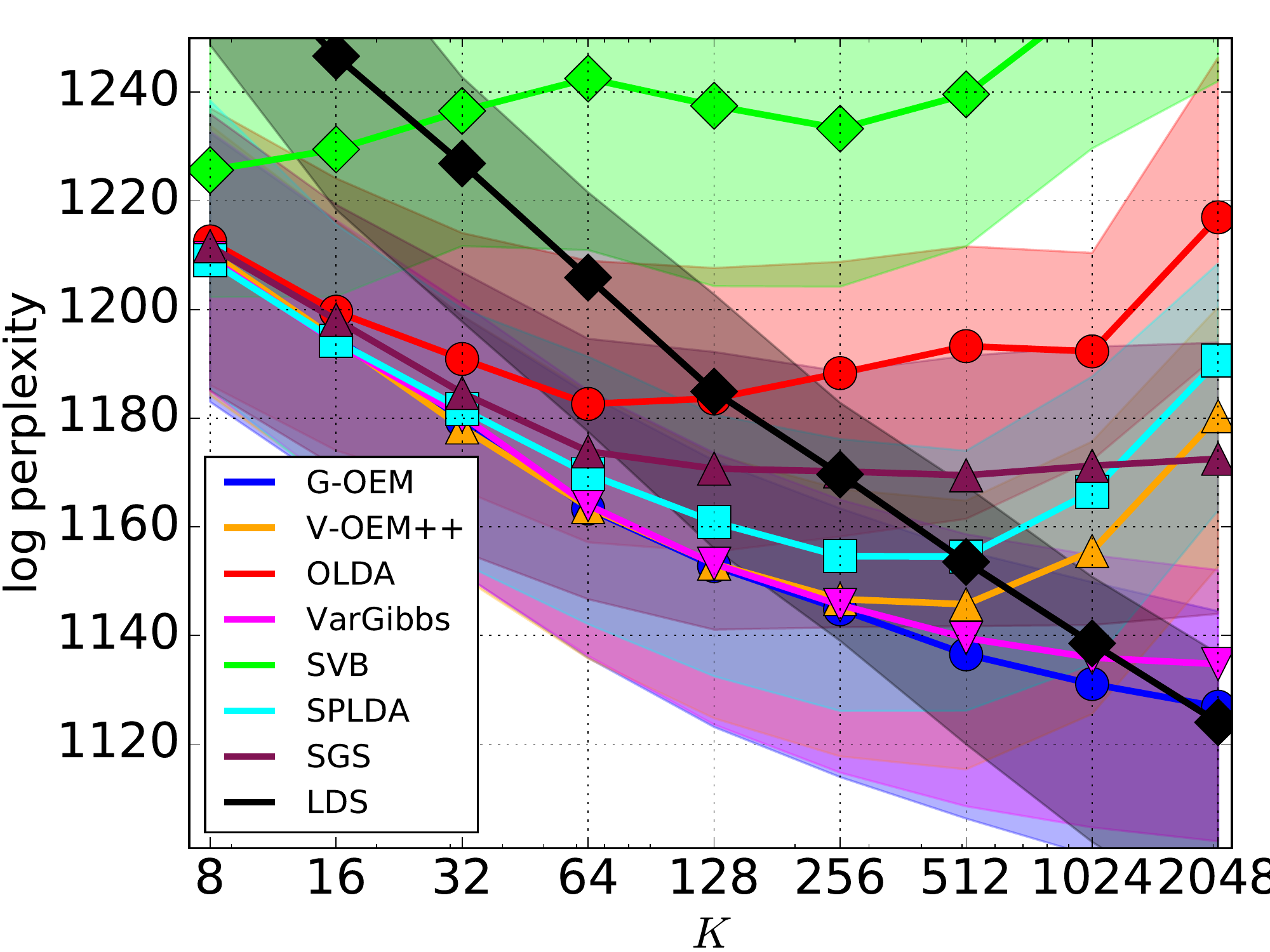}\label{fig:wiki-Kcomp_var}}
\subfigure[Dataset: New York Times]{\includegraphics[width=0.49\textwidth]{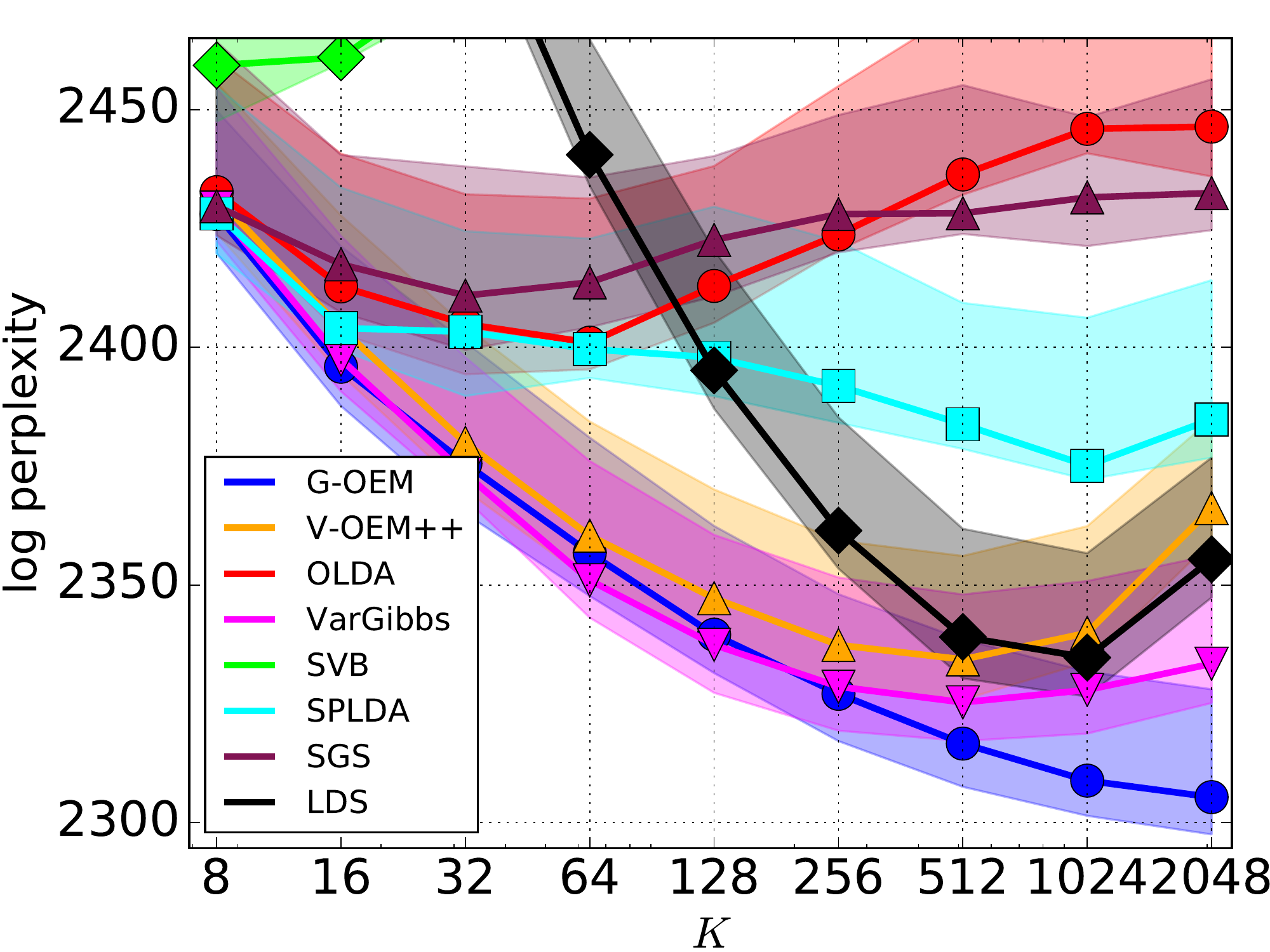}\label{fig:NYT-Kcomp_var}}

\subfigure[Dataset: Pubmed]{\includegraphics[width=0.49\textwidth]{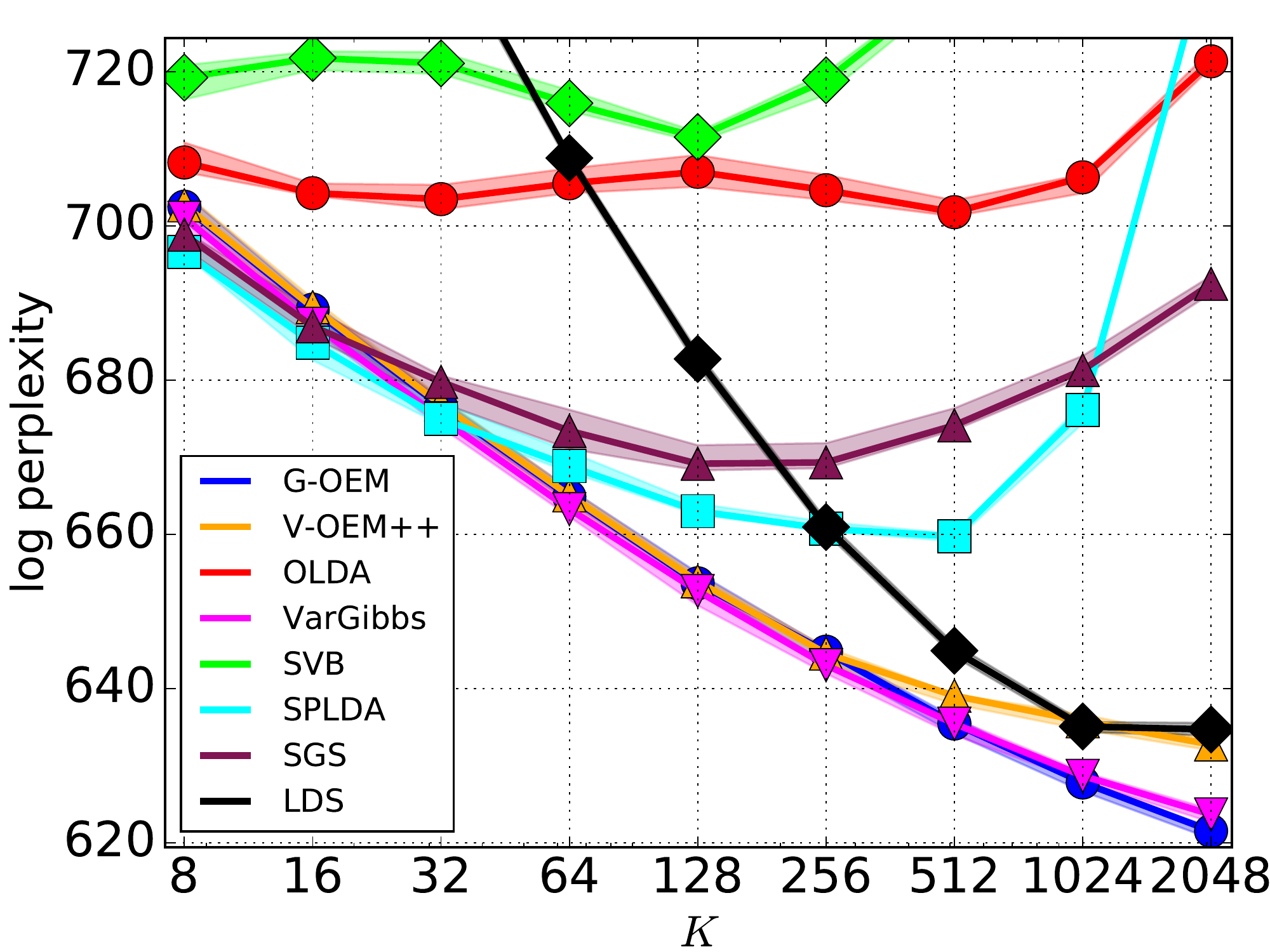}\label{fig:Pubmed-Kcomp_var}}
\subfigure[Dataset: Amazon movies]{\includegraphics[width=0.49\textwidth]{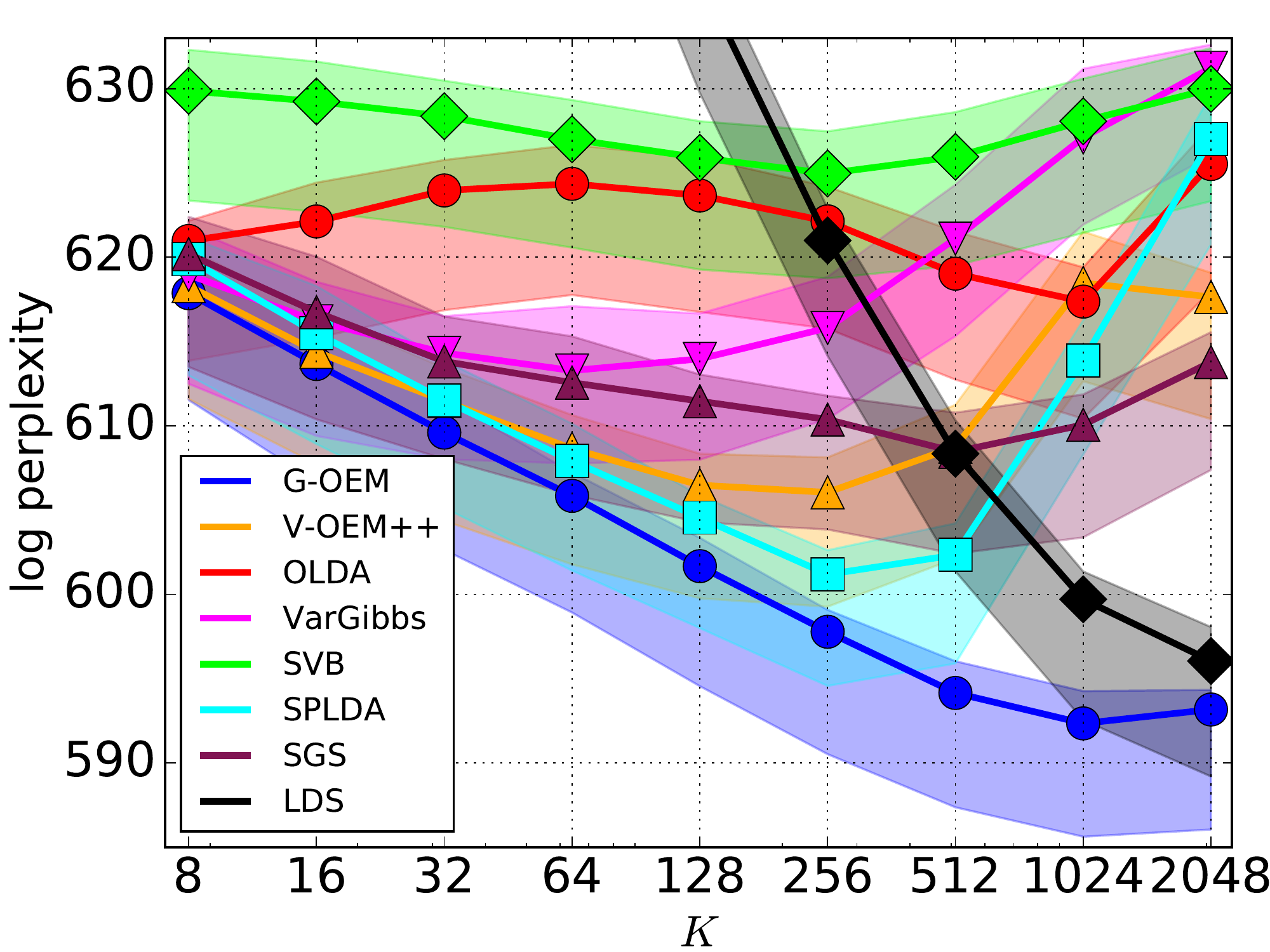}\label{fig:Amazon-Kcomp_var}}

\caption{Perplexity on different test sets as a function of $K$, the number of topics inferred. Same as Figure~\ref{fig:Kcomp}, but with error bars. Best seen in colors.}
\label{fig:Kcomp_var}
\end{center}
\vskip -0.3in
\end{figure}

The performance of the presented methods for different values of $K$ on the different datasets is presented in Figure~\ref{fig:Kcomp_var}. We plot the median from the 11 experiments as a line---solid or dashed---and a shaded region between the third and the seventh decile.

\section{Performance Through Iterations, with Error Bars}
\label{app:ite}
\begin{figure}[!t]
\begin{center}
\subfigure[\label{fig:Synth_ite_var}Synthetic, $K=10$]{\includegraphics[width=0.49\textwidth]{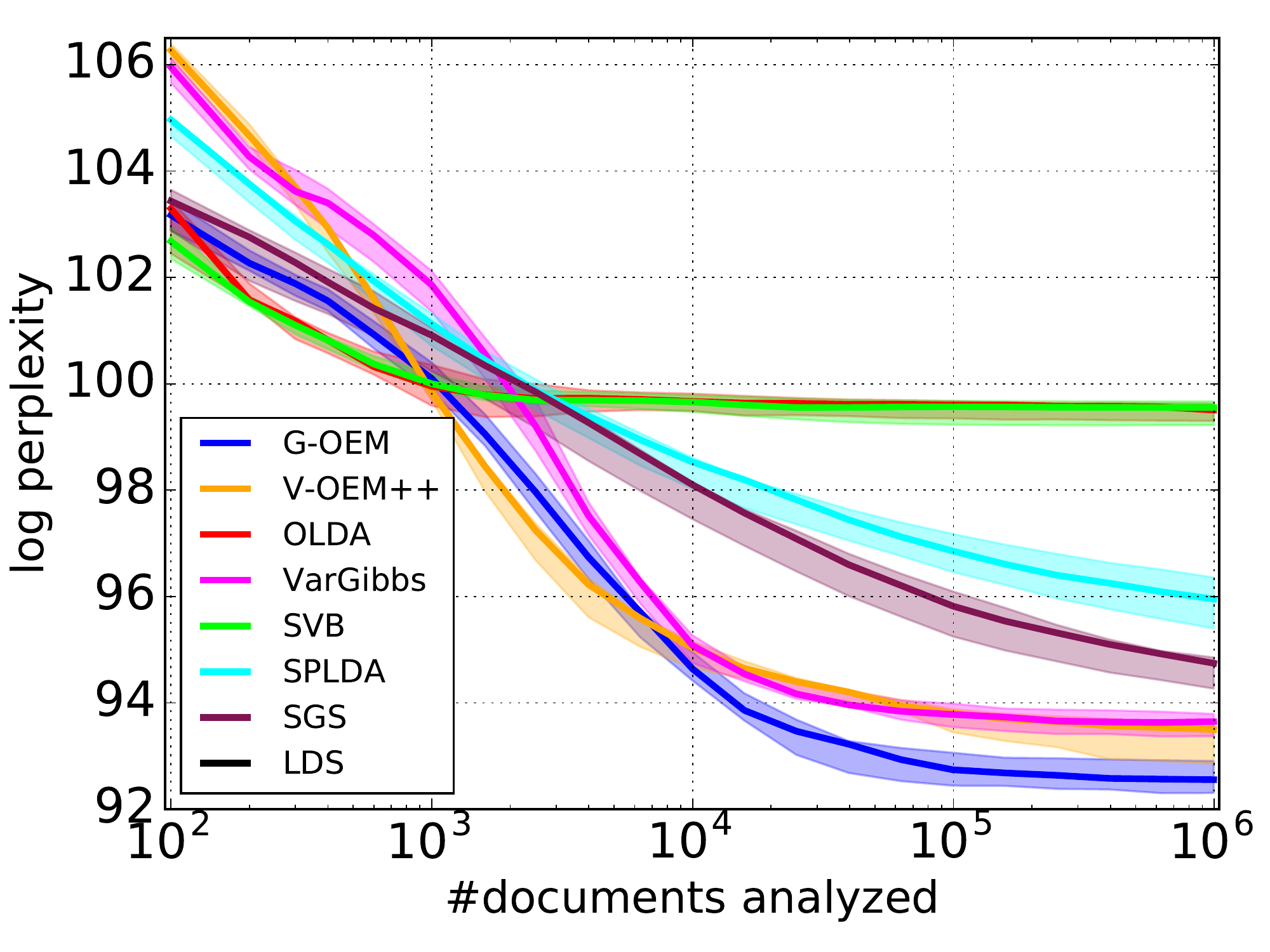}}
\subfigure[\label{fig:IMDB_ite_var}IMDB, $K=128$]{\includegraphics[width=0.49\textwidth]{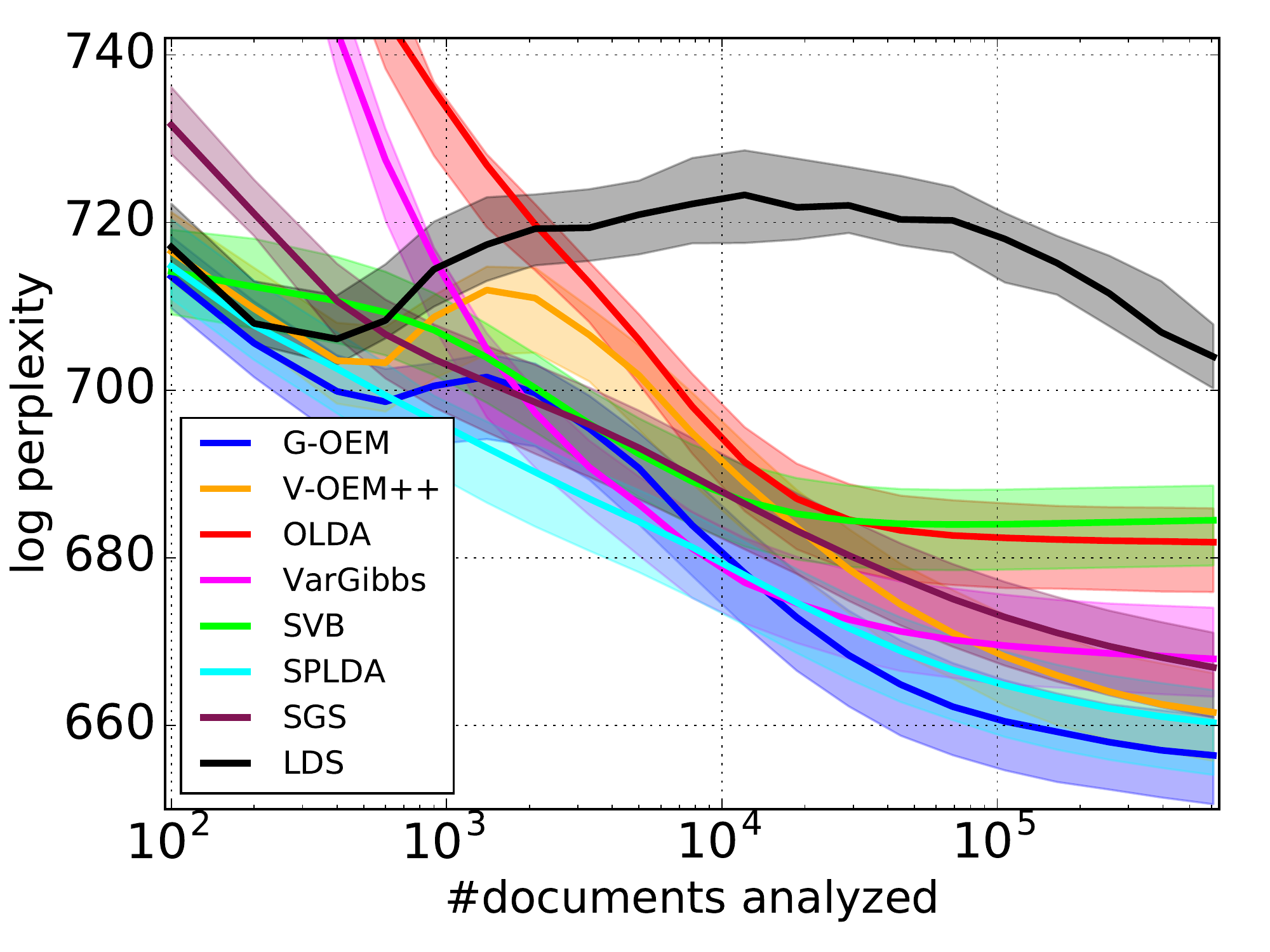}}

\subfigure[\label{fig:wiki_ite_var}Wikipedia, $K=128$]{\includegraphics[width=0.49\textwidth]{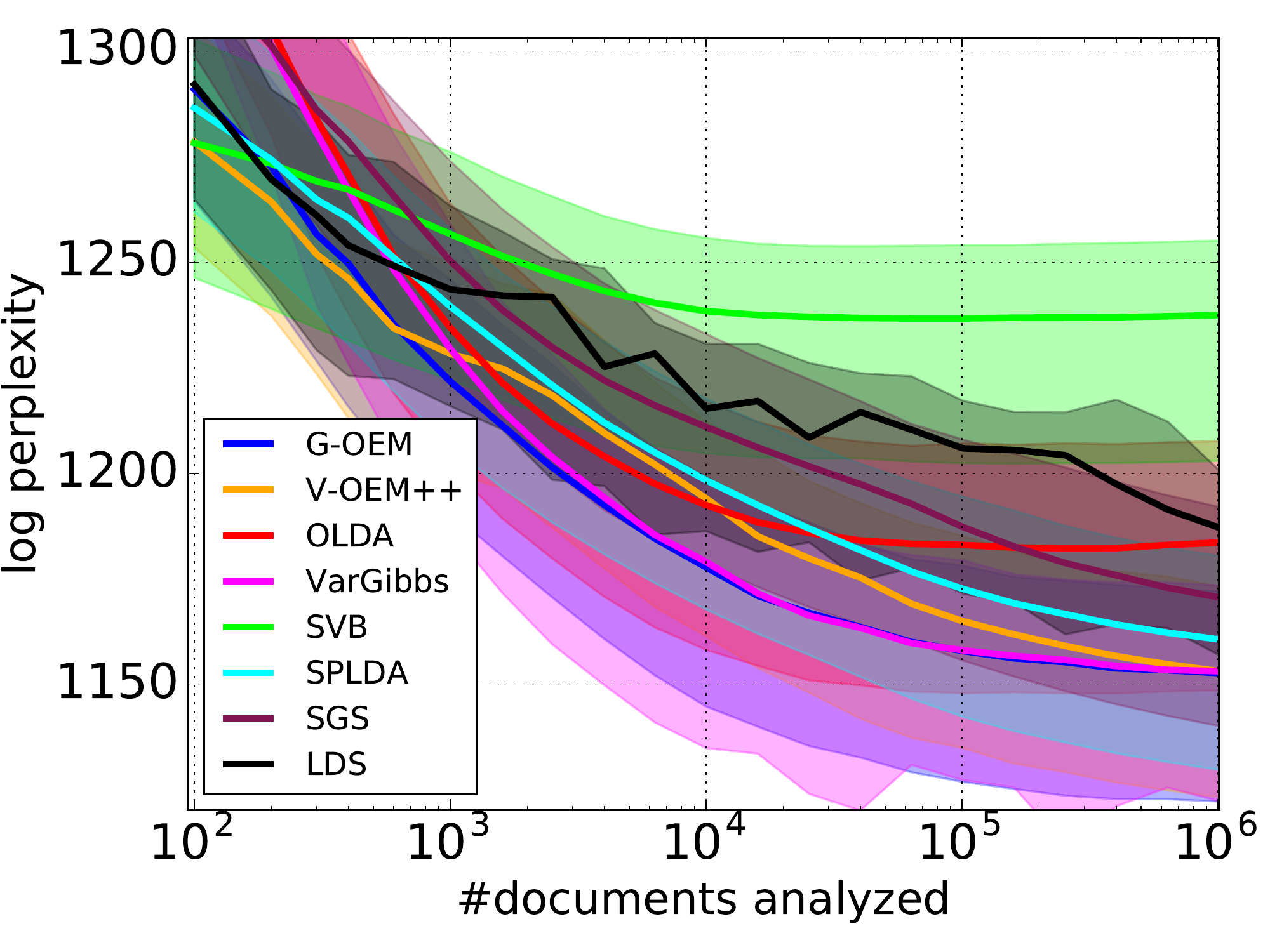}}
\subfigure[\label{fig:NYT_ite_var}New York Times, $K=128$]{\includegraphics[width=0.49\textwidth]{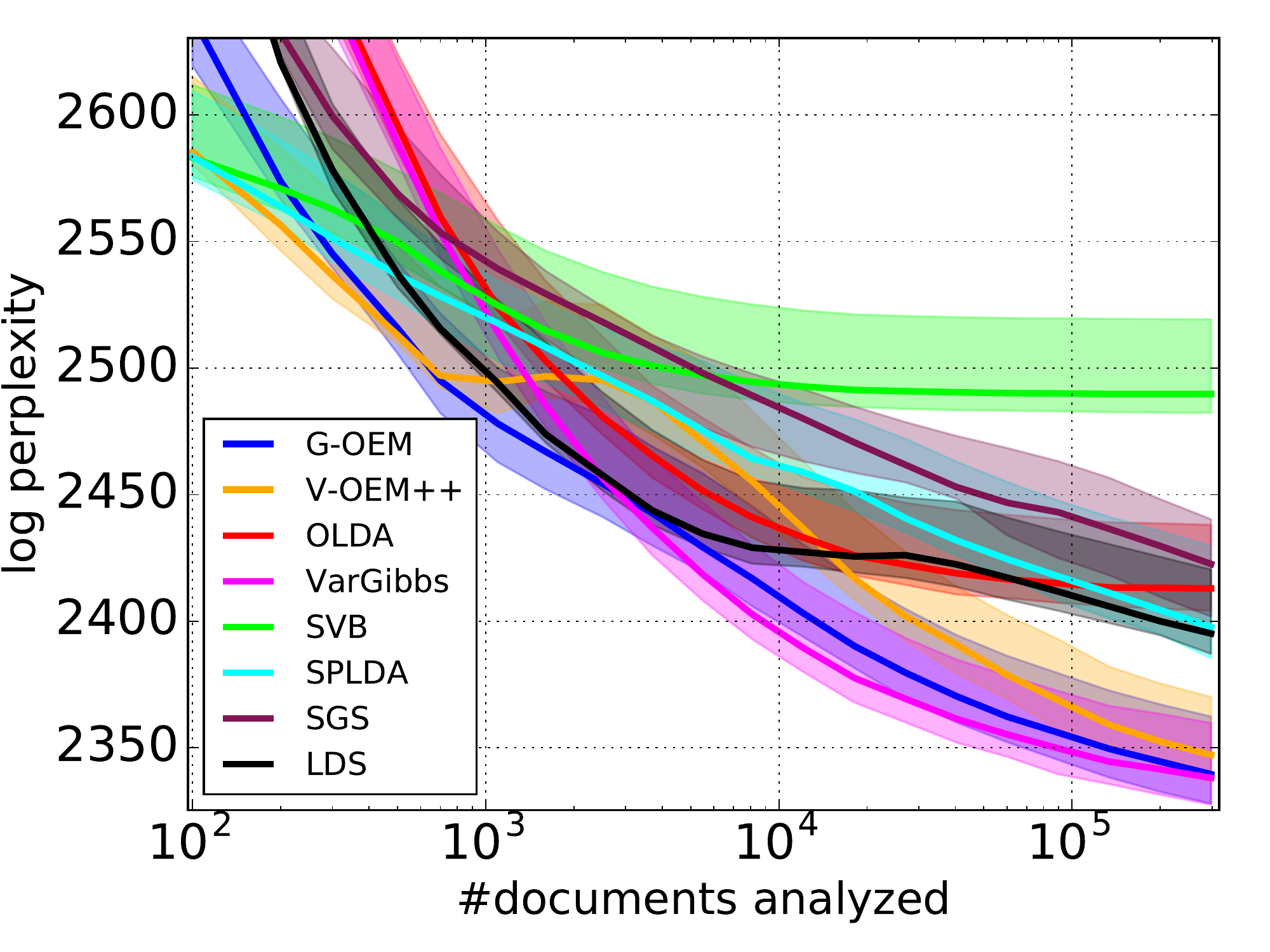}}

\subfigure[\label{fig:Pubmed_ite_var}Pubmed, $K=128$]{\includegraphics[width=0.49\textwidth]{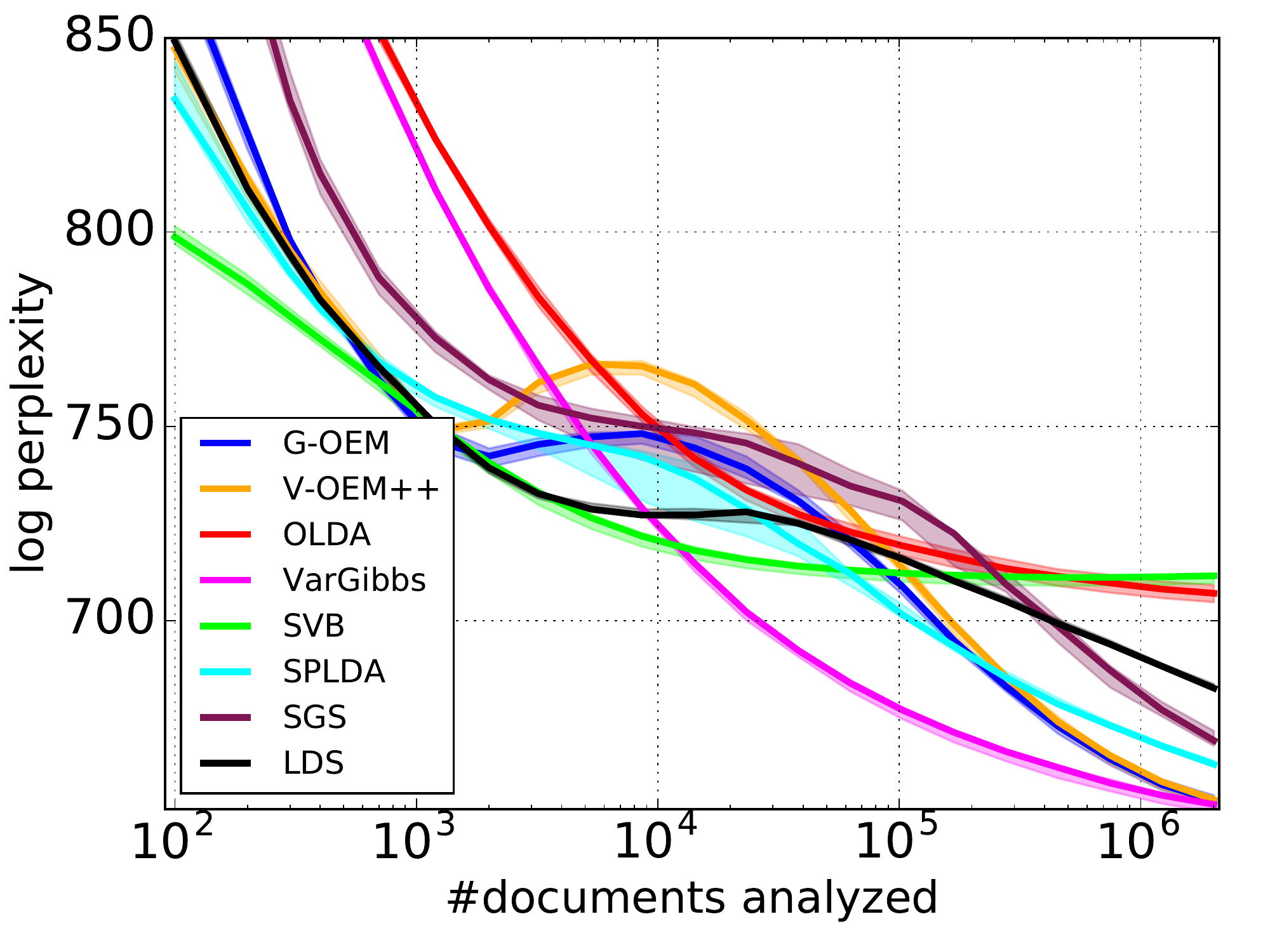}}
\subfigure[\label{fig:Amazon_ite_var}Amazon movies, $K=128$]{\includegraphics[width=0.49\textwidth]{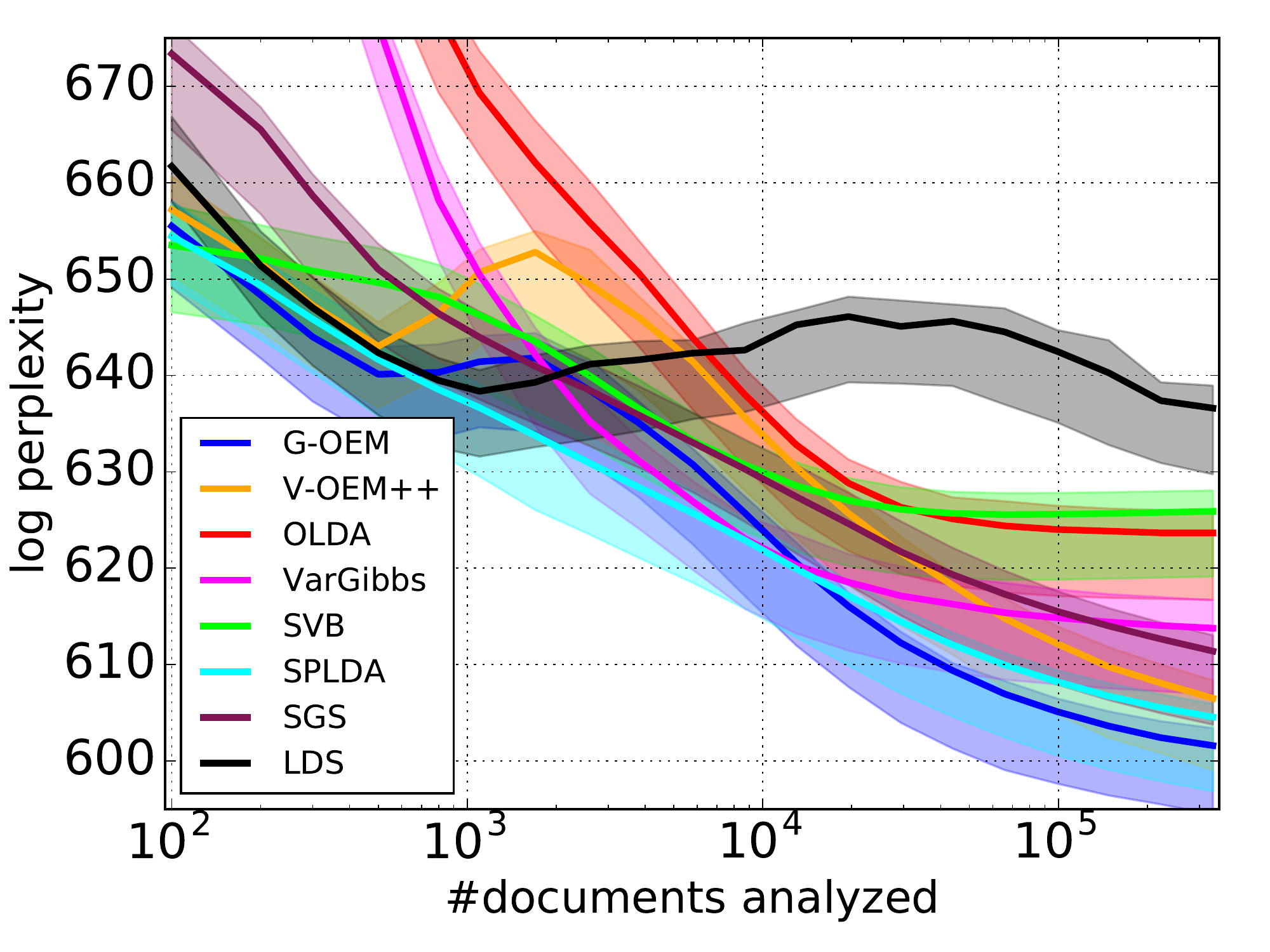}}
\caption{Perplexity through iterations on different test sets with the presented methods. Same as Figure~\ref{fig:ite}, but with error bars. Best seen in colors.}
\label{fig:ite_var}
\end{center}
\end{figure}

The performance through iterations of the presented methods on the different datasets is presented in Figure~\ref{fig:ite_var}. We plot the median from the 11 experiments as a line---solid or dashed---and a shaded region between the third and the seventh decile.

\section{Results on HDP, with Error Bars}
\label{app:hdp}
The performance through iterations of the \texttt{G-OEM} and \texttt{VarGibss} applied to both LDA and HDP is presented in Figure~\ref{fig:ite_var}. We plot the median from the 11 experiments as a line---solid or dashed---and a shaded region between the third and the seventh decile.
\begin{figure}
\begin{center}
\subfigure[IMDB.]{\includegraphics[width=0.48\columnwidth]{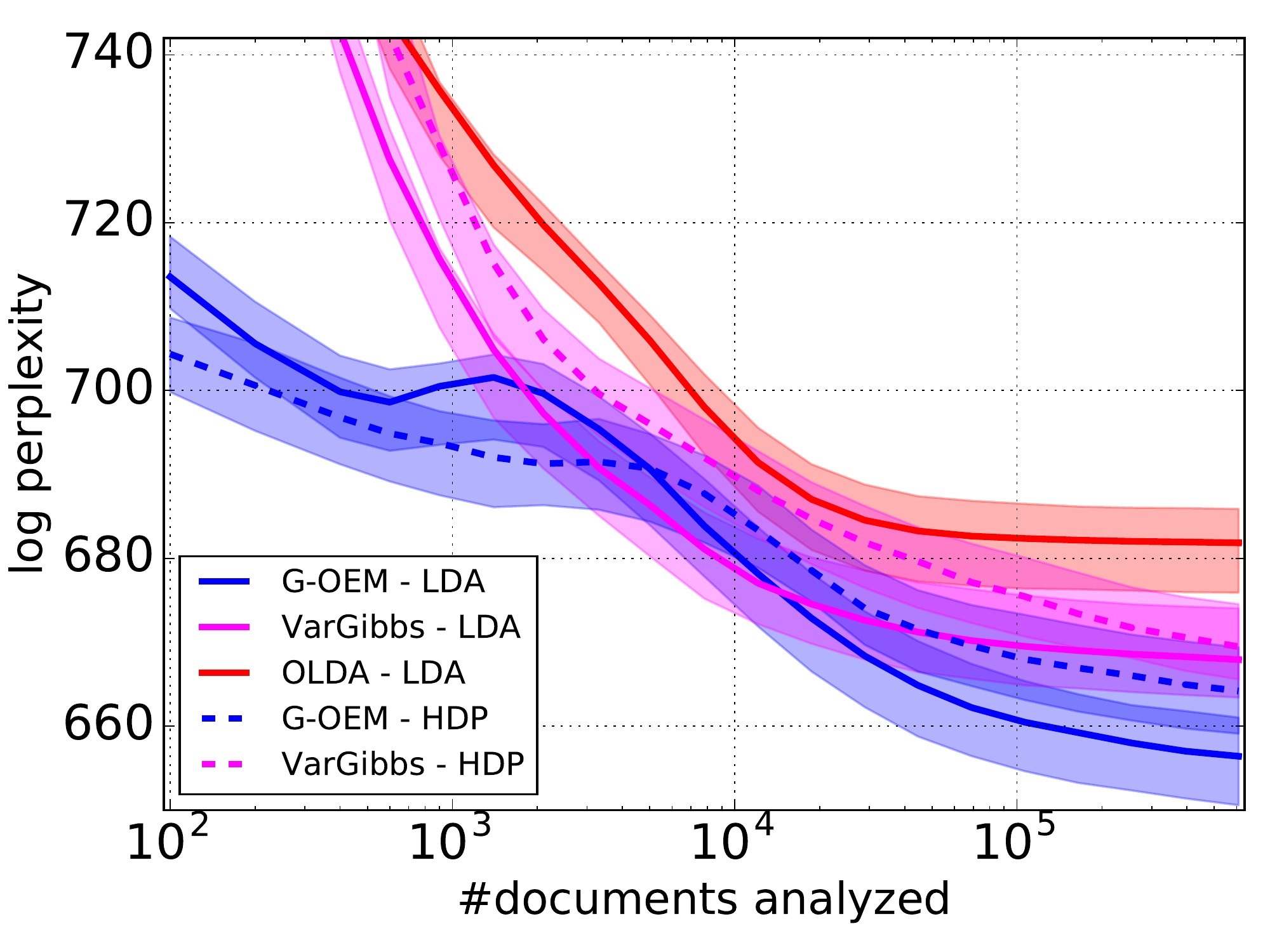}}
\subfigure[Wikipedia]{\includegraphics[width=0.48\columnwidth]{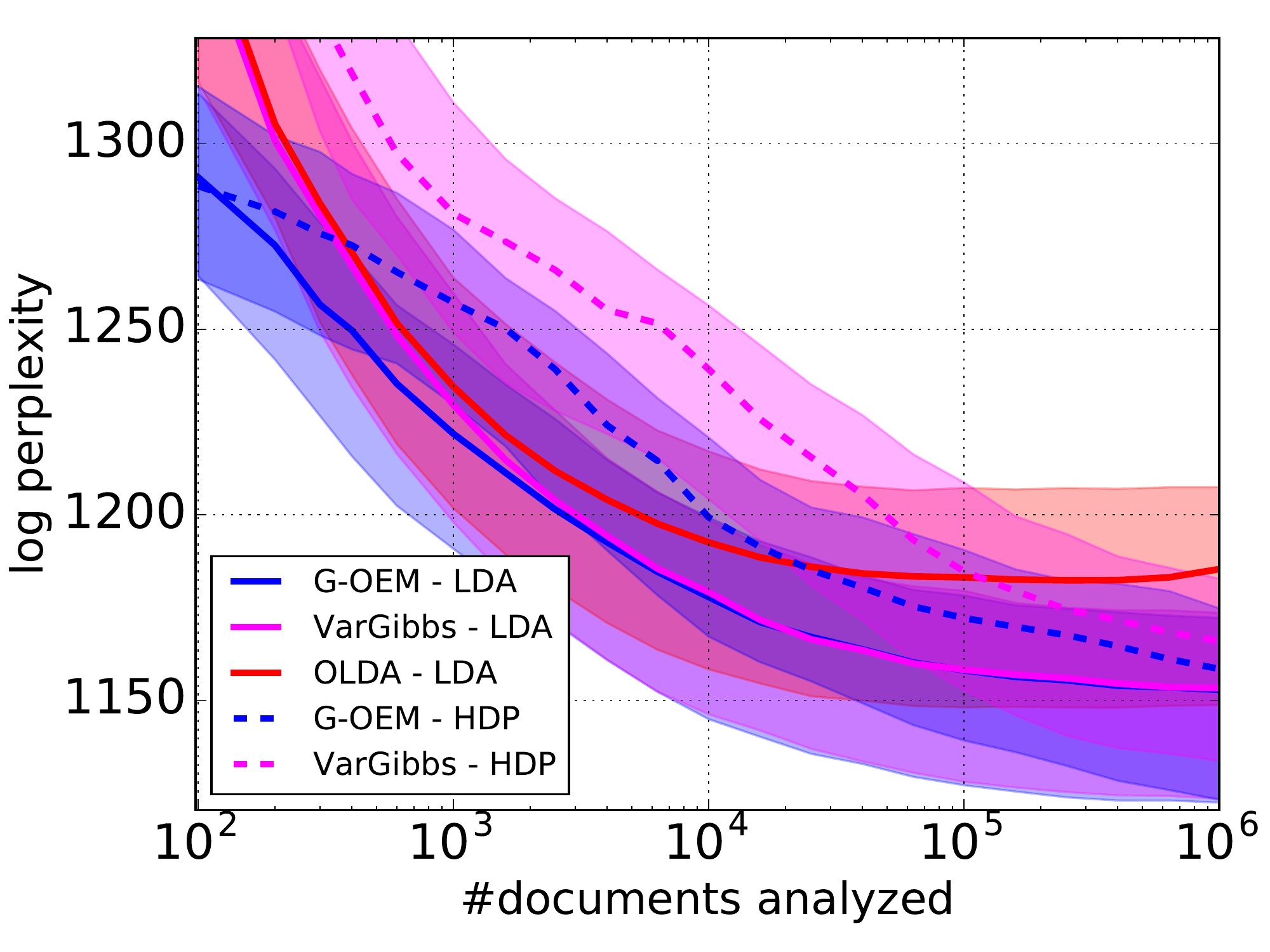}}
\caption{Perplexity through iterations on different test sets with \texttt{G-OEM} and \texttt{VarGibbs} applied to both LDA and HDP. Best seen in color.}
\label{fig:HDP_var}
\end{center}
\end{figure} 

\clearpage
\bibliography{bib}

\end{document}